\documentclass[letterpaper]{article} % DO NOT CHANGE THIS
\usepackage{aaai24}  % DO NOT CHANGE THIS
\usepackage{times}  % DO NOT CHANGE THIS
\usepackage{helvet}  % DO NOT CHANGE THIS
\usepackage{courier}  % DO NOT CHANGE THIS
\usepackage[hyphens]{url}  % DO NOT CHANGE THIS
\usepackage{graphicx} % DO NOT CHANGE THIS
\usepackage{multirow}
\usepackage{booktabs}
\usepackage{natbib}  % DO NOT CHANGE THIS AND DO NOT ADD ANY OPTIONS TO IT
\usepackage{caption} % DO NOT CHANGE THIS AND DO NOT ADD ANY OPTIONS TO IT
\usepackage{algorithm}
\usepackage{algpseudocode}
\usepackage{amsmath}
\usepackage{amssymb}
\usepackage{amsfonts}
\usepackage{amsthm, bm}
\usepackage[dvipsnames]{xcolor}

\usepackage{tikz}
\usetikzlibrary{calc,matrix,positioning, patterns}
\usepackage{esvect}
\usepackage{xspace, pifont}
\usepackage{xurl}
\usepackage{tcolorbox}
\usepackage{newfloat}
\usepackage{listings}
\DeclareCaptionStyle{ruled}{labelfont=normalfont,labelsep=colon,strut=off} % DO NOT CHANGE THIS
\floatstyle{ruled}
\newfloat{listing}{tb}{lst}{}
\floatname{listing}{Listing}
\usepackage{hyperref}
\definecolor{mydarkblue}{rgb}{0,0.08,0.45}
\hypersetup{ %
pdftitle={},
pdfsubject={Proceedings of the International Conference on Machine Learning 2024},
pdfkeywords={},
pdfborder=0 0 0,
pdfpagemode=UseNone,
colorlinks=true,
linkcolor=mydarkblue,
citecolor=mydarkblue,
filecolor=mydarkblue,
urlcolor=mydarkblue,
}

\newcommand{\cmark}{\ding{51}}%
\newcommand{\xmark}{\ding{55}}%

\usepackage{amsmath, mathtools, amssymb}
\usepackage[capitalise]{cleveref}

\usepackage{subcaption, caption}
\usepackage{bm}
\usepackage{adjustbox}
\usepackage{tikz}

\usetikzlibrary{calc,matrix,positioning,patterns,shapes.geometric}
\DeclareMathOperator{\Var}{Var}

\newtheorem{definition}{Definition}
\newtheorem{lemma}{Lemma}

\newtheorem{theorem}{Theorem}
\newtheorem{corollary}{Corollary}
\crefname{condition}{condition}{conditions}
\Crefname{condition}{Condition}{Conditions}
\crefname{example}{example}{example}
\Crefname{example}{Example}{Example}
\Crefname{section}{Section}{Section} % Use small cref to print Sec., Fig. Tab.
\crefname{section}{Sec.}{Sec.} % Use small cref to print Sec., Fig. Tab.
\crefname{figure}{Fig.}{Figs.} 
\Crefname{figure}{Figure}{Figures} 
\Crefname{table}{Table}{Tables} 
\crefname{table}{Tab.}{Tab.} 
\Crefname{equation}{Equation}{Equations} 
\crefname{equation}{Eqn.}{Eqns.} 

\DeclareMathOperator*{\argmin}{arg\,min}

\newcommand{\aii}[0]{Availability Inference Restriction\xspace}

\newcommand{\air}[0]{Availability Inference Restriction\xspace}
\newcommand{\puc}[0]{Protected User Consent\xspace}
\newcommand{\PUC}[0]{PUC\xspace}
\newcommand{\bx}[0]{\mathbf{x}} % The entire feature vector
\newcommand{\Bx}[0]{\mathbf{X}} % The entire feature vector RV
\newcommand{\bw}[0]{\mathbf{w}} 
 
\newcommand{\bv}[0]{\mathbf{v}}
\newcommand{\bu}[0]{\mathbf{u}} 
\newcommand{\bq}[0]{\mathbf{q}} 
\newcommand{\indset}[0]{\mathcal{I}}
\newcommand{\BOne}[0]{\bm{1}}
\newcommand{\bz}[0]{\bm{z}} % The base feature vector
\newcommand{\Bbz}[0]{\bm{Z}} % The base feature vector
\newcommand{\bb}[0]{\mathbf{b}} % The base feature vector
\newcommand{\Bb}[0]{\mathbf{B}} % The base feature vector
 
\newcommand{\bzstar}[0]{\bm{z}^{*}} % The imputed
\newcommand{\Bbzstar}[0]{\bm{Z}^{*}} % The imputed
\newcommand{\Bbm}[0]{\bm{a}}
\newcommand{\bbm}[0]{a}
\newcommand{\Bba}[0]{\bm{A}}
\newcommand{\bba}[0]{\bm{a}}
\newcommand{\spaceXb}[0]{\mathcal{X}^b}
\newcommand{\spaceXz}[0]{\mathcal{X}^z}

\newcommand{\spaceX}[0]{\mathcal{X}}
\newcommand{\spaceY}[0]{\mathcal{Y}}
\newcommand{\distp}[0]{\mathbf{p}}
\newcommand{\distpo}[0]{{\overline{\mathbf{p}}}}
\newcommand{\distq}[0]{\mathbf{q}}

\newcommand{\expect}[0]{\mathbb{E}}
\newcommand{\fexpect}[0]{\mathbb{F}^\mathcal{L}}
\newcommand{\expectw}[2]{\mathbb{E}_{#1}\left[#2\right]}
\newcommand{\OFFIDA}[0]{\texttt{PUCIDA}\xspace}
\newcommand{\bfm}[0]{base feature model\xspace}
\newcommand{\bfms}[0]{base feature models\xspace}
\newcommand{\ffm}[0]{full feature model\xspace}
\newcommand{\ffms}[0]{full feature models\xspace}
\newcommand{\Bfm}[0]{Base feature model\xspace}
\newcommand{\Ffm}[0]{Full feature model\xspace}
\newcommand{\indicone}[0]{\mathbb{I}}

\newcommand{\indep}{\perp\!\!\!\!\perp} 
\title{I Prefer Not To Say: Protecting User Consent in Models with\\ Optional Personal Data}
\author{
    Tobias Leemann\textsuperscript{\rm 1,\,2}, Martin Pawelczyk\textsuperscript{\rm 3}, Christian Thomas Eberle\textsuperscript{\rm 1},
    Gjergji Kasneci\textsuperscript{\rm 2}
}
\affiliations{
    \textsuperscript{\rm 1}University of Tübingen, Tübingen, Germany \\
    \textsuperscript{\rm 2}Technical University of Munich, Munich, Germany \\
    \textsuperscript{\rm 3}Harvard University, Cambridge, MA, USA \\
    tobias.leemann@uni-tuebingen.de, martin.pawelczyk.1@gmail.com,
    ct.eberle@protonmail.ch,
    gjergji.kasneci@tum.de
}

\newcommand{\wappendix}{1}
\newcommand{\appref}[2]{\ifdefined\wappendix\Cref{#1}\else Appendix~#2\fi}
\newcommand{\aaaisec}[1]{\ifdefined\wappendix\section{#1}\else \section*{#1}\fi}
\newcommand{\aaaisubsec}[1]{\ifdefined\wappendix\subsection{#1}\else \subsection*{#1}\fi}

\begin{document}
\maketitle

\begin{abstract}
We examine machine learning models in a setup where individuals have the choice to share optional personal information with a decision-making system, as seen in modern insurance pricing models.
Some users consent to their data being used whereas others object and keep their data undisclosed. 
In this work, we show that the decision not to share data can be considered as information in itself that should be protected to respect users' privacy.
This observation raises the overlooked problem of how to ensure that users who protect their personal data do not suffer any disadvantages as a result.
To address this problem, we formalize protection requirements for models which only use the information for which active user consent was obtained. 
This excludes implicit information contained in the decision to share data or not.
We offer the first solution to this problem by proposing the notion of Protected User Consent (PUC), which we prove to be loss-optimal under our protection requirement. 
We observe that privacy and performance are not fundamentally at odds with each other and that it is possible for a decision maker to benefit from additional data while respecting users' consent.
To learn PUC-compliant models, we devise a model-agnostic data augmentation strategy with finite sample convergence guarantees. Finally, we analyze the implications of PUC on challenging real datasets, tasks, and models.
%Finally, we analyze the implications of PUC on a variety of challenging real datasets
\end{abstract}

\aaaisec{Introduction}\label{sec:intro}
%Data-driven models are being widely deployed to support critical decision-making. 
%Such crucial scenarios include hiring decisions \citep{bogen2018help, raghavan2020mitigating, tambe2019artificial}, loan approvals \citep{dastile2020statistical, xia2017boosted, ala2022deep}, %judicial parole decisions\,\cite{Angwin2016, dressel2018accuracy},
%personalized pricing \cite{ban2021personalized, den2015dynamic}, and disease risk prediction or diagnosis~\cite{park2018methodologic,louro2019systematic}. %, or facial-recognition~\cite{dooley2022robustness}.
While the day-to-day impact of automated data processing is steadily growing, modern regulations such as the European Union's General Data Protection Regulation (GDPR)~\cite{regulation2016regulation} or the California Consumer Privacy Act (CCPA)~\cite{ccpa2021} strive to give individuals more control over their personal data. 
%In particular, the principle of \textit{data minimization} stated in Article 5 of the GDPR demands that personal data shall be \textit{``adequate, relevant and limited to what is necessary in relation to the purposes for which they are processed''}\cite{regulation2016regulation}. 
In light of these regulations, we consider machine-learned classifiers in which individuals have the freedom to decide themselves on which data they would like to provide to an automated decision system.
%In light of these regulations, we explore the increasingly relevant scenario where machine learning classifiers are involved, and individuals can make their own choices on the information they wish to provide to a data-driven decision system.

Such systems are increasingly being deployed \cite{sydneymorning2023insurerssmartwatch}: As a running example, we consider a realistic use-case of health insurance pricing: 
Suppose in an automated pricing model all potential customers are asked to fill out an application form where they enter certain \emph{base features}, for instance information such as their state of residence and age.
To improve the pricing model, the insurance offers an additional service, a ``companion fitness app'', through which additional health data about the customer's physical condition are collected. 
%and workouts are collected. %transmitted to the company.
The customers decide whether to use the app or not; alternatively, customers can sign up for a policy without consenting to use the app. 
The health data that customers share may however influence the premium of the insurance policy they receive. 
We refer to data that provide additional, non-mandatory information beyond the base features as \textit{optional features}. 
With fitness trackers and smartwatches rapidly gaining popularity \cite{reeder2016health, zimmer2020there, statista2023wearables}, such systems are already being deployed in practice, e.g., by major health insurance firms in Australia \cite{sydneymorning2023insurerssmartwatch}.

% and show an example of this kind of dataset in \Cref{tab:datasample}.
\begin{figure}
\input{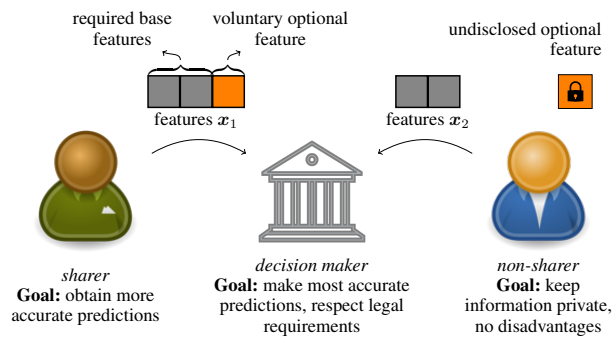}
\caption{
Overview of the relevant stakeholders. We consider a case where users can voluntarily provide information on optional features or choose to leave them undisclosed. The goals of sharers, non-sharers, and the decision maker have to be reconciled.}
%We show calibration curves for a model without fairness considerations (a) and with common fairness constraints enforcing statistical parity (b) and equalized odds (c).
%The first three models can penalize users not sharing the optional information (fairness gap in left panel), whereas a model trained with Optional Feature Fairness (d) exhibits no systematic bias.
%Models are probabilistic Random Forests trained on a synthetic dataset (see \Cref{sec:app_teaseroff}) \textbf{New Figure needed}. \label{fig:teaseroff}}
\end{figure}

The outlined scenario is challenging as there are three groups of stakeholders whose interests need to be reconciled:
(1) The group of non-sharing individuals who do not want to provide additional information, for instance due to privacy concerns. 
We refer to them as \textit{non-sharers}.
For this group, the decision maker does not want to or cannot force them to provide the additional information for legal reasons. 
Consequently, the non-sharers do not want the additional information to be considered in the decision making process; 
in return, they are willing to sacrifice some accuracy, but they do not want to face other systematic disadvantages. 
(2) On the other hand, individuals who voluntarily share data (\textit{sharers}) explicitly want the additional information to be considered and want to obtain more accurate predictions.
(3) Finally, the decision makers themselves desire the most accurate predictions with the lowest overall costs while respecting the users' privacy and legal requirements.

Among these requirements, it is crucial to the non-sharers to explicitly exclude the information contained in the decision to share or not to share.
To see this, we note that smartwatch users are more likely to exercise in general than non-wearers \cite{valuepenguin2023healthcare} which usually create lower costs for the insurance company as fitter customers take less sick days on average.
Thus, only through observing the decision to share data, the insurance firm could make inferences about a person's fitness. 
This is problematic for two reasons: 
% Reason 1: "Right to resonalbe inference", it is a privacy infringement"
First, the company would unethically infer private data, that the non-sharers  explicitly did not give consent to. Prior work \cite{wachter2019right} has argued for a ``right to reasonable inferences".
This rules out inferences from unrelated factors that are purely predictive and may infringe privacy, as they open the door for discriminatory and invasive decision-making \cite{mittelstadt2016ethics}.
%This might be considered illegal disparate treatment \cite{lohaus2022two}. 
% Reason 2: It is deemed illegal in many countries.
Second, this would lead to non-sharers being assigned a higher insurance premium than the estimate of the legacy model which only considered their base features. 
% We will show that this increase can be considerable  in our experimental section. 
% A premium increase for non-sharers can be interpreted as implicitly forcing users to provide data and is considered illegal discrimination in many countries.
Many countries have laws that prohibit insurers from raising the base premium for users who do not share their data, as this is seen as a coercive and unfair practice.
For example, the US only permits five factors to affect the premium, which are location, age, tobacco use, plan category, and dependent coverage \cite{usgov2023healthcare}.
It is however possible for insurers -- and desired by many users -- to award bonuses which reduce the premium based on participation in optional reward and incentive programs \cite{madison2013smoking, sydneymorning2023insurerssmartwatch}. %In other cases, the implicit usage of this information could introduce some bias in the decision.

To summarize, we study machine learning models that can handle optional features and meet legal requirements and desiderata of three groups of stakeholders: the sharers, the non-sharers, and the decision makers.
We consider it essential for these models to not make inferences based on the unavailability of a feature value for the non-sharers, a constraint that we term \textit{\air (AIR)}.
%Additionally, we require the model to incentivize users to provide optional data by effectively using this data if it is made available, which we call \emph{Incentivization}. 
Finally, we are interested in obtaining models with optimal performance under this requirement. %We show that the resulting model manages to reconcile the desiderata of the three stakeholders and can be interpreted as a bonus system. Such systems are already being deployed in practice, for instance by insurance firms. Therefore, our work provides them a solid theoretic grounding by explicitly distilling the premises underlying such systems and deriving them from first principles. We further present a constructive machine learning approach to directly learn optimal and  compliant models from scratch.

%Our desiderata balance the interests between three groups of stakeholders: \emph{individuals who do not want to share information} on optional features and demand that only the provided data should be used in decision-making; \emph{individuals who share information} on the optional features, who want this information to make a difference for the outcome if it is informative; and \emph{decision makers}, who want to make the most accurate decision.
%

% OFF is loss optimal under req (i) and (ii) (lemma 4.1)
% OFF obeys "predictive non-degradation"
% Guarantees for OFF-inducing data augmentation
% -> equivalence result (optimal model on phat is equivalent to OFF)
% -> finite sample consistency result (Theorem 5.2): how well can we achieve this optimal model using finite samples

\textbf{Contribution.} 
We address the problem of how to fairly and privately predict outcomes for users who share optional data and those who do not. 
We tackle this overlooked issue by making the following contributions:
\begin{itemize}
\setlength\itemsep{0.0cm}
\item  \textbf{Definition.} We introduce models with \puc (PUC), which are optimal under our protection requirement AIR. %ensures that the predictions respect the \aii~of the non-sharers and achieve the best possible model performance.
%PUC models outperform or match the performance of a model trained only on the base features, 
We derive performance guarantees, which formally show that it is possible to reconcile the decision maker’s interest in improved predictions and the non-sharer’s privacy preferences.
\item \textbf{Algorithm.} We propose a PUC-inducing data augmentation (\OFFIDA) technique that can be applied to any type of predictive architecture (e.g., tree or neural network) and any convex loss function (e.g., mean squared error or cross-entropy loss) to obtain such models
\item \textbf{Analysis.} We prove that predictive models trained with \OFFIDA satisfy PUC asymptotically, and provide finite sample convergence results that demonstrate that \OFFIDA produces PUC-compliant models in practice.
\item \textbf{Empirical evaluation.} We empirically show that without enforcing PUC, the average absolute prediction outcome (e.g., insurance quote) of users who do not share data can be almost 20\,\% worse than justified by their base data. 
We then evaluate our data augmentation technique on various ML models and show that PUC is achieved regardless of the model.
\end{itemize}

\aaaisec{Related Work}
In this Section, we review the most relevant streams of related work (see \appref{sec:app_relatedwork}{A.1} for additional references).

\label{sec:relatedwork}
\textbf{Classification with Missing Values.} %or Labels.} 
Classification models that can handle missing data have been studied previously with the goal of minimizing costs or increasing performance \cite{zhang2005missing, aleryani2020multiple}, obtaining uncertainty estimates \cite{kachuee2020generative}, or fulfilling classical fairness notions \cite{zhang2021assessing, jeong2022fairness, wang2021analyzing, fernando2021missing}. 
However, the mechanisms underlying missingness is different in this work, as missing values indicate explicit non-consent by the user, leading to different implications.
In a related line of work, classification with noisy \cite{fogliato2020fairness} or missing labels \cite{kilbertus2020fair, rateike2022dontthrowaway}  has been investigated, where the missingness is often a result of \textit{selection bias}. 
The setting considered in this work is different in the sense that we are not concerned with fulfilling a fairness notion with respect to a sensitive attribute, %in the presence of missing values or labels
but consider the interests of subjects that have and have not provided optional information.

\textbf{Data Minimization.} The principle of Data Minimization is anchored in the GDPR \cite{regulation2016regulation}. Data Minimization demands minimal data collection. Several works are concerned with implementing \cite{goldsteen2021data} or auditing compliance with this principle \cite{rastegarpanah2021auditing}. Rastegarpanah et al. \cite{rastegarpanah2020fair} consider decision systems that can handle optional features from a data minimization perspective where the decision maker decides which features are collected for each individual.
This principle is distinct from the ``right to be forgotten'' \cite{biega2020operationalizing}, which enables individuals to submit requests to have their data deleted.
In response to these regulations, several works consider the problem of updating an ML model without the need of retraining the entire model \cite{wu2020deltagrad,ginart2019making,izzo2021,Golatkar_2020_CVPR} or the effect of removals on model explanations \cite{rong2022evaluating, pawelczyk2022trade}.
%Finally, removing subsets of training data is an essential ingredient for the evaluation of explanation techniques \citep{hooker2019bench,rong2022evaluating,pawelczyk2022trade}. 
Our work differs from these works as 
%certain feature values may be unavailable from the data collection process. 
our goal is to train a model where users decide themselves which data they deem relevant through sharing one or many optional features.

\textbf{Algorithmic Fairness.} 
A multitude of formal fairness definitions have been put forward in the literature \cite{verma2018fairness}.
Examples include statistical parity \cite{dwork2012fairness}, predictive parity \cite{chouldechova2017fair}, equalized odds, equality of opportunity \cite{hardt2016equality}, and individual fairness \cite{dwork2012fairness}. However, they are still a topic of discussion, for instance, because these definitions are known to be incompatible \cite{kleinberg2016inherent, lipton2018does}. % and it has recently been argued that even fair models may be prone to disparate treatment \cite{lohaus2022two}.
%However, even some of the basic definitions are incompatible with one another \cite{kleinberg2016inherent} forcing a choice between conflicting fairness notions.
Additionally, there are a several definitions that rely on causal mechanisms to assess fairness, e.g., counterfactual fairness \cite{kusner2017counterfactual}, and the notion of unresolved discrimination \cite{kilbertus2017avoiding}. 
While causal approaches to fairness might be preferable, they require information about the causal structure of the data generating process. 
Moreover, it has recently been shown that causal definitions may lead to adverse consequences, such as lower diversity \cite{nilforoshan2022causal}.
%For these reasons, we take a non-causal, observational perspective in this work.
We discuss how existing fairness definitions could possibly be applied to the setting with optional features, but we find that none of the fairness definitions aligns with our desiderata theoretically and experimentally (see \appref{sec:presentdefinitions}{A.2}).

\textbf{Strategic Classification.}
In an even broader context, this work also relates to the field of strategic classification \cite{hardt2016strategic}. However, it is worth noting that in strategic classification research, the focus primarily revolves around users strategically manipulating their features for optimal outcomes, which may also involve information withholding \cite{krishnaswamy2021classification}. In contrast to our work, privacy concerns are neglected in this research stream.
As far as we are aware, there are no prior works on the specific problem of balancing the interests of \emph{all three} groups of stakeholders (the non-sharers, sharers, and the decision makers).

\aaaisec{Problem Formulation}
\label{sec:form}
\aaaisubsec{Formalization and Notation}
In this work, each data instance contains a realization of a number of base features $\bb \in \spaceXb$, where $\spaceXb \subseteq \mathbb{R}^n$ is the space of the base features. 
Furthermore, let there be some optional information $z \in \spaceXz$, where $\spaceXz \subseteq \mathbb{R}$ is the value space of the optional feature.\footnote{We extend our definitions to integrate multiple optional features a later section.}
It is the users' choice to decide if they want to disclose $z$ to the system, which results in an availability variable $a \in \left\{0,1\right\}$. Accordingly, only imputed samples $z^*= \left\{z~\text{if}~a{=}1,~\text{else}~ \texttt{N/A}\right\}$ are observed, where a value of \texttt{N/A} indicates that a user did not reveal the optional information, e.g., did not use the companion app.
In summary, the data observations are tuples $\bx = (\bb, a, z^*)$ that reside in $\spaceX = \spaceXb \times \left\{0,1\right\} \times (\spaceXz \cup \left\{\texttt{N/A}\right\})$. 
Each training sample comes with a label $y \in \spaceY$. Further, there is a data generating distribution $\distp$ with support $\spaceX \times \spaceY$ and we have access to an i.i.d.\ training sample $(\bx, y) \sim \distp$. \Cref{tab:datasample} shows such a data sample. We denote the random variables for the respective quantities by $\bm{B}, A, Z, Z^{*}, Y$. The label is probabilistically determined through the base features $\Bb$ and the hidden feature $Z$ but the sharing decision does not influence the true label for a given $\Bb, Z$, such that $Y\indep A | \Bb, Z$. %that are governed by the graphical model depicted in \cref{fig:gengraphmodel}.
\begin{figure}
\ifdefined\arxiv
\setlength{\tabcolsep}{4pt}
\else
\setlength{\tabcolsep}{0.5pt}
\fi
\centering
\begin{tcolorbox}[boxsep=-1.5mm,hbox]
\adjustbox{max width=0.44\textwidth}{
\begin{tabular}{cccccc}
\multicolumn{2}{c}{base features~$\bb$} & opt. feat.~$z^{*}$ & $a$ & label $y$\\
\cmidrule(lr){1-2}\cmidrule(lr){3-3}\cmidrule(lr){4-4}\cmidrule(lr){5-5}
state & plan & fitness score & avail. & treatment costs \\
\cmidrule(lr){1-1}\cmidrule(lr){2-2}\cmidrule(lr){3-3}\cmidrule(lr){4-4}\cmidrule(lr){5-5}
New South Wales & basic & 87\,\% & 1 & 3k\$ \\
Queensland  & gold & $\texttt{N/A}$ & 0 & 17k\$ \\
New South Wales & basic & 92\,\% & 1 &  5k\$ \\
New South Wales & basic & $\texttt{N/A}$ & 0 & 64k\$\\
Victoria & premium & 56\,\% & 1 & 22k\$ \\
\end{tabular}}
\end{tcolorbox}\ifdefined\arxiv\vspace{2mm}\fi
\caption{Samples for the insurance use-case. We have two base features $\bb$ and one optional feature $z^*$, which either takes an observed value $z$, or it takes a value of \texttt{N/A} if unobserved. The variable $a \in \left\{0,1\right\}$ indicates the availability of the feature. The goal is to predict the label $y$.
}
\label{tab:datasample}
\end{figure}
% \begin{wrapfigure}[10]{l}{0.33\textwidth}
% \centering
% \vspace{-0.5cm}
% \input{graphicalmodel}
% \caption{Most general graphical model of missingness considered in this work. The true $Z$ is unobserved and $Y$ is only observed for observations in the train set.}
% \label{fig:gengraphmodel}
% \end{wrapfigure}
% \begin{figure}
%     \centering
%     \input{graphicalmodel}
%     \caption{Most general graphical model of missingness covered in this work. The true $Z$ is unobserved and $Y$ is only observed for observations in the train set.}
%     \label{fig:gengraphmodel}
% \end{figure}

In many applications, the goal is to find a function $f: \spaceX \rightarrow \spaceY$ that models the observed data. %Note that this formulation already describes a function that can handle both samples with and without the optional feature present. 
%In the scope of this work, we consider functions $f$ that map to a continuous domain $\spaceY$. 
In particular, $f: \spaceX \rightarrow [0,1]$ may predict a probability of a positive outcome or $f: \spaceX \rightarrow \mathbb{R}$ may return a numerical score.
The test data for which the model will be used come from the same distribution $\distp$, though with the label $y$ unobserved, and we suppose that the information provided is always correct.
We consider a convex loss function $\mathcal{L}: \spaceY \times \spaceY \rightarrow \mathbb{R}$, e.g., mean-squared-error (MSE) or binary cross entropy (BCE), for which we minimize the expected loss for a sample from the data distribution. 
For instance, using the common MSE loss $\mathcal{L}(f(\bx),y)=(f(\bx)-y)^2$, an optimal predictor is given by $f^{*}_\mathcal{L}(\bx) = \argmin_{f(\bx)}\expectw{\distp(Y|\bx)}{(f(\bx)-Y)^2} = \expect[Y|\bx]$, the conditional expectation.
%\footnote{Supposing $\distp(Y|\bx)$ is square-integrable for all $\bx$.}
However, this notion can be generalized to other loss functions: An optimal predictor $f^*_\mathcal{L}(\bx)$ for the loss function $\mathcal{L}$ fulfills $\forall \bx$:
\begin{align}
f^*_\mathcal{L}(\bx) = \mathbb{F}^{\mathcal{L}}_\distp\left[Y|\bx\right] \coloneqq \argmin_{f(\bx)} \expectw{\distp(Y|\bx)}{\mathcal{L}(f(\bx),Y)}.
\label{eq:generalized_erm}
\end{align}
We use $\mathbb{F}^{\mathcal{L}}[Y|\bx]$ to denote a generalized expected value that minimizes the expected loss conditioned on $\bx$. 
To ease our derivations, we suppose this minimum to be unique and finite. %\footnote{If the $\argmin$ is not unique, the same results can be obtained when requiring $f^*_\mathcal{L}(\bx)$ to return one specific member from the set of minima.}
Intuitively, it represents the best guess of $Y$ given $\bx$.
For the MSE-Loss, $\fexpect$ is equivalent to the expectation operator $\expect$.
In the following statements, the reader may thus mentally replace $\fexpect$ with an expectation $\mathbb{E}$ without further ramifications in order to get the high level intuition.
%For notational convenience, we define an extraction operator $\extrb: \spaceX \rightarrow \spaceXb$ that returns only the base features from a sample.
Finally, we introduce two key terms, namely, \emph{\bfm} and \emph{\ffm}.
The former refers to a model trained on the base features only, while
the latter refers to a model trained on all features where some strategy is used to replace unavailable feature values.
Typically these strategies are called \emph{imputation} and replace unavailable values by zeros, a feature's mean or median \citep{emmanuel2021survey}.%; these strategies are usually called \emph{imputation}.

\aaaisubsec{Desiderata}
Our goal is to learn models $f: \spaceX \rightarrow \spaceY$ that comply with the desideratum of \emph{\aii}, which we briefly introduced in Section \ref{sec:intro}, to protect the interests of the non-sharers. Under this constraint, the model should provide the best predictive performance to reflect the need of the sharers and the decision maker for most accurate predictions.

%Below we crisply formalize these two notions.
%\textbf{Desideratum 1: ncentivization.}
%First, we consider the case where optional information is provided. 
%We define Incentivization to enforce that individuals, who \emph{share optional data}, receive the most accurate prediction score possible in the decision-making process. This constrains $f_{\mid_{a=1}}$, i.e., the prediction for the set of inputs where optional information is available.
%\begin{definition}[Incentivization]
%\label{def:incentivization}
%\textit{Using the generalized expectation introduced in \cref{eq:generalized_erm}, Incentivization requires:}
%\begin{align}
%\begin{split}
%& f_{\mid_{a=1}}(\bb, a, z^*) = \fexpect\left[Y|\bb, A=1, z^{*}\right] \\
%& \coloneqq\argmin_{f(\bb,a,z^{*})} \expectw{\distp(Y|\bb,A=1,z^*)}{\mathcal{L}(f(\bb,a,z^{*}),Y)}.
%\end{split}
%\end{align}
%\end{definition}
%Definition \ref{def:incentivization} ensures that for individuals providing optional information, all available information is used to make the most accurate prediction of the outcome, satisfying the group's desire for the optional information to impact the outcome and the decision maker's interest in accurate predictions. %i.e., the outcome with minimal loss. This covers th

\textbf{Desideratum 1: \aii.} We start by considering the intricate case of  individuals who \emph{do not want to share optional information}.
In this case, the model should compute the prediction based on the information the user gave their consent to. 
In particular, (a) the model should only use the base features \emph{and} (b) should not use information that could be derived from the unavailability of the optional features to compute the prediction to avoid violating the user's consent. %and implicit force to reveal information. %[\textbf{}\textbf{GIVE EXAMPLE}]

For (a), this requires that the predictor does not use the information as an explicit input, i.e., the predictor should behave as if it only used base features $\bb$ via some function $g: \spaceXb \rightarrow \spaceY: f_{\mid_{a=0}}(\bb, a, z^*) = g(\bb)$. 
% Let us restrict ourselves to the classification case first, where $\spaceY = [0,1]$ %predicts a probability such that and a conditional probability $\distp_g$ of the outcome $\distp_g(Y=1|\bb) \coloneqq g(\bb)$ can be defined.
For (b), although a=0 is not an explicit input to $g$, a sufficiently complex function may still be implicitly adapting to the group $a=0$ and thus incorporate information that the user did not give their consent to.
%On the other hand, we have $\distp(Y|\bb)$. 
We would like to make sure that the predictions of $g$ cannot use more information than contained in the overall conditional distribution, given the base features $\bb$. This overadaption can be prevented by constraining the model's loss on the population of non-sharers to match the loss of the optimal base model $f_{\mathcal{L}}^{*}$ on this population. The reasoning behind this rationale is that all models that would beat the performance of this model must implicitly use some additional side knowledge about this group that was not provided by the users.

\begin{definition}[\aii]
\label{def:nonpenalization}
%\textit{
For individuals that choose not to provide the optional feature ($a{=}0$), only the provided data $\bb$ is used to compute the outcome in the decision process, i.e., $f_{\mid_{a=0}}(\bb, a, z^*){=}g(\bb)$,
where $g:\spaceXb \rightarrow \spaceY$ is a \bfm. Further, we require
%In the probabilistic two-class case where we obtain probablistic predictions $G(\bb) \sim \distp_g(Y|\bb)$ and have  marginal likelihood  independent of $A$ is given by $\hat{Y}(\bb) \sim \distp(Y|\bb)$ this requires
%\begin{align}
%    I(G(\Bb);Y|A=0) \leq I(\hat{Y}(\Bb);Y|A=0)
%\end{align}
%where $I$ denotes mutual information. This notion can be generalized to non-classification problems with an arbitrary loss function $\mathcal{L}$ where it requires %Generalized to arbitrary loss functions $\mathcal{L}$ (\Cref{sec:generalizednonpen})  this requires:
%, i.e., it cannot be more informative about the label for this group than an optimal predictor $f_{\text{BCE}}^{*}(\bb)=\expect\left[Y|\bb\right]$ trained on the entire dataset using the base features, i.e.,}
% \[
% \text{MI}\left[Y,g(\Bb)\middle|A=0\right] \leq \text{MI}\left[Y,f_{\text{BCE}}^{*} (\Bb)\middle|A=0\right].
% \]
\begin{align}
\expect\left[\mathcal{L}\left(g(\Bb), Y\right)\middle|A=0\right] \geq \expect\left[\mathcal{L}\left(f_{\mathcal{L}}^{*} (\Bb), Y\right)\middle|A=0\right].
\end{align}
%Furthermore, this outcome should not be specifically adapted to this subgroup. Therefore, letting $f_{\mid_{a=0}}$ indicate the restriction of $f$ to samples without optional features ($a=0$), we require 
\end{definition}
This definition summarizes our intuition that the information encoded through the unavailability of feature information should neither be used explicitly (a) nor implicitly (b).
We show how this constraint can analogously be derived from information-theoretic considerations in \appref{sec:app_probderivationnonpen}{B.3}.
%through the Kullback-Leibler Divergences as a measure. %predictors theoretic quantities
%used when computing the prediction line with requirement (b). % score in the presence of undisclosed features. 
%Thus, we make sure that $g$ does not implicitly leverage information specifically contained in the group of non-consenting users with $a{=}0$.

\textbf{Desideratum 2: Optimality.} Our \Cref{def:nonpenalization} restricts the information that the predictor can use when the optional information is unavailable. 
To meet the interests of the decision maker and the sharers, we also want to find models with optimal performance, i.e., lowest loss, under this constraint.

\aaaisec{Protecting User Consent}\label{sec:off_definition}
%In the previous Section, we have established that common fairness definitions fail to conform to our requirements.
We are therefore looking for an \textit{optimal} model within the class of predictors that comply with \air.
In this Section, we derive a novel notion called \puc (\PUC) that fulfills this purpose.
% We 

%When combined in Section %\ref{sec:off_definition}

% While Non-Penalization explicitly restricts the information that can be used, we do not wish to obtain models that perform arbitrarily bad. 
%, our desiderata will result in a predictor that uses the maximum amount of information while maintaining fairness.
%In Section \ref{sec:off_approaches}, we present an algorithm for finding this optimal predictor and show that it performs well even with finite data.

%To learn this kind of classifier $\bff_{\mid_{a=0}}$, we could train a model with the MSE loss on the subset of the data where $a=1$. Thus, the information in the incomplete samples will not be used at all for training. 
%As we will detail in the next section, we some advantages to constrain $\bff_{\mid_{a=0}}$ to be the average over all samples with identical base features, without constraining the missingness value, to leverage the full potential of the data while still maintaining an intuitive notion of fairness.

%As the common definitions of fairness considered in the previous section have their disadvantages when applied to the setup of optional features, we propose the novel notion of Optional Feature Fairness (OFF).

%

\aaaisubsec{One-Dimensional PUC}
%We first discuss what should happen in the case of $a=0$, i.e., an individual has not submitted optional information. To fulfill the requirement of Non-Penalization, we can only use the information in the base features $\bb$. The estimate with the lowest error would be $f_{\mid_{a=0}}(\bb) = \expect[Y|\bb]$. In the case $a=1$, by Incentivization the outcome $f_{\mid_{a=1}}(\bb, a, z^*)$ is determined to be $f_{\mid_{a=1}}(\bb, a, z^*)=\expect[Y|\bb, A=1, Z^*=z^*]$, the justified outcome. Combining these two cases results in a novel fairness notion that we term Optional Feature Fairness (OFF):

The next result encodes an intuitive notion of protection for the users that do not want to share data on the optional features ($a{=}0$): Their prediction under $f$ is then constrained to the best estimate for a user with the same base characteristics, no matter if additional data was provided. 
Contrarily, when additional information through the optional feature is provided, the predictor returns the best estimate using the available optional information:
\begin{theorem}[1D-\PUC] Let $f:\spaceX \rightarrow \spaceY \subseteq \mathbb{R}$ be a \ffm (i.e., including optional features). Among all predictors compatible with the \aii, a model $f$ with minimal loss is given by:
\begin{align*}
f^{*}_{\text{PUC}}\left(\bb, \bbm, z^*\right) =
\begin{cases}
\mathbb{F}^{\mathcal{L}}[Y|\bb], & \text{if}~a=0 \\
\mathbb{F}^{\mathcal{L}}[Y|\bb, A=1, Z^*=z^*] & \text{if}~a=1. 
\end{cases}
\end{align*}
%We denote this handling of additional information by one-dimensional Optional Feature Fairness (1D-OFF). 
\label{def:off1d}   
\end{theorem} 
We defer all proofs in this work to \appref{sec:app_proofs}{D}. \PUC is different from existing notions of group fairness, that do not fulfill the two desiderata in general (see \appref{sec:presentdefinitions}{A.2} for a discussion). Under the mentioned requirements, there is no model that can outperform $f^{*}_{\text{PUC}}$.
We stress that 1D-\PUC-compliant models have performance guarantees. 
These models match or improve upon an optimal \bfm $f^{*}_\mathcal{L}(\Bb) = \mathbb{F}_\mathcal{L}\left[Y|\bb\right]$. This model can be seen as an upper bound for practical models obtained after model selection. Therefore, models that can beat its performance may offer improvements even after extensive hyper-parameter tuning and model selection, a property which we refer to as Predictive Non-Degradation (PND): a model $f$ fulfills PND if its loss is smaller than that of the \bfm:
\begin{align}
\expect[\mathcal{L}\left(Y, f^{*}_{\mathcal{L}}(\Bb)\right)] \geq \expect[\mathcal{L}\left(Y, f(\Bb, A, Z^*)\right)].
\end{align}
We prove the following result:
\begin{corollary}[Predictive Non-Degradation of $f^{*}_ {\text{PUC}}$]
\label{corr:offnondegradation}
For any density $\distp$, a PUC-compliant model $f^{*}_ {\text{PUC}}$ fulfills Predictive Non-Degradation, i.e., it has a loss upper-bounded by the optimal \bfm $f^{*}_{\mathcal{L}}$.
%that of perfect model on the base features.
\end{corollary}
This is a remarkable result as it testifies that the decision maker can benefit from additional information in terms of loss, while protecting the privacy of users. 
This highlights that the interests of the different stakeholders are not contradictory and models that benefit all stakeholders do exist. 

\aaaisubsec{\PUC under Strategic Considerations and Monotonicity Constraints}%actions}
We have initially considered the case where the users desire the highest possible accuracy under data usage restrictions. 
%This is sensible if two outcomes with the same error are equally desirable. 
However, in some cases such as our initial insurance example, the motivation to receive a lower premium might be a more important concern to some users than receiving an accurate prediction or their privacy concerns. 
%In such scenarios, some users might be interested in obtaining the lowest premium instead of obtaining the most accurate decision. 
%Their decision to share optional information might also be based on obtaining the lowest premium possible.
%They might further base their decision to share optional information on this goal.
%We would like to highlight that the \PUC model is still optimal in such a setup if the decision maker cannot systematically increase the premium beyond the one demanded from the base model for the non-sharers. 
%This is reasonable in many cases, where legal constraints mandate that the decision maker cannot increase the premiums for non-sharers over this model, as outlined in the example in the introduction. 
If all users have full information (i.e., they see premiums with and without their optional data) and act strategically by sharing the value of $z$ only if it would decrease their premiums, we obtain the following result.
\begin{theorem}[Optimality of $f^{*}_ {\text{PUC}}$ under strategic actions]
Let $\distp'(\Bb, Z, Y)$ be any prior density on base features, true optional features and labels and let  $f(\bb, a=0, z)=\mathbb{F}^{\mathcal{L}}[Y|\bb]$, i.e., the decision maker uses the \bfm when no optional data is available.
Further suppose that users strategically choose to share the optional feature $z$ only if $f(\bb, a=1, z) \leq f(\bb, a=0, \texttt{N/A})$. Under these conditions, the model $f^{*}_{\text{PUC}}$ (\cref{def:off1d}) has minimal loss among all predictors. %that comply with the above conditions.
\label{corr:offstrategic}
%that of perfect model on the base features.
\end{theorem}
% \textbf{Proof Sketch.} To see this, we first note that the decision maker can only realize improvements for individuals by offering them a lower premium than the prediction of the base model -- otherwise, they would strategically not provide their data. It is only useful for the decision maker to do so if there exists an $y' \leq \mathbb{F}^{\mathcal{L}}[Y|\bb]$ with a lower expected loss. Due to the convexity of the loss we will also have $\mathbb{F}^{\mathcal{L}}[Y|\bb, z]\leq \mathbb{F}^{\mathcal{L}}[Y|\bb]$ and thus the optimal prediction would be $f(b, a = 1, z) =\mathbb{F}^{\mathcal{L}}[Y|\bb, z]$, which would again results in an OFF-model.
%decision maker
%OFF-model $f^{*}_{\text{OFF}}$ is not affected by the sharing decisions of the users given the prior density $\distp'(\Bb, Z, Y)$. The the base model stays unaffected and $\mathbb{F}^{\mathcal{L}}[Y|\bb, A=1, Z^*=z^*] = \mathbb{F}^{\mathcal{L}}[Y|\bb, A=1, Z=z] = \mathbb{F}^{\mathcal{L}}[Y|\bb, Z{=}z]$ as the sharing decision is independent of the label given the base and the optional features.
This result underlines that PUC models remain optimal if the decision maker cannot increase the premiums beyond the predictions of the current base model for the non-sharers. 
This is reasonable in many cases, where legal constraints mandate that the decision maker cannot implicitly force users to share data by inflating the base premium, as outlined in the introduction.
The sharing decision can also be automated for the users by simply dropping the optional feature if it does not lead to a decrease in premiums. This would result in the aforementioned bonus systems, where sharing more data cannot increase the premium. We show that among the class of models with such a monotonicity constraint, the outlined PUC-model with automatic sharing decisions is still optimal under the same conditions as in \Cref{corr:offstrategic} in \appref{sec:app_offstrategic}{D.5}.
\aaaisubsec{r-dimensional \PUC}
Next, we generalize our notion such that $r$ features can be provided optionally.
%and independently of each other. 
%While we have previously considered a single optional feature, one can easily construct examples where multiple features can be provided voluntarily and independently of each other. 
For example, the insurance firm might also accept voluntary results from prior medical examinations or diagnostic tests. 
Therefore, let there now be $r$ optional features such that $\bz{\in}\spaceXz_1 \times \cdots \times \spaceXz_r$ and $\bba{\in}\left\{0,1\right\}^r$, where $\spaceXz_i$ are the respective supports of each optional feature. 
By $\indset{\subseteq} [r]{=}\left\{1,\ldots,r\right\}$, we denote an index set that contains all feature indices present, i.e., $\indset(\bba){=}\left\{i~\middle|~\bba_i = 1, i=1,\ldots, r \right\}$. When we index vectors with this set, e.g., $\Bbz_\indset$, we refer to the subvector that only contains the indices in $\indset$.

\begin{definition}[\puc, \PUC]
Let $f: \spaceX \rightarrow \spaceY \subseteq \mathbb{R}$ be a \ffm.
%a model that can handle multiple optional features. 
The model $f^{*}_{\text{PUC}}$ that fulfills Protected User Consent is given by 
%the following equality holds:
\begin{align*}
&f^{*}_{\text{PUC}}\left(\bb, \Bbm, \bzstar\right) =\\ &\mathbb{F}^{\mathcal{L}}_{(\Bb,\Bba, \Bbz^*) \sim \distp}\left[Y\middle| \Bb=\bb, \Bba_{\indset(\bba)} = \BOne, \Bbz_{\indset(\bba)}=\bzstar_{\indset(\bba)} \right],
%\expect\left[\bff\left(\bx'\right)|\extrb\left(\bx'\right)=\bx\right].
\end{align*}
where $\Bba_{\indset(\bba)} = \BOne$ means that each element that is set to 1 in $\bba$ needs to be one in $\Bba$ as well.
\label{def:off_fairness}
\end{definition}
For a single feature ($r{=}1$), the index set can either be $\indset{=}\emptyset$ or $\indset{=}\left\{1\right\}$ and the definition corresponds to 1D-\PUC. The conditional expectation with $\Bba_{\indset(\bba)}{=}\BOne$ effectively constrains the features in $\indset$ to be available, but marginalizes over samples with or without further information.

\aaaisec{Implementing Protected User Consent}
\label{sec:off_approaches}
In this section, we derive a model-agnostic approach called \emph{PUC-inducing data augmentation} (\OFFIDA) to achieve protected user consent. %using any predictive model. 
By using theoretical analysis, we establish that \OFFIDA will result in exact protected user consent. %when using infinite sample sizes. 
Furthermore, we establish performance guarantees that provide an upper bound on the deviation between practical, finite sample-based \PUC-compliant models and their theoretical infinite sample limits.

%To implement Optional Feature Fairness (\PUC), we derive a \emph{model-agnostic} data augmentation technique that allows to approximate OFF with any type of predictive model. 
% We accordingly call this approach \textit{OFF-inducing data augmentation}.
% Using theoretical analysis, we show that estimators on the augmented data yield exact optional feature fairness under infinite sample limits. 
% Finally, we derive performance guarantees which give upper bounds on the deviation between practical finite sample based OFF-models and their theoretical infinite sample limits.

\aaaisubsec{\OFFIDA: \PUC-inducing Data Augmentation}

\newcommand{\augment}[1]{\textcolor{BrickRed}{\textbf{#1}}}
\begin{figure}
\centering\fontsize{9}{11}\selectfont
%\vspace{-1.75cm}
\setlength{\tabcolsep}{2.5pt}
\centering
\begin{tcolorbox}[boxsep=-1mm,hbox]
\adjustbox{max width=0.3\textwidth}{
\begin{tabular}{ccccccc}
& state & plan & score & costs \\
\cline{2-5}
& NSW & basic & 87\,\%  &  3k\$\\
\textcircled{+}~& \augment{NSW} & \augment{basic} & \augment{\texttt{N/A}} & \augment{3k\$} \\
%& chemistry, math  & 3.7 & $\texttt{N/A}$ & Yes \\
& NSW & basic & 92\,\%  &  5k\$\\
\textcircled{+}~& \augment{NSW} & \augment{basic} & \augment{\texttt{N/A}} & \augment{5k\$} \\
& NSW & basic & $\texttt{N/A}$  & 64k\$\\
%& biology & 3.0 & 56\,\% & No \\
% \textcircled{+}~& \augment{biology} & \augment{3.0} & \augment{\texttt{N/A}} &\augment{No} \\
%& \multicolumn{2}{c}{$\aoverbrace[l1D1r]{~~~~~~~\text{base features}~\bb~~~~~~}$} & %$\aoverbrace[@{\hfill}lDr@{\hfill}]{\text{opt. feat.}~z^{*}}$ & %$\aoverbrace[@{\hfill}lDr@{\hfill}]{\text{label}~y}$ \\
\end{tabular}}
\end{tcolorbox}\ifdefined\arxiv\vspace{2mm}\fi\vspace{-1mm}
\caption{Explaining \OFFIDA. Our data augmentation procedure expands each instance with optional information into two samples: The original instance and a synthetic sample (\textcircled{+}). 
The synthetic samples retain the base features and the labels, but the information on the optional features is dropped (fitness score $\xrightarrow{}$ \texttt{N/A}). 
The model sees samples with the same base features with a missing value and will thus base its decision only on the base features.
In this example, given the base features (``NSW'', basic) and no optional statements, the model would estimate the costs to be 24k\$, which is the dataset average conditioned on these values.}
\label{tab:resampling}
\end{figure}

% \begin{algorithm}[tb]
% \caption{PUC-SGD: SGD with Protected User Consent\label{alg:offsgd}}
% \begin{algorithmic}
% \Require Data set $\mathcal{D}$, Loss function $\mathcal{L}$, predictor $f_{\bm{\theta}}$ with parameters $\bm{\theta}$
% %\Procedure{OFF-SGD}{$\mathcal{D}, \mathcal{L}, f_{\bm{\theta}}, \bm{\theta}$}\Comment{Perform fair SGD}
% \State $\bw \gets \{\text{Distribution over}~\mathcal{D}~ \text{with}~\bw(\bx) \propto w(\bx)\}$
% \While{$r\not=0$}
% \State Sample batch $(\bx^{(1)}, y^{(1)}),\ldots, (\bx^{(k)},y^{(k)}) \sim \bw$
% \For{$j = 1,\ldots,k$} \Comment{$\bx^{(j)}{=}(\bb^{(j)}, \bba^{(j)}, \bz^{*(j)})$}
% \State $\bq \gets \text{Bernoulli}(0.5)$\Comment{iid. Bernoulli vector}
% \State $\overline{\bba}^{(j)} = \bq \odot \bba^{(j)}$
% \State $\overline{\bz}^{*(j)}_i = \left\{\bz_i^{*(j)}~\text{if}~\overline{\bba}^{(j)}_i{=}1,~\text{else}~ \texttt{N/A}\right\}i\in[r]$
% \State $\overline{\bx}^{(j)} \gets (\bb^{(j)}, \overline{\bba}^{(j)}, \overline{\bz}^{*(j)})$
% \EndFor
% \State $\bm{d}{\bm{\theta}} \gets \nabla_{\bm{\theta}}\left(\frac{1}{k}\sum_{j=1}^k\mathcal{L}\left(f_{\bm{\theta}}(\overline{\bx}^{(j)}), y^{(j)}\right)\right)$
% \State $\bm{\theta} \gets \bm{\theta} - \gamma \bm{d}{\theta}$
% \EndWhile\label{euclidendwhile}
% \State \textbf{return} $\bm{\theta}$
% %\Comment{The gcd is b}
% %\EndProcedure
% \end{algorithmic}
% \end{algorithm}

Intuitively, we want to prevent the model from making inference from a feature's missingness patterns.
The core insight is to leverage synthetic samples that make the \emph{distribution of the labels given missingness equal to the overall label distribution}. 
Thereby, we prevent the derivation of predictive information from the missingness itself (see Table \ref{tab:resampling}). 

For a single optional feature, extensively enumerating all samples as in the table is possible while for multiple features this may be intractable.
Therefore, we do not list all samples but propose a stochastic, multifeature variant of the algorithm:
\textbf{(1)} Instead of drawing samples with uniform probability from the distribution $\distp$, we use non-normalized weights $w$:
\begin{align}
   w(\bx) =  w(\bb, \bba, \bzstar) = 2^{\lvert\indset(\bba)\rvert}.
\end{align}
This step corresponds to the expansion of an instance into $2^{\lvert\indset(\bba)\rvert}$ synthetic ones; e.g., a sample with a single optional feature is assigned a weight of two (cf.\ \Cref{tab:resampling}).
Training instances are drawn with a probability proportional to these weights. 
This results in data instances with optional information being more frequently sampled. 
\textbf{(2)} We require a sample modification where optional features are randomly dropped from the samples. 
For each sampled item, we drop each available optional feature with probability $p{=}0.5$:%, thus setting
\ifdefined\arxiv
\begin{align}
%\begin{split}
\bq_i &\sim \text{Bern}(0.5), i=1,\ldots,r; ~~~~ \overline{\bba} = \bq \odot \bba;\\
\overline{\bz}^{*}_i &= \left\{\bzstar_i~\text{if}~\overline{\bba}_i{=}1,~\text{else}~ \texttt{N/A}\right\}, i=1,\ldots,r.
%\end{split}
\end{align}
\else
\begin{align}
%\begin{split}
\bq_i &\sim \text{Bern}(0.5), i=1,\ldots,r; ~~~~ \overline{\bba} = \bq \odot \bba;\\
\overline{\bz}^{*}_i &= \left\{\bzstar_i~\text{if}~\overline{\bba}_i{=}1,~\text{else}~ \texttt{N/A}\right\}, i=1,\ldots,r.
%\end{split}
\end{align}
\fi
% To understand why this is necessary, consider the case of an instance with a single optional feature: this optional information will be maintained with a probability of $0.5$ or dropped otherwise.
\textbf{(3)} We train the predictive model on the modified samples $\left(\overline{\bx},y\right){=}\left((\bb, \overline{\bba}, \overline{\bz}^{*}),y\right) \sim \overline{\distp}$ derived through this procedure.

\aaaisubsec{Theoretical Analysis}
We summarize \OFFIDA in pseudo-code in \appref{app:algorithms}{D.8}
and provide the following theorem to demonstrate that \OFFIDA leads to \PUC-compliant models.
\begin{theorem} The loss-minimal model $f\left(\bb, \Bbm, \bzstar \right) =\fexpect_{\overline{\distp}}\left[Y|\bb, \Bba=\bba, \Bbzstar=\bzstar \right] $ on the modified distribution $\overline{\distp}$ fulfills Protected User Consent with respect to $\distp$, i.e.,\label{thm:offresampling}
\begin{align*}
%\begin{split}
&\fexpect_{\overline{\distp}}\left[Y|\Bb=\bb, \Bba=\bba, \Bbzstar=\bzstar \right]=\\
&\mathbb{F}^{\mathcal{L}}_{\distp}\left[Y\middle| \Bb{=}\bb, \Bba_{\indset(\bba)}{=}\BOne,\Bbz_{\indset(\bba)}{=}\bzstar_{\indset(\bba)}\right] 
= f^{*}_{\text{\PUC}} \left(\bb, \Bbm, \bzstar\right).
\end{align*}
\end{theorem}
This result is remarkable in its generality as it enables \PUC-compliant models using standard optimization procedures by modifying the distribution of the data; 
i.e., \emph{\OFFIDA can be combined with any existing model and training pipeline}. 
%\textbf{Finite sample performance.}
Next, `we' study the theoretical convergence behavior for \OFFIDA on finite samples.
To this end, we define the \PUC-Gap as the expected squared deviation from \PUC:
%optimal
\begin{align}
&\text{\PUC-Gap}^2(f, \distp)  =\label{eqn:offgap2}\\  &\expect_{\tiny (\Bb, \Bba, \Bbzstar)\sim\distp}\big[\big( f(\Bb, \Bba, \Bbzstar) - f^{*}_{\text{PUC}}(\Bb, \Bba, \Bbzstar)\big)^2 \big]\nonumber.
\end{align}
We will restrict ourselves to $\mathcal{L}\equiv\text{MSE}$ and thus $\fexpect\equiv\expect$, and study a \emph{baseline conditional expectation estimator} $\hat{\mu}$ which averages the labels conditional on all observations with the same features $\mathbf{x}$. 
For brevity, we refer to \appref{app:finite_sample_proof}{D.7} (Eqn. 51) for a formal definition of this estimator.
Since we usually cannot compute the exact expectation from \cref{thm:offresampling}, we are interested in the number of samples required from $\overline{\distp}$ to obtain a fixed average estimation error for which we establish the following result.
\begin{theorem}[Finite Sample Convergence]
Let $\spaceX=\spaceXb \times (\spaceXz \cup \{\texttt{N/A}\})$ be finite feature space and let $\spaceY \subseteq \mathbb{R}$ be the label space. All conditional expectations $\mu(\bx) {\coloneqq}\expect_{\overline{\distp}}\left[y\middle|\bx\right]$ and the conditional variances $\sigma^2(\bx){\coloneqq} \Var_{\overline{\distp}}\left[y\middle|\bx\right]$ exist and are finite. 
Then there exists a baseline non-parametric regressor $\hat{\mu}: \spaceX \mapsto \mathbb{R}$ from a finite number of $N$ independent, identically distributed observations $\left(\overline{\bx}_i, y_i\right)_{i=1\ldots N}$ from $\overline{\distp}$ with a convergence rate of $\mathcal{O}(N^{-1})$; more specifically
\begin{align}
%\begin{split}
\text{\normalfont \PUC-Gap}^2(\hat{\mu}, \distp) = \expect_{\Bx \sim \distp}\left[\left(\hat{\mu}(\Bx)-\mu(\Bx)\right)^2\right]\nonumber\\ \leq  \frac{2^r \lvert\spaceX\rvert^2(\sigma_{\max}^2+\mu_{\max}^2)}{N} +\mathcal{O}\left(\frac{1}{N^2}\right)\nonumber,
%\end{split}
\end{align}
with $\sigma_{\max}^2{\coloneqq}\max_{\bx \in \spaceX} \sigma^2(\bx)$ and $\mu_{\max}^2{\coloneqq} \max_{\bx \in \spaceX} \mu^2(\bx)$.
\label{thm:offfinitesample}
\end{theorem}
% The expected squared deviation from the naive estimator $\hat{ \mu}$ to the optimal estimator converges at an order of $\mathcal{O}\left(\frac{1}{N}\right)$.
\newcommand{\res}[2]{#1 \small$\pm #2$}
\newcommand{\bres}[2]{\textbf{#1} \small$\pm #2$}

\begin{table}[tb]
\centering
\begin{subtable}{.48\textwidth}
\centering\fontsize{9}{11}\selectfont
\setlength{\tabcolsep}{2pt}
\begin{tabular}{cccc}
\toprule
data & base model & \Ffm & \texttt{\OFFIDA}\\
\cmidrule(lr){1-1} \cmidrule(lr){2-2} \cmidrule(lr){3-3} \cmidrule(lr){4-4}
diab.(\texttt{C})  & \res{33.84\%}{2.47}  & \textcolor{black}{\res{31.44\%}{2.19}} & \textcolor{black}{\res{34.01\%}{1.71}}\\
compas (\texttt{C})  & \res{44.47\%}{0.37}  & \textcolor{black}{\res{41.47\%}{1.09}} & \textcolor{black}{\res{44.54\%}{0.54}}\\
adult (\texttt{C}) & \res{13.37\%}{0.07}  & \textcolor{black}{\res{12.84\%}{0.28}} & \textcolor{black}{\res{13.41\%}{0.12}}\\
\midrule
 water (\texttt{C}$^{*}$) & \res{10.65\%}{1.64}  & \textcolor{black}{\res{10.00\%}{1.58}} & \textcolor{black}{\res{10.97\%}{1.21}}\\
 colic (\texttt{C}$^{*}$)  & \res{13.81\%}{0.82}  & \textcolor{black}{\res{11.34\%}{0.46}} & \textcolor{black}{\res{15.05\%}{0.68}}\\
\midrule
income (\texttt{R})& \res{109.56}{1.00} & \textcolor{black}{\res{109.11}{1.29}} & \textcolor{black}{\res{110.73}{1.29}}\\
calif. (\texttt{R}) & \res{15.79}{0.10} & \textcolor{black}{\res{15.16}{0.28}} & \textcolor{black}{\res{16.18}{0.06}}\\
insurance (\texttt{R}) & \res{283.47}{0.53} & \textcolor{black}{\res{279.78}{0.42}} & \textcolor{black}{\res{285.31}{0.39}}\\
\bottomrule
\end{tabular}
\end{subtable}
\caption{\air is violated by \ffms (Random Forests).  As expected, the \ffms always have lower losses than the base-models, indicating that \air is violated while \OFFIDA fullfils \air. We report misclassification error rates for classification models and MSE loss ($\times$ 100) for regression models.\label{tab:A}}
\end{table}

\begin{table*}
\centering\fontsize{9}{11}\selectfont
\setlength{\tabcolsep}{2pt}
\begin{tabular}{cccccccc}
\toprule
& & & &  \multicolumn{2}{c}{\Ffm} &  \multicolumn{2}{c}{\OFFIDA}\\
\cmidrule(lr){5-6}\cmidrule(lr){7-8} 
 task & data & optional & \Bfm & pred. & change & pred. & change\\
\cmidrule(lr){1-3} \cmidrule(lr){4-4} \cmidrule(lr){5-6}\cmidrule(lr){7-8} 
\texttt{C} & diab. & Glucose & 60.27\% & 45.19\% & \textcolor{black}{\res{-15.08\%}{2.01}} & 61.20\% & \textcolor{black}{\res{0.93\%}{0.93}} \\
\texttt{C} & compas & \#priors & 51.19\% & 32.86\% & \textcolor{black}{\res{-18.33\%}{0.89}} & 51.34\% & \textcolor{black}{\res{0.15\%}{0.59}} \\
\texttt{C} & adult & edu-num & 13.86\% & 11.44\% & \textcolor{black}{\res{-2.42\%}{0.07}} & 13.92\% & \textcolor{black}{\res{0.06\%}{0.05}} \\
\midrule
\texttt{C}$^{*}$ & water & oxygen. dem. & 87.10\% & 84.52\% & \textcolor{black}{\res{-2.58\%}{2.81}} & 87.42\% & \textcolor{black}{\res{0.32\%}{1.58}} \\
\texttt{C}$^{*}$ & colic & abdom. app. & 6.39\% & 1.24\% & \textcolor{black}{\res{-5.15\%}{0.92}} & 7.01\% & \textcolor{black}{\res{0.62\%}{1.64}} \\
\midrule
\texttt{R} & income & WKHP & 100.0\% & 81.2\% & \textcolor{black}{\res{-18.8\%}{0.61}} & 101.2\% & \textcolor{black}{\res{1.2\%}{0.19}}\\
\texttt{R} & calif. & m\_income & 100.0\% & 94.4\% & \textcolor{black}{\res{-5.6\%}{0.67}} & 103.8\% & \textcolor{black}{\res{3.8\%}{0.42}}\\
\texttt{R} & insurance & experience & 100.0\% & 94.8\% & \textcolor{black}{\res{-5.2\%}{0.09}} & 100.1\% & \textcolor{black}{\res{0.1\%}{0.05}}\\
\bottomrule
\end{tabular}
\caption{Measuring the average predictions for non-sharers. For classification tasks we report the positive outcomes (in \%), and for regression tasks, we report relative predictions to the \bfm (set to 100\,\%). The non-sharers face disadvantages for not providing the voluntary information and are assigned less favorable prediction outcomes by the \ffms. This discrepancy vanishes when \OFFIDA is applied.
\label{tab:B} \label{tab:fairnessgapreal}}
\end{table*}

\begin{table*}[htb]
\centering
\begin{subtable}{\textwidth}
\centering\fontsize{9}{11}\selectfont
\begin{tabular}{cccccc}
\toprule
task & data & opt. feature  & base model & \texttt{\OFFIDA} & \Ffm \\
\cmidrule(lr){1-3} \cmidrule(lr){4-4} \cmidrule(lr){5-5}\cmidrule(lr){6-6} 
\texttt{C} & diab. & Glucose & \res{29.30\%}{0.62}   & \bres{26.61\%}{0.56} & \res{23.41\%}{0.69}\\
\texttt{C} & compas & \#priors & \res{42.89\%}{0.10}   & \bres{40.85\%}{0.15} & \res{36.67\%}{0.36}\\
\texttt{C} & adult & edu-num & \res{16.05\%}{0.03} & \bres{15.94\%}{0.05}  & \res{14.86\%}{0.06} \\
\midrule
\texttt{R} & income & WKHP & \res{85.07}{0.17} & \bres{80.22}{0.15} & \res{73.25}{0.16} \\
\texttt{R} & calif. & m\_income & \res{15.62}{0.14} & \bres{14.79}{0.08} & \res{13.40}{0.03} \\
\texttt{R} & insurance & experience & \res{262.43}{0.21} &  \bres{254.35}{0.39} & \res{236.92}{0.42} \\
\bottomrule
\end{tabular}
\caption{One dimensional case, strategic withholding. Metrics: \texttt{C}: (1-Acc)$\times$100, \texttt{R}: MSE$\times$100\label{tab:strategicwithhold}}
\end{subtable}
\hfill
\begin{subtable}{\textwidth}
\centering\fontsize{9}{11}\selectfont
\setlength{\tabcolsep}{3pt}
\begin{tabular}{cccccrc}
\toprule
& & \multicolumn{4}{c}{Fair models} & \Ffm \\ 
\cmidrule(lr){3-6}\cmidrule(lr){7-7}
task & data (\# opt.) & \Bfm & \OFFIDA (f) & \OFFIDA (e) &  ($\times$) & zero-imputed\\
\cmidrule(r){1-2} \cmidrule(lr){3-3}\cmidrule(lr){4-4} \cmidrule(lr){5-6}  \cmidrule(lr){7-7}
\texttt{C} & diab. (2) & \res{29.74}{2.92} & \res{26.23}{4.42} & \bres{25.58}{3.69} & 2.2 & \res{24.16}{4.18}\\
\texttt{C} & compas (5) & \res{40.83}{0.56} & \res{37.65}{0.23} & \bres{37.21}{0.71} & 7.6 & \res{36.86}{1.20}\\
\texttt{C} & adult (5) & \res{17.98}{0.37} & \res{15.35}{0.36} & \bres{15.27}{0.25} & 7.9 & \res{15.15}{0.33}\\
\midrule
\texttt{R} & income (3) & \res{52.40}{0.92} & \bres{49.47}{1.71} & \res{51.21}{0.86} & 3.4 & \res{46.15}{1.60}\\
\texttt{R} & calif. (4) & \res{6.64}{0.79} & \res{6.83}{0.32} & \bres{6.36}{0.08} & 5.1 & \res{5.69}{0.22}\\
\texttt{R} & insurance (3) & \res{271.72}{4.14} & \bres{242.99}{4.47} & \res{260.77}{2.74} & 3.2 & \res{232.59}{2.39}\\
\bottomrule
\end{tabular}
\caption{$r$-dimensional case. Metrics: \texttt{C}: (1-Acc)$\times$100, \texttt{R}: MSE$\times$100 \label{tab:multiplefeatures}}
\end{subtable}
\caption{\PUC-compliant models leverage optional information to improve predictive performance relative to \bfms This is in line with  \Cref{corr:offnondegradation}. In the bottom table, two strategies are considered to achieve \PUC: \emph{fixed-size (f)} and \emph{exhaustive (e) \OFFIDA.} When using exhaustive \OFFIDA, the predictive performance is always better than the performance of the \bfm, and often similar to the performance of the \ffms. \label{tab:costsoffairness}\ifdefined\arxiv\newline \newline \fi}
\label{tab:exp_costs}
\end{table*}
In conjunction with \Cref{thm:offresampling}, this result provides a bound on the expected gap to perfect protected user consent that is dependent of the sample size, which decreases with a rate of $\mathcal{O}\left({N}^{-1}\right)$.
%While the bound decreases with  
Several remarks are in place: We obtain a multiplicative constant which depends on the number of optional features $r$ and the size of the feature space $\lvert\spaceX\rvert$. The square of this quantity enters the result because the number of samples available to estimate each conditional mean is not independent, as they need to sum up to $N$. For large feature spaces, however, they are almost independent and we expect the constant to scale almost linearly in $\lvert\spaceX\rvert$. 
The growth of $2^r$ is attributed to the re-sampling strategy which might assign a very low probability to certain inputs, which may only be well approximated with a high number of samples. 
As the number of optional features is typically limited in realistic use-cases it will be well outgrown by $N$.
Note that more powerful model (e.g., Tree based model + \OFFIDA) usually outperform this baseline.

\textbf{Practical considerations.} For smaller datasets, an alternative approach to random sampling is to use all possible samples to approximate the distribution $\distpo$ by a method we call ``exhaustive augmentation''. This involves enumerating all possible variations of the original samples, including any optional features, to form a larger dataset $\mathcal{D}'$. The model is then trained on this expanded dataset.

\aaaisec{Experimental Evaluation}
\label{sec:experiments}
Here, we empirically validate the effectiveness of our methods using eight real-world datasets and one synthetic dataset.
In particular, we highlight that (a) \ffms violate the \aii and make it harder for non-sharers to obtain the positive outcome, (b) \OFFIDA results in \PUC-compliant models as suggested by our theory, and that (c) the reduction in terms of model performance due to using \PUC are moderate relative to deploying a \ffm. 

\textbf{Common datasets.} 
We use eight real-world datasets commonly found in the related literature.
For classification (\texttt{C}), the Diabetes (diab) and the horse colic  dataset (colic) study the prediction of diseases, the COMPAS dataset is concerned with estimating likelihood of recidivism and UCI Adult income dataset requires to predict whether individuals have an income of over 50k\$.
The water treatment dataset (water) predicts the operational state of a facility.
We also study the regression tasks (\texttt{R}) of house price estimation in California (calif), income prediction (income), and inferring information from insurance claims (insurance) to link to our initial example.
Details about preprocessing, dataset sources and model hyperparameters are provided in \appref{sec:app_preprocessing}{F.2}.

\textbf{Availability.} The colic and the water dataset  come with inherent missing values that we use (indicated through ${*}$). For six more datasets we introduce availability dependent on a feature's value. We compute the probability of feature unavailability $\distp(A_i=0|z_i)$ by applying a sigmoid function centered at the feature mean and sample the availability $a$ from the respective conditional distribution. 
%This results in feature values equivalent to the mean having a 50\,\% chance to be unavailable.
We additionally study these datasets in the setting of strategic withholding. 
%The probability that the feature values are unavailable increases as the feature values grow above the mean. The 
%However, we invert the probability if higher values are beneficial to obtain the positive outcome.  
%%FIGSPACE1
\begin{figure}[tb]
\includegraphics[width=\columnwidth]{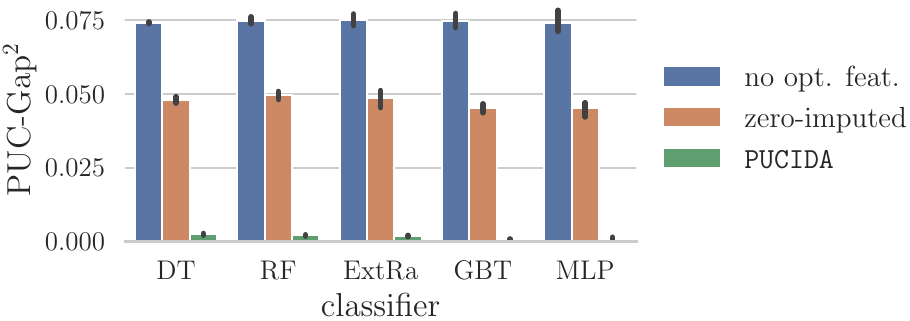}\ifdefined\arxiv\vspace{0mm}\else\vspace{-0mm}\fi
\caption{\texttt{\OFFIDA} is model-agnostic. The PUC-gaps are close to zero when applying our technique across a variety of common models on the simulated dataset.\label{fig:offconvergencemodels}\ifdefined\arxiv\newline \fi}
\end{figure}
\begin{figure}[tb]
\includegraphics[width=\columnwidth]{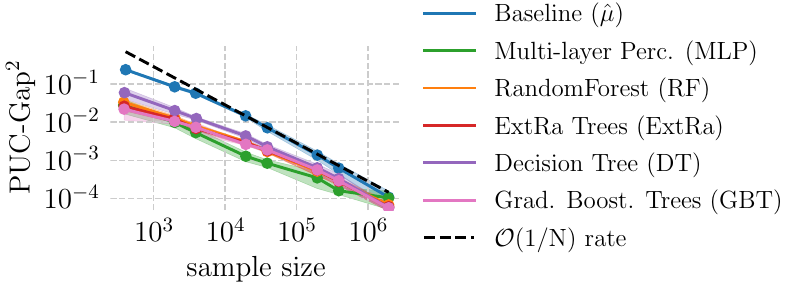}\ifdefined\arxiv\vspace{0mm}\else\vspace{-0mm}\fi
\captionof{figure}{Convergence rate of models under \OFFIDA. The estimate of \PUC converges to the true value at a rate of $\mathcal{O}(\frac{1}{N})$ for the baseline estimator $\hat{\mu}$ and other commonly used models. \label{fig:offconvergence}}
\end{figure}%
\aaaisubsec{Evaluating \OFFIDA}
\textbf{\air is violated by \ffms.}
First, we demonstrate the effect that \ffms have on \air.
We follow common practices and use zero-imputation to deal with unavailable feature values \citep{emmanuel2021survey}.
Then, we train a Random Forest model on all features of the dataset where we have introduced stochastic availability into one feature (see previous paragraph).
%, and where we (b) filled unavailable values with zeros (``zero imputed''). 
We also train a \bfm that fully drops the optional feature from the dataset. 
We consider the subset of individuals with unavailable feature values (i.e., $a{=}0$) and report the average loss and absolute prediction of the positive class for both models in \Cref{tab:fairnessgapreal}. 
We observe that the full feature models use the information contained in the missingness to obtain a lower loss. This can reduce the chance of obtaining the positive outcome from the \ffm compared to the \bfm  by significant margin of up to 18\,\% for non-sharers.
Hence, these results impressively show how the \ffm implicitly infers information from missingness and thereby violates protection requirements. This stays the same when applying established fairness constraints on the models (see \appref{sec:app_empiricalfairnessnotions}{F.1}).
In contrast, when applying \PUC using \OFFIDA this gap vanishes or is significantly reduced. We show that the same effect can be observed independently of the imputation techniques, the model class, and the model hyperparameters in \appref{sec:app_exp1}{F.3}.

\aaaisubsec{Evaluating the Theoretical Bounds}
\textbf{\texttt{\OFFIDA} guarantees Predictive Non-Degradation.}\label{sec:expcostoffairness}
Usually model performance degrades when training models with additional constraints (e.g., see \citet{corbett2017algorithmic}). 
%We follow the same setup but now make multiple numerical features optional in each dataset (see \Cref{sec:app_exp3real} for the detailed procedure). 
To measure model performance, we use the misclassification rate for classification tasks (ROC-AUC scores lead to qualitatively similar results, see \appref{sec:app_exp3real}{F.4}) and the MSE for regression tasks. 
The results in \Cref{tab:strategicwithhold} confirm that \OFFIDA (using exhaustive augmentation) improves over the \bfm, suggesting that \OFFIDA models benefits from using optional information. This is the case even under under strategic actions where users only provide data if it improves their outcome, and aligns with our theoretical result in \Cref{corr:offnondegradation}. Under non-strategic actions, the performance figures show the same characteristics (\appref{sec:app_exp3real}{F.4}).
%In \Cref{tab:offcostswithcsp} (Appendix), we show that CSP-compliant models do not improve over the \bfm as predicted by our analytical result in \Cref{lem:cspcounterexample}.
As expected, \PUC-compliant models fare moderately worse than \ffms which have no protection requirements.

We now compare two different \OFFIDA variants on multiple optional features: the first strategy ensures a fixed dataset size, i.e., the number of samples is equivalent to the original dataset size.
The second strategy, which uses exhaustive data augmentation, leads to an increased dataset size.
The factor by which the dataset size is increased is indicated by ($\times$) along with the results in \Cref{tab:multiplefeatures}. 
We observe that competitive results can often be obtained without any dataset increase; fixed-size \OFFIDA even outperforms the exhaustive variant on the larger income and the insurance dataset, whereas the exhaustive augmentation leads to a more reliable performance increase. We study the performance for sharers in Table 6 (Appendix) and find that it remains on par with the \ffm. Overall, our results demonstrate that optional information can be leveraged in a conscious way through \PUC-inducing data augmentation without suffering from prohibitive performance decrease for the decision maker and the sharers.%, while still protecting the data of non-sharers.

\textbf{Convergence of \OFFIDA.} 
Finally, we study the convergence behavior of \OFFIDA.
As a measure of approximation quality, we use the 
$\text{\PUC-Gap}^2$ defined in \Cref{eqn:offgap2}, which measures the squared deviation from perfect \PUC.
As this notion requires the knowledge of the ground truth distribution, we use a synthetic dataset for this experiment. 
The dataset consists of eight binary features (five base, three optional).
All features in this dataset are sampled independently.
Labels are induced via a logistic distribution, and availability of the optional information depends on the label.
For experiments on a second synthetic dataset with five continuous features (two base, three optional) and more details, see \appref{sec:app_exp3converge}{F.5}.

First, we observe that \OFFIDA is model agnostic, i.e., it works with a variety of state-of-the-art models leading to negligible \PUC-gaps (see \Cref{fig:offconvergencemodels}).
Second, we verify that the \PUC-Gaps converge to zero  at the rate of $\mathcal{O}(\frac{1}{N})$ as the sample size increases (\Cref{fig:offconvergence}), confirming what we derived in \Cref{thm:offfinitesample}. While common models (e.g., RandomForest, MLP) have a lower error than the baseline estimator $\hat{\mu}$ the models approach the baseline estimator with larger datasets and the gap closes at the suggested rate.

\aaaisec{Conclusion and Future Work}
% In this work, we studied machine learning predictions in the setting where users can choose to provide optional information. To respect legal requirements and user consent,
% we proposed two desiderata that balance the interests between sharers, non-sharers, and decision-makers. 
% We established Protected User Consent (\PUC) and showed that is remains possible to leverage optional information from consenting users and to improve over models that do not consider optional information at all. 
In this work, we studied machine learning predictions where users have the option to disclose optional information. 
To comply with legal regulations and respect user consent, we introduced the notion of Protected User Consent (\PUC) that strikes a balance between the interests of sharers, non-sharers, and decision-makers. 
We demonstrated that leveraging optional information from consenting users through \PUC results in superior performance compared to models that disregard the optional information entirely.

Our work gives raise to several follow-up questions. It would be interesting to study possible long-term effects of \PUC and how \PUC incentivizes improvements. 
Furthermore, we have only considered users that act entirely strategic or on privacy grounds.
Modeling heterogeneous users, who might be willing to accept a certain increase in costs in return for their privacy could be a meaningful extension. %an important extension of this work.
%Finally, motivated by the insurance example it would be interesting to include policymakers as a forth actor, and study what they can do to improve the overall status of their population.

\section*{Additional Material}
An extended version of this work including technical appendices is available online\footnote{\url{https://arxiv.org/abs/2210.13954}}.
We also publish our code as an open-source project\footnote{ \url{https://github.com/tleemann/protectedconsent}}.

\bibliography{references}
\clearpage

% UNCOMMENT FOR APPENDIX
\ifdefined\wappendix
\onecolumn
\appendix

\section{Related Work and other Fairness Notions}
% \subsection{Tables for regression setting (delete later)}
% \begin{table}[h]
%     \centering
%     \begin{tabular}{rcccc}
%     \toprule
%      data & opt. feature & imputed & base model & change \\
%     \midrule
%     heloc & AverageMInFile & 6251.09 & 6256.25 & \textcolor{BrickRed}{\res{5.16}{0.59}} \\
%     law & LSAT & 2.93 & 3.14 & \textcolor{BrickRed}{\res{0.21}{0.17}} \\
%     \bottomrule
%     \end{tabular}
%     \caption{Regression setting, Exp. 1. Score is average regression output times 100. (Multiplying with 100 makes sense for law data, but not for heloc - will fix this with the next commit)}
%     \label{tab:my_label}
% \end{table}

% \begin{table}[h]
%     \centering
%     \begin{tabular}{rcccc}
%     \toprule
%     & \multicolumn{3}{c}{Fair models} & No fairness \\
%     \cmidrule(lr){2-4}\cmidrule(lr){5-5}
%     data & base model & cond.\ parity & OFF & imputed\\
%     \cmidrule(r){1-4}\cmidrule(l){5-5}
%     heloc & \res{18.54}{0.02} & \res{18.86}{0.14} & \res{18.51}{0.02} & \res{18.52}{0.02}\\
%     law & \res{22.95}{0.01} & \res{22.96}{0.02} & \res{22.95}{0.01} & \res{22.95}{0.01}\\
%     \bottomrule
%     \end{tabular}
%     \caption{Regression setting, Exp. 2. The Score is 100 times MSE.}
%     \label{tab:my_label}
% \end{table}

\subsection{Additional Related Work}
\label{sec:app_relatedwork}
\textbf{Estimation of causal effects in the presence of missing data.} The works by \citet{mohan2018estimation, mohan2021graphical} introduce graphical models for incomplete data and study the consistent estimation of causal effects amidst missing values. Our work differs as we are not concerned with estimating true causal effects but focus on building a definition of fairness in the presence of optional data.
% that also covers the non-linear case.

\textbf{Implementing Fairness in ML Systems.} There are different strategies to implement fairness and mitigate bias in practical decision-making systems.
This can be done by adding additional constraints to the optimization problem (e.g., \citet{zafar2019fairness}).
To solve the such an optimization problem, one can employ the reductions approach \cite{agarwal2018reductions}, where the fairness constraint is reduced to a series of classification problems with different costs assigned to each sample. 
Furthermore, another line of work consists of preprocessing approaches to obtain models that are compliant with classical fairness notions. 
These work through sample selection \cite{roh2021sample, abernethy22activesampling} and reweighting approaches \cite{chai2022fairness, li2019repair, li2022achieving} or through resampling of the sensitive attribute \cite{romano2020achieving}.
While these approaches can help fulfill common fairness notions, they cannot easily be applied to obtain \PUC.

\textbf{Trade-offs between Privacy and Fairness.} Possible trade-offs between classical notions of privacy such as Differential Privacy (DP, \citealp{dwork2006calibrating}) have been previously studied \cite{bagdasaryan2019differential, ganev2022robin, amiri2022impact}, showing that imposing DP may lead to disparate outcomes across sensitive groups or reinforce existing biases. \citet{suriyakumar2021chasing} recently found that imposing privacy constraints can lead to an undue influence of majoritiy groups over miniories, thus possibly impacting fairness. Although we are considering personal data in this work, this paper differs from classical privacy literature because we are not concerned with data leakage. Instead, we strive to give users a better choice of which data to provide in the first place. 
%\textbf{Unsupervised Fairness}. Furthermore, there is a stream of works around the question of how fairness can be achieved without group labels.
\subsection{Common fairness notions are not applicable}
\label{sec:presentdefinitions}
As many definitions of fair outcomes between an advantaged and a disadvantaged group exist, we investigate whether existing definitions can readily be applied or easily adapted to the optional feature setting considered in this work. 
In other words, here we study whether existing fairness notions comply with our desiderata of Optimality and \aii.
In the conventional fairness literature, the impact of a sensitive attribute on the prediction is restricted.
However, in the optional feature setting the point of departure is different since the optional feature may contain discriminative information that we explicitly want to use in come cases (recall that the sharers would like to obtain the most accurate prediction given their information).
If not stated otherwise, we consider the availability feature $A$ (see \Cref{tab:datasample}) to be the sensitive attribute.
We denote the predicted label by $\hat{Y}$ and discuss binary labels $Y\in\left\{0,1\right\}$ as in most of the original definitions.

\textbf{Fairness through Unawareness.}
%\cite{grgic2016case,kusner2017counterfactual}.} 
This notion demands that the availability indicator $A$ is not used as an explicit input in the decision-making process. Removing explicit information on the availability can be done easily by dropping the feature $A$. 
This makes ``Fairness through Unawareness''
very easy to implement. 
However, the group information is still implicitly encoded in the optional feature through the value $\texttt{N/A}$ (see \cref{tab:datasample}).
%is typically replaced by a placeholder (e.g., with the mean or the median) imputed optional feature through the value $\texttt{N/A}$.
A sufficiently complex classifier can infer this group information and include it into its decision-making. %(cf.\ experiments in \Cref{sec:expunfairtreat}).
Therefore, this fairness notion cannot be applied in the optional feature setting as it violates \air.
 
\textbf{Predictive Parity.} %\cite{chouldechova2017fair}.} 
This notion of fairness constrains the False Discovery Rates to be equal across groups, i.e.,
$P(Y=0|\hat{Y}=1,A=0) = P(Y=0|\hat{Y}=1,A=1)$.
We argue that this definition and other error rate-based ones will not work in our setup because they bound performance an thus violate Optimality. It is desired by the sharers and the decision maker that the predictions will be more accurate when the feature $z$ is present ($A=1$) because the information in $z$ should explicitly be used in the decision-making process if users decide to share their data on the optional features.
%Constraining this threshold for the case with the additional information will result in a classifier where the $P(Y=0|\hat{Y}=1,A=1)$ is artificially low, resulting in a higher $P(Y=1|\hat{Y}=0,A=1)$. This means persons that have provided additional information and are wrongfully rejected, which is not desirable.
One can make an analogous argument for other error-rate based notions such as equalized odds and equal opportunity.

\textbf{Equalized Odds and Equal Opportunity \cite{hardt2016equality}.} Equalized odds requires the predicted label $\hat{Y}$ and the the protected attribute $A$ to be conditionally independent given the true label $Y$. 
Formally, this means $P(\hat{Y}| A,  Y) = P(\hat{Y}|Y)$ for all values of $Y, A, \hat{Y}$. 
This effectively constrains the true and false positive rates to be equal across groups. 
However, by the desideratum of Optimality, it is required to use class-discriminative information in the optional feature, which will necessarily lead to lower missclassification rates for subjects with $A{=}1$.
Another fairness notion is Equal Opportunity which is a relaxation of Equalized Odds that only demands $P(\hat{Y}| A{=}1, Y{=}1) = P(\hat{Y} | A{=}0, Y{=}1)$, thus constraining the true positive rates across groups.
To fulfill this notion, for $A{=}1$, the true positive rate would have to be kept artificially low to match that of the case $A{=}0$, with less information.
This would thus result in a lower $P(\hat{Y}{=}1 | A{=}0, Y{=}1)$, than could be achieved otherwise. 
Let $Y{=}1$ be the desirable outcome (e.g., being assigned a low insurance quote); this means that less subjects are rewarded with the justified positive outcome. This is incompatible with our desideratum of Optimality.

\textbf{Statistical Parity \cite{dwork2012fairness, kusner2017counterfactual}.} This definition is satisfied by a classifier if subjects in both protected and non-protected groups have an equal probability of getting a positive classification outcome: $P(\hat{Y}{=}1|A{=}0) = P(\hat{Y}{=}1|A{=}1)$. If the set of people providing additional information has more favorable base features in general, this definition may lead to different thresholds where people that choose to provide information are getting a lower score to achieve parity. This definition would even forbid using this base features' full distinctive power, because one has to equalize over both missingness classes, thus contradicting Optimality. %Furthermore, statistical parity is known to be notoriously unfair on an individual level \cite{dwork2012fairness, zemel2013learning}.

\textbf{Individual fairness \cite{dwork2012fairness}.} Fairness definitions in this category use a distance metric $m$ to define similarities $m(\bx_i, \bx_j)$ between individuals $x_i$ and $x_j$. Considering the application in mind, the sensitive attributes should not play a role in determining the distance. The classifier output distributions for $f(\bx_j)$ and $f(\bx_i)$ that are compared by some divergence $D$ should not differ more than the distance between these individuals, i.e., $D(f(\bx_j), f(\bx_i)) \leq m(\bx_i, \bx_j)$ \cite{dwork2012fairness}. In the considered setting, following the proposition by \citet{verma2018fairness}, we could define the distance to be $0$, if individuals have the same base features $\bb$. This would effectively constrain the classification outcome to be identical independently of the optional feature specified, effectively prohibiting its use. Even when defining other distance metrics, the classification outcome will still be constrained to a certain range, again contradicting our desideratum of Optimality.

\textbf{(Conditional) Statistical Parity.} Statistical parity (SP) is known to be notoriously unfair on an individual level \cite{dwork2012fairness}.
Therefore, \citet{corbett2017algorithmic} define the notion of conditional statistical parity (CSP), which is an extension of SP, where some attributes are allowed to affect the decision. If we allow all base features $\bb$, the resulting definition expressed in expectations would be $P[\hat{Y} |\bm{B}=\bb, A=0] = P[\hat{Y} |\bm{B}=\bb, A=1]$.
% This definition would require individuals that do not provide additional information to be treated like the average of the population that provided these features. 
% While this definition comes close to our goals of Incentivization and Non-Penelization we show that when Incentivization is fulfilled, the performance of a model with this constraint can be worse than that of a model trained only on the base features even under idealized conditions.
While this definition can be compliant with \air, we show that  CSP-compliant models cannot meet the desire of the sharers for most accurate predictions. They cannot assign most accurate predictions to sharers or would encounter prohibitively high costs due to the CSP constraint if they did so. Indeed, they can be worse than the performance of a \bfm, even when we assign the sharers the most accurate predictions.
This is the case even under idealized conditions (i.e., known expectations) and when Incentivization is perfectly fulfilled. \PUC-models do not suffer from this limitations and can assign sharers most accurate predictions while always matching the performance of the \bfm.

\begin{lemma}[CSP-compliant models can degrade model performance over \bfm]
\label{lem:cspcounterexample}
There exists a density $\distp$ for which a CSP-compliant model $f^{*}_{\text{CSP}}: \spaceX \rightarrow \left[0,1\right]$ which assigns the most accurate predictions to the sharers, i.e., $f^{*}_{\text{CSP}\mid_{a=1}}(\bb, a, z^*) = \mathbb{F}_{\mathcal{L}}\left[Y|\bb, A=1, z^{*}\right]$ leads to higher expected losses (for MSE and BCE losses) than an optimal \bfm $g^*_{\text{BCE}}:$ %(\bb) = P\left[Y=1|\bb\right]$, i.e.,
\begin{align}
\expect_{\distp}[\mathcal{L}\left(g^*_{\text{BCE}}(\Bb), Y\right)] < \expect_{\distp}[\mathcal{L}\left(f^{*}_{\text{CSP}}(\Bb, A, Z^*), Y\right)].
\end{align}
%even when we suppose perfect estimators. 
\end{lemma}
A proof is provided in \Cref{sec:counterexamplecsp}.
There, we give a density $\distp$ that serves as such a counterexample. 
We argue that this fairness notion is incompatible with the desire of the sharers for accurate predictions and the decision makers desire for low overall costs.
Thus, we have established that common fairness definitions fail to conform to our desiderata of \air or result in models with unreasonable performance characteristics.
%This findings is illustrated in Figures \ref{fig:vanilla_classifier} - \ref{fig:equalizedoss}.

\section{Intuition and Additional Examples}
In this section, we provide a simple example to show the problem of possible unfairness and provide more intuition for our notion of \puc.

\subsection{Standard losses may lead to unfair treatment}
We revisit the example of college admission, to show how imputation leads to possibly unfair treatment. Suppose we are given the samples $\left\{(\bx^i, y^i)\right\}_{i=1..N}$ with $N=5$ from \cref{tab:datasample}. Using the standard Mean-Squared Error (MSE) loss, we solve the following empirical risk minimization problem: 
\begin{align*}
f^* = \argmin_{f\in \mathcal{F}} \sum_{i=1}^N (f(\bx^i)-y^i)^2,
\end{align*}
with a sufficiently expressive function class $\mathcal{F}$. For samples in the example data set, this will yield the outcome $f^*(\bx) = \frac{1}{|\{\bx^i = \bx\}|}\sum_{\{\bx^i = \bx\} } y_i$, the empirical mean. Consider the samples $\bx^1$ and $\bx^4$ in \cref{tab:datasample}. Candidate 1 chose to share an additional feature, while candidate 4 did not. Although they have the same base features, their classification will be different at test time (as the data in the training set) with $f^*(\bx^1)=3k\$$ and $f^*(\bx^4)=64k\$$.

We argue that in the case of candidate 4, the availability information was implicitly used to compute the score and resulted in a lower outcome. If only the base features had been available, i.e., $f^*$ would have been trained on the data set $\left\{(\bb^i, y^i)\right\}_{i=1..k}$, the model outcome would be $f^*(\bb) = \frac{1}{|\{\bb^i = \bb \}|}\sum_{\{\bb^i = \bb \} } y^i$ with $f^*(\bb^4) = 24k\$$ which is the dataset average for all customers from NSW with a basic coverage plan. In this work, we argue that the unavailability of certain features itself should not be used in the determination of the model outcome when no additional information is available.

\subsection{Example: Missing at random (MAR) data.} For missing at random data \cite{rubin1976inference}, the likelihood of unavailability can be entirely accounted for by the observed base features $\bb$ and is not affected by the partially observed $z$ and the label $y$. Formally, for a single optional feature with random availability $A$, $\distp(A=0|\bb) = \distp(A=0|\bb, z, y)$ for every $z \in \spaceXz, y \in \spaceY$. Therefore,
\begin{align}
\distp\left(y|\bb,A=0\right)
&=\frac{\distp(y,\bb, A=0)}{\distp(\bb, A=0)}=\frac{\distp(y)\distp(\bb|y)\distp(A=0|\bb,y)}{\sum_{y'} \distp(y')\distp(\bb|y')\distp(A=0|\bb,y')}\\
&=\frac{\distp(y)\distp(\bb|y)\distp(A=0|\bb)}{\sum_{y'} \distp(y')\distp(\bb|y')\distp(A=0|\bb)}=\frac{\distp(y)\distp(\bb|y)\distp(A=0|\bb)}{\distp(A=0|\bb)\sum_{y'} \distp(y')\distp(\bb|y')}\\
&= \frac{\distp(y)\distp(\bb|y)}{\sum_{y'} \distp(y')\distp(\bb|y') }= \distp(y|\bb).
\end{align}
Therefore, we also have $\expect\left[Y|B=\bb, A=0\right] = \expect_{\distp}\left[Y\middle| B=\bb\right]$, indicating that the missingness does not affect the expected value of the label (or that of any other functional of $p(y|\bb)$)  over the entire data distribution. Therefore, a perfect discriminative model with $f(\bx)=\fexpect_{\distp}\left[Y|\bx\right]$ will fulfill \cref{def:off1d}, our definition of \PUC, right away.

\subsection{A probabilistic derivation of Non-Penalization}
\label{sec:app_probderivationnonpen}
First, we require that the predictor does not use the information as an explicit input, i.e., the predictor should behave as if it only used base features $\bb$ via some function $g: \spaceXb \rightarrow \spaceY$: %i.e.,
\begin{align}
f_{\mid_{a=0}}(\bb, a, z^*) = g(\bb).
\label{eq:basic_constraint}
\end{align} 
For (b), although $a{=}0$ is not an explicit input to $g$, a sufficiently complex function may still be implicitly adapting to the group $a{=}0$ and thus incorporate information that the user did not give their consent to.
Therefore, on its own, the constraint in eqn.\ \eqref{eq:basic_constraint} is insufficient to enforce \air, and we need to formally define which predictors $g$ are not specific to the information provided by the group of non-consenting users.
To make matters more concrete, we first consider the case of binary classification with $\spaceY \in \left\{0,1\right\}$ from a probabilistic perspective and suppose $g(\bb)$ returns a numerical probability socre in $[0,1]$. We let $\distp_g(\hat{Y}|\bb)$ denote stochastic predictions $\hat{Y}$ defined through $g$ where $\distp_g(\hat{Y}=1|\bb) \coloneqq g(\bb)$. We would like to constrain the information contained in the predictions $G(\bb)$ to not use any additional information that the users did not actively consent to. To come up with a suitable constraint, we consider two simple others predictors, where one should be allowed and the other should be ruled out.

% In the probabilistic two-class case where we obtain probabilistic predictions $G(\bb) \sim \distp_g(\hat{Y}|\bb)$ and have predictions from the marginal likelihood independent of $A$ is given by $\hat{Y}(\bb) \sim \distp(Y|\bb)$, This requires
% \begin{align}
%    \text{MI}(G(\Bb);Y|A=0) \leq \text{MI}(\hat{Y}(\Bb);Y|A=0).
% \end{align}
% where MI denotes mutual information. 

%This notion can be generalized to non-classification problems with an arbitrary loss function $\mathcal{L}$ where it requires %Generalized to arbitrary loss functions $\mathcal{L}$ (\Cref{sec:generalizednonpen})  this requires:
% , i.e., it cannot be more informative about the label for this group than an optimal predictor $f_{\text{BCE}}^{*}(\bb)=\expect\left[Y|\bb\right]$ trained on the entire dataset using the base features, i.e.,}
% \[
% \text{MI}\left[Y,g(\Bb)\middle|A=0\right] \leq \text{MI}\left[Y,f_{\text{BCE}}^{*} (\Bb)\middle|A=0\right].
% \]

%\left(Y=1|\bb\right){=}g(\bb)$. %, i.e., the predictor needs to be further away from the non-compliant predictor than the global average. 
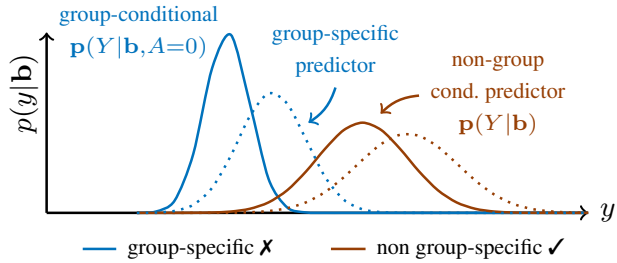
\begin{figure}[ht]
\centering
\newcommand{\lgendline}[0]{-0.4}
\newcommand{\lgendlengthoff}[0]{0.2}
\scalebox{1.2}{
\begin{tikzpicture}
  \draw[->, thick] (-3, 0) -- (3, 0) node[right] {\small$y$};
  \draw[->, thick] (-3, 0) -- (-3, 2) node[left, rotate=90, yshift=0.25cm, xshift=-0.1cm] {\small$p(y|\bb)$};
  \draw[scale=0.25, domain=-8:12, smooth, variable=\x, RoyalBlue, thick] plot ({\x}, {8*2.7^(-0.5*(\x+4)*(\x+4))});
   \draw[scale=0.25, domain=-8:12, smooth, variable=\x, RawSienna, thick] plot ({\x}, {4*2.7^(-0.125*(\x-2)*(\x-2)))});
  \draw[scale=0.25, domain=-8:12, smooth, variable=\x, RawSienna, thick, dotted] plot ({\x}, {3.5*2.7^(-0.10*(\x-4)*(\x-4)))});
   \draw[scale=0.25, domain=-8:12, smooth, variable=\x, RoyalBlue, thick, dotted] plot ({\x}, {5.33*2.7^(-0.222*(\x+2)*(\x+2)))});
   \node at (2.0,1.3) (unspec){\parbox{1.5cm}{\centering\scriptsize{\textcolor{RawSienna}{non-group cond. predictor\\$\distp(Y|\bb)$}}}};
  \node at (-1.98,2.0) {\parbox{2cm}{\centering\scriptsize\textcolor{RoyalBlue}{group-conditional\\$\distp(Y|\bb{,}A{=}0)$}}};
    \node at (0.2,1.8) (part){\parbox{2cm}{\centering\scriptsize\textcolor{RoyalBlue}{group-specific predictor}}};
\draw[->, thick, RoyalBlue] (part.south) to [bend left=20]  (-0.1, 1);
\draw[->, thick, RawSienna] (unspec.west) to [bend right=20]  (0.7, 1.1);

\node at (1.7,\lgendline) (ok){\scriptsize non group-specific \cmark};

\node at (-1.3, \lgendline) (notok){\scriptsize group-specific \xmark};

\draw[thick, RoyalBlue] (notok.west) to (-2.4-\lgendlengthoff, \lgendline);
\draw[thick, RawSienna] (ok.west) to (0.35-\lgendlengthoff, \lgendline);
 
\end{tikzpicture}}
\caption{In our definition, predictors are called group-specific (blue) if they are closer to the group conditional distribution than to the overall label distribution (pink). Our requirement forbids the use of such group-specific predictors for the group of users with no additional information.\label{fig:groupspecific}}
\end{figure}

There are two canonical examples: a probabilistic estimator that is certainly not adapted to a specific group would be the one matching the ground truth overall conditional probability $\distp\left(Y|\bb\right)$. 
On the other extreme, the predictor $\distq$ equivalent to $\distq\left(Y|\bb\right) \coloneqq \distp\left(Y|\bb, A{=}0\right)$ is fully leveraging the protected information and would thus be non-compliant. 
Generalizing this insight, we rule out all probabilistic predictors that are closer to the most non-compliant predictor than the overall conditional predictor, which we consider valid. These forbidden, group-specific predictors are visualized in \Cref{fig:groupspecific}.

To this end, a suitable distance metric is required between the predictive distributions. 
A common choice is the Kullback-Leibler divergence $\mathcal{D}_{\text{KL}}$, which results in the following requirement for a predictor $g$:
\begin{align}
\mathcal{D}_{\text{KL}}\left(\distp(Y|\Bb,A{=}0)\middle|\middle|\distp_g(Y|\Bb)\right)  \geq \mathcal{D}_{\text{KL}}\left(\distp(Y|\Bb,A{=}0)\middle|\middle|\distp(Y|\Bb)\right).
\label{eq:kl_condition}
\end{align}
%While this formulation is suitable for binary classification, we identify similar constraints for other loss functions in \Cref{sec:generalizednonpen}.
%In \Cref{sec:generalizednonpen}, we show that
The condition from eqn.\ \eqref{eq:kl_condition} can be equivalently stated in terms of expected loss for binary classification and regression problems; 
%under a homoscedasticity assumption. 
i.e., the above condition allows to derive a generalized principle of \air (see \Cref{sec:generalizednonpen} for the proof).

\section{Effects of user choices in \PUC-compliant models}
In this section, we provide a brief discussion on the effect a user's choice to provide or not provide optional information has for the decision maker and for the affected end users.

\textbf{Balancing the interests of consenting users, non-consenting users and decision makers.}
From the user's viewpoint, we would like to outline that a user's choice to provide optional information or not may depend on different factors: 
\begin{itemize}
\item[(a)] \textbf{Relevance of information:} How relevant does the individual deem the information that is asked for. The user may (correctly or incorrectly) deem certain information irrelevant for the decision and therefore they may not be willing to provide information on optional features. 
\item[(b)] \textbf{Sensitivity of information:} How concerned is the individual about the optional information being unintentionally leaked or intentionally passed on to a third party. 
\item[(c)] \textbf{Prior beliefs and expected outcomes:} The user's mental models of the decision system and the role their information plays in the system could be essential as a user may be more inclined to provide information which they deem beneficial for their prediction.
\end{itemize}
In summary, the \emph{utility} of an individual for providing data is composed of several factors, including sensitivity, perceived relevance and anticipated outcomes.

From the perspective of the decision maker, \PUC models become increasingly more accurate with more data being voluntarily provided. 
Therefore, from the perspective of the decision maker, it is important that users can also benefit from providing optional information.
This desiderata is captured by optimality requirement, which allows the decision maker to make the most of all voluntarily provided information and allows the users to obtain more accurate decisions when additional information is provided.

\textbf{Can less information lead to more favorable predictive outcomes for users?}
A user's predictive outcome depends on which information was provided by the user, and the predictive outcomes do not need to be monotonic in the number of optional features being provided; i.e., \emph{providing more information on optional features does not necessarily lead to a better outcome for the user}.

To see why this behavior is necessary and desired, we consider the two extreme cases on either side of the spectrum, where (a) optional information may not impact the prediction outcome, and where (b) predictions can only get worse when providing strictly less features.
In case (a), where no changes to the predictions occur, the setup becomes trivial and results in the \bfm. 
If this is the goal, then the collection of any additional information is useless for the decision and one should refrain from collecting these data, directly following the principle of Data Minimization. 
In case (b), where users can only get worse predictions with less information, a machine learning model would always have to treat users, who did not provide information, as if the worst possible value of a feature was provided. 
This is to make sure that the prediction outcomes of consenting users remains higher than the outcomes of non-consenting users with identical features. 
In the real-world setting of college admission we considered throughout the main text, this would lead to severe penalization of users who did not share optional test score results. 
This could de-facto rule them out entirely.
\emph{We argue that this behavior is not desired as it implicitly forces users to provide their data}.
If this is desired, then the decision maker should make this choice explicit and the feature should then be mandatory. 
To conclude, for the setting of optional features with a decision maker who is interested in providing users a real choice, any sensible notion of fairness must allow for differences in the outcomes depending on the provided information.

\section{Proofs}
\label{sec:app_proofs}
\subsection{Counterexample: Conditional statistical parity can be inferior to the base model}
\label{sec:counterexamplecsp}

Expressed in terms of expectations, the notion of conditional statistical parity $\expect[\hat{Y} |\bm{B}=\bb, A=0] = \expect[\hat{Y} |\bm{B}=\bb, A=1]$ requires the prediction averages conditioned on $\bb$ to be equal among groups that provided the optional features and those that did not. We now consider a non-probabilistic prediction function $\hat{Y} = f(\bb, a, z^*)$. Plugging in the functional form would result in the following definition: $f_{\mid_{\bba=0}}(\bb, \bba=0, z^*)=\expect_{Z^* \sim \distp(Z^*|\bb,\bba=1)}\left[f(\bb, \bba=1, Z^*)\right], \forall \bb$. In the case $a=0$, $z^*$ is constrained to be $\texttt{N/A}$ so we can ignore its value. The subscript is used to indicate the restriction of $f$ on the set of points with $\bba{=}0$. This definition constrains the output $f_{\mid_{m=0}}$, when no additional features provided, to match the average output of the individuals that provided features. 

We follow the requirement most accurate predictions by the sharers which requires $f_{\mid_{a=1}}$ to be the best approximation of $\expect\left[Y|\bm{B}, A=1, Z^*\right]$. Thus, we would have to set $f_{\mid_{a=0}}$ to be $f_{\mid_{a=0}}(\bb, a, z^*)= \expect_{Z^* \sim \distp(z^*|\bb,a=1)}\left[ \expect\left[Y|\bm{B}=\bb, A=1, Z^*\right]\right]=\expect\left[ Y | B=\bb, A=1 \right]$ when marginalizing over $Z^*$. Overall, this derivation results in a function $f_{csp}$ of the following form:
\begin{align}
f_{\text{csp}}(\bb,a,z^*)= \left\{
\begin{matrix}
\expect[Y|\bb, A=1] & \text{if}~a=0 \\
\expect[Y|\bb, A=1, Z^*=z^*], & \text{if}~a=1 
\end{matrix}\right. .
\end{align}
In this section we present a simple example to show that this function $f_{\text{csp}}$ derived from notion of conditional statistical parity may lead to an increased Mean-Squared-Error (MSE) and Binary Cross Entropy (BCE) loss compared to the base model (not using the optional feature) even when the estimators of the conditional means are perfect.
Note that in a binary space $\spaceY=\left\{0,1\right\}$ for both losses, predicting the conditional expectation is optimal.
% \begin{definition}[Natural Loss Function]
%     We call a loss function $l:\mathbb{R} \times \mathbb{R} \rightarrow \mathbb{R}$ natural, if there exist $a, b \in \mathbb{R}$, such that $l(a,a) < l(b,a)$ and $l(b,b)<l(a,b)$ and if $\argmin_{z\in \mathbb{R}} l(z,b) = b$.
% \end{definition}

%First, we would like to remind the reader that the expected MSE-Loss is minimized by models $f^*$ exactly predicting the conditional expectation, i.e.,
%\[
%f^*(\bx) = \argmin_{f(\bx)} \expect_{Y \sim \distp(y|\bx)} \left[(Y-f(\bx))^2 \right]  = \expectw{Y \sim \distp(y|\bx)}{Y|\bx}
%\]

For the example, we take any value $\bb$ and suppose $p(y|\bb, A, Z)$ depends on $A$ but that $Z$ is useless and does not contribute any new information, i.e. $\forall z \in \spaceXz, A \in \left\{0,1\right\}: p(y|\bb, A, z) = p(y| \bb, A)$.
Furthermore, we set the outcome to be deterministic of $A$:
\begin{align}
\mathbb{E}\left[Y|\bb, A=0 \right]&=0\\
\mathbb{E}\left[Y|\bb, A=1 \right]&=1\\
\mathbb{E}\left[Y|\bb\right]&=\distp(A=1|\bb)\mathbb{E}\left[Y|\bb, A=0 \right] + \distp(A=0|\bb)\mathbb{E}\left[Y|\bb, A=1 \right] = \distp(A=1|\bb) \coloneqq \alpha.
\end{align}
Let $\distp(A=1|\bb) = \alpha$ be in the range $0 < \alpha < 1$.
The optimal \bfm $g^{*}$ would predict:
\begin{align}
g^{*}(\bb) = \alpha,
\end{align} 
whereas the model based on CSP is given by:
\begin{align}
f_{csp}(\bb) = 1,
\end{align}
independently of the realization of $A$ (because it is not allowed to use this information). The expected MSE Loss is given by:
\begin{align}
L_{base, MSE}&=\distp(A=0|\bb)(\alpha-0)^2+\distp\left(A=1|\bb\right)(\alpha-1)^2\\
&=(1-\alpha) \alpha^2 + \alpha(1-\alpha)^2
=(1-\alpha)\alpha(\alpha+1-\alpha)=(1-\alpha)\alpha,
\end{align}
\begin{align}
L_{base, BCE}&=-\distp(A=0|\bb)\log (1-\alpha) -\distp(A=1|\bb) \log \alpha < \infty.
\end{align}
If the notion derived from Conditional statistical parity is used, we would use $f_{csp}(\bb) = \expect\left[Y|\bb, A=1\right]=1$ to predict in both cases and obtain:
\begin{align}
L_{csp, MSE}=\distp(A=0|\bb)(0-1)^2 + \distp(A=1|\bb)(1-1)^2=p(A=0|\bb) = 1-\alpha,
\end{align}
\begin{align}
L_{csp, BCE}=- \distp(A=0|\bb) \log (1-1) - \distp(A=1|\bb)\log 1 =p(A=0|\bb) = \infty.
\end{align}
For the BCE, we already see that the loss is unbounded in the case of CSP. One can construct the same example with non-infinite losses by adding a slight probability of the other outcome, i.e., setting  $\mathbb{E}\left[Y|\bb, A=1 \right]=1-\epsilon$ with  some small $\epsilon>0$ and obtain an analogous result.

For the MSE, in every case with $\alpha = \distp (A=1|\bb)<1$ this results in:
\begin{align}
L_{csp}= 1-\alpha > (1-\alpha)\alpha = L_{base}.
\end{align}
We have now shown that for an arbitrary $\bb$, the loss can be higher that of the base feature model. We can complete the example to the overall loss over a distributions of $\bb$'s by supposing $\distp(\Bb=\bb)=1$, which would however be a degenerate distribution. As a broader alternative, one can assume the above  for a set of $\bb \in \mathcal{B}$ and suppose any probability distribution with support in $\mathcal{B}$, i.e., $\distp(\Bb \notin \mathcal{B})=0$.

\subsection{Proof: The notion of \puc is optimal in the set of predictors conforming to the two desiderata}

We can consider both predictors (for the case with and without optional features) independently.
On the one hand, the notion of \air demands that the base predictor $f_{\mid_{a=0}}(\bb) = g(\bb)$ should not outperform the optimal base predictor $f^*_\mathcal{L}$ trained on the full data set,  
\begin{align}\expect_{\distp}\left[\mathcal{L}\left(f^*_\mathcal{L}(\Bb), Y\right)\middle| A=0 \right] \leq \expect_{\distp}\left[\mathcal{L}\left(g(\Bb), Y\right)\middle|A=0\right]. \label{eqn:incenctivationproofoff}
\end{align}
This directly provides us with one predictor $g(\bb)$, that is optimal in terms of loss for these individual namely, $g\equiv f^*_{\mathcal{L}}$ where $f^*_{\mathcal{L}}(\bb)=\mathbb{F}_{\mathcal{L}}\left[Y|\bb\right]$

On the other hand, for the group of individuals with optional information, we face no constraints and thus use the best predictor possible, i.e., 
\begin{align}
    f_{\mid_{a=1}}(\bb, a, z^*) = \mathbb{F}_{\mathcal{L}}\left[Y|\bb, A=1, z^{*}\right]=\argmin_{f(\bb,a,z^{*})} \expectw{\distp(Y|\bb,A=1,z^*)}{\mathcal{L}(f(\bb,a,z^{*}),Y)}.
\end{align}

Together, this results in the given definition of \PUC.
\hfill$\square$

\subsection{Proof: 1D-PUC obeys Predictive Non-Degradation}
\label{sec:app_predictivenondegr}

For the case of optional features ($A=1$), we have:
\begin{align}
    f^{\text{\PUC}}_{\mid_{a=1}}(\bb, a, z^*) = \mathbb{F}_{\mathcal{L}}\left[Y|\bb, A=1, z^{*}\right]=\argmin_{f(\bb,a,z^{*})} \expectw{\distp(Y|\bb,A=1,z^*)}{\mathcal{L}(f(\bb,a,z^{*}),Y)}.
\end{align}
As is is the optimal predictor, its loss on these samples is smaller than that of any model, including the optimal model on the base features. Therefore, for each $\bb, z^*$, we have:
\begin{align}
\expectw{\distp(Y|\bb,A=1,z^*)}{\mathcal{L}(f^{\text{\PUC}}(\bb,a{=}1,z^{*}),Y)} \leq  \expect_{\distp(Y|\bb,A=1,z^*)}\left[\mathcal{L}\left(f^*_\mathcal{L}(\bb), Y\right) \right].
\end{align}
Averaging over the entire class of samples with $A=1$, we obtain:
\begin{align}
\expect_{\distp}\left[\mathcal{L}\left(f^{\text{\PUC}}(\Bb, A, Z^*), Y\right)\middle|A=1\right] \leq \expect_{\distp}\left[\mathcal{L}\left(f^*_\mathcal{L}(\Bb), Y\right)\middle| A=1 \right].
\end{align}

%Note that it is easily possible to construct destributions where \Cref{eqn:incenctivationproofoff} holds with equivalence. For instance, this is the case, when missingness is independent of the label and the base features and the optional information also only contains random information.

On the other hand, the definition of \PUC demands that the predictor in case $A=0$ is equivalent to the optimal predictor on the base features. Thus they have equal loss and:
\begin{align}
   \expect_{\distp}\left[\mathcal{L}\left(f^{\text{\PUC}}(\Bb, A, Z^*), Y\right)\middle|A=0\right]=\expect_{\distp}\left[\mathcal{L}\left(f^*_\mathcal{L}(\Bb), Y\right)\middle| A=0 \right].
\end{align}
In total, we have 
\begin{align}
\mathcal{L}^{\text{\PUC}} &= \expect_{\distp}\left[\mathcal{L}\left(f^{\text{\PUC}}(\Bb, A, Z^*), Y\right)\middle|A=0\right]\distp(A{=}0) +\expect_{\distp}\left[\mathcal{L}\left(f^{\text{\PUC}}(\Bb, A, Z^*), Y\right)\middle|A=1\right] \distp(A{=}1)\\
&\leq \expect_{\distp}\left[\mathcal{L}\left(f^*_\mathcal{L}(\Bb), Y\right)\middle| A=0 \right] \distp(A{=}0) +\expect_{\distp}\left[\mathcal{L}\left(f^*_\mathcal{L}(\Bb), Y\right)\middle| A=1 \right] \distp(A{=}1)\\
&=\expect_{\distp}\left[\mathcal{L}\left(f^*_\mathcal{L}(\Bb), Y\right)\right]=\mathcal{L}^{\text{base}}.
\end{align}
\hfill$\square$

\subsection{The Generalized Principle of \air}
\label{sec:generalizednonpen}
We can reformulate the probabilistic definition of \air in terms of loss functions, which allows for generalization. We define $\distp_0 = \distp(\Bb|A=0)$. We start by the notion given in the definition: 
%\text{MI}(G(\Bb);Y|A=0) &\leq %\text{MI}(\hat{Y}(\Bb);Y|A=0).\\
%H(Y|A=0)-H(G(\Bb)|Y,A=0) &\leq %H(Y|A=0)-H(\hat{Y}(\Bb)|Y,A=0)\\
%H(G(\Bb)|Y,A=0) &\geq H(\hat{Y}(\Bb)|Y,A=0)\\
%\expect_{\bb \sim \distp_0}\mathcal{D}_{KL}\left(\distp(Y|\bb,A{=}0)\middle|\middle|\distp_g(Y|\bb)\right) &\geq \expect_{\bb \sim \distp_0}\mathcal{D}_{KL}\left(\distp(Y|\bb,A{=}0)\middle|\middle|\distp(Y|\bb)\right).\\
\begin{align}
\mathcal{D}_{KL}\left(\distp(Y|\Bb,A{=}0)\middle|\middle|\distp_g(Y|\Bb)\right) &\geq \mathcal{D}_{KL}\left(\distp(Y|\Bb,A{=}0)\middle|\middle|\distp(Y|\Bb)\right) \\
\expect_{\bb \sim \distp_0}\mathcal{D}_{KL}\left(\distp(Y|\bb,A{=}0)\middle|\middle|\distp_g(Y|\bb)\right) &\geq \expect_{\bb \sim \distp_0}\mathcal{D}_{KL}\left(\distp(Y|\bb,A{=}0)\middle|\middle|\distp(Y|\bb)\right).
\end{align}
The Kullback-Leibler divergence can be decomposed as $\mathcal{D}_{KL}(\distp||\distq) = H(\distp)+CE(\distp||\distq)$, which results in: 
\begin{align}
\Longleftrightarrow  \expect_{\bb \sim \distp_0} \text{CE}\left(\distp(Y|\bb,A{=}0)\middle|\middle|\distp_g(Y|\bb)\right) &+ \expect_{\bb \sim \distp_0} H\left[Y|\bb,A{=}0\right] \\
\geq \expect_{\bb \sim \distp_0}\text{CE}\left(\distp(Y|\bb,A{=}0)\middle|\middle|\distp(Y|\bb)\right) &+  \expect_{\bb \sim \distp_0} H\left[Y|\bb,A{=}0\right]\\
\Longleftrightarrow  \expect_{\bb \sim \distp_0}\text{CE}\left(\distp(Y|\bb,A{=}0)\middle|\middle|\distp_g(Y|\bb)\right) &\geq \expect_{\bb \sim \distp_0}\text{CE}\left(\distp(Y|\bb,A{=}0)\middle|\middle|\distp(Y|\bb)\right)\\
\Longleftrightarrow  \expect_{\bb \sim \distp_0}\expect_{Y \sim \distp(Y|\bb,A=0)}\left[-\log \distp_g(Y|\bb)\right] & \geq \expect_{\bb \sim \distp_0}\expect_{Y \sim \distp(Y|\bb,A=0)}\left[-\log \distp(Y|\bb) \right]
\end{align}
The inner expectation is equivalent to the BCE loss for a specific $\bb$. Averaged over all $\bb \sim \distp_0$ we obtain.
\begin{align}
\Rightarrow \expect_\distp\left[\text{BCE}\left(g(\Bb), Y\right)\middle| A=0 \right] &\geq \expect_\distp\left[\text{BCE}\left(f_{\text{BCE}}^{*}(\Bb), Y\right)\middle| A=0 \right].
\end{align}
This notion allows for generalization by replacing $\text{BCE}$ with some general loss function $\mathcal{L}$. Doing so results in 
\begin{align}
\expect_\distp\left[\mathcal{L}\left(g(\Bb), Y\right)\middle| A=0 \right] &\geq \expect_\distp\left[\mathcal{L}\left(f_{\mathcal{L}}^{*}(\Bb), Y\right)\middle| A=0 \right],
\end{align}
the version of the desideratum of \air mentioned in the main paper.
\hfill$\square$ 
\subsection{PUC under strategic withholding of data}
\label{sec:app_offstrategic}
To prove \Cref{corr:offstrategic}, we first note that the decision maker can only realize improvements over the base model in the setup of strategic interactions for individuals by offering them a lower premium than the prediction of the base model. Otherwise, they would strategically not provide their data.

It is only beneficial for the decision maker to do so if there exists an $y^\prime \leq y_{\text{base}} \coloneqq \mathbb{F}^{\mathcal{L}}[Y|\bb]$ with a lower expected loss, i.e.,
\begin{align}
\mathbb{E}_Y\left[\mathcal{L}(y^\prime, Y) | \Bb=\bb, Z=z\right] \leq \mathbb{E}_Y\left[\mathcal{L}(y_{\text{base}}, Y) | \Bb=\bb, Z=z\right]
\end{align}
Due to the convexity of the loss $\mathcal{L}$, this expected value will as well be convex in the prediction $y'$ and we will also have $\mathbb{F}^{\mathcal{L}}[Y|\bb, z]\leq \mathbb{F}^{\mathcal{L}}[Y|\bb]$. The loss-minimal prediction would be $f(b, a = 1, z) =\mathbb{F}^{\mathcal{L}}[Y|\bb, z]$, which will not be hindered through strategic actions. This however results in a \PUC-model again, as $\mathbb{F}^{\mathcal{L}}[Y|\bb, z] = \mathbb{F}^{\mathcal{L}}[Y|\bb, A=1, z]$, because the sharing decision does not influence the label given $\Bb, Z$.
\hfill$\square$\par

\textbf{PUC with monotonicity constraints.} A similar argument can be made when monotonicity constraints need to be enforced, i.e., the outcome can only decrease over the base model with more information provided.  We can consider each optional feature value $z$ separately for sharers. If the sample comes with a better average $\mathbb{F}^{\mathcal{L}}[Y|\bb, z]\leq \mathbb{F}^{\mathcal{L}}[Y|\bb]$ than the base prediction, we can confidently return this full-feature optimal prediction. In the contrary case, where $\mathbb{F}^{\mathcal{L}}[Y|\bb, z] > \mathbb{F}^{\mathcal{L}}[Y|\bb]$, the best prediction that the decision maker is allowed to make is the base feature models prediction (due to the convexity of the loss function). This is equivalent to dropping the optional feature in this case and using the corresponding \PUC model.

\subsection{Equivalence of Expectations for the Resampling model}
In this section, we show that the resampling technique proposed in this work converges to the desired outcome. Therefore, we show that in the infinite sample-limit, the optimum reached when optimizing the loss over the modified distribution corresponds to the desired \PUC model. 

We introduce the usual mapping $\indset(\bba) \coloneqq \left\{i~\middle|~\Bbm_i = 1, i=1,\ldots, r \right\}$ to denote the set of all indices that are 1 in the vector $\bba$ but also use $\indicone_S$ to denote the binary indicator vector where all components corresponding to indices in $S$ are set to 1 and to zero otherwise, i.e., $\left(\indicone_S\right)_i = \left\{1~\text{if}~i \in S, \text{else}~0 \right\}$. Note that these operations invert each other such that $\indicone_{\indset(\bba)} = \bba$.
We can show that the optimal prediction $\hat{y} = \hat{y}(\bb, \bba, \bzstar)$ is given by:
\begin{align}
\hat{y} = &\fexpect_{((\bb, \bba, \bzstar),y)\sim\overline{\distp}}\left[Y|\bb, \Bba=\bba, \Bbzstar=\bzstar \right] = \argmin_{\hat{y}} \expect_{\left((\bb, \bba, \bzstar),y\right)\sim\overline{\distp}} \left[\mathcal{L}(\hat{y} , y)|\bb, \Bba=\bba, \Bbzstar=\bzstar\right]\\
&=\argmin_{\hat{y} } \sum_{\indset(\bba)\subseteq S}\distp(\Bba{=}\indicone_S|\Bb{=}\bb, \Bbz_{\indset(\bba)}{=}\bz_{\indset(\bba)})\expect_\distp\left[\mathcal{L}(\hat{y}, y)\middle| \Bba=\indicone_S, \Bb=\bb, \Bbz_{\indset(\bba)}=\bz_{\indset(\bba)}\right]\label{eqn:refpluginpbar}\\
&= \argmin_{\hat{y}} \expect_\distp\left[\mathcal{L}(\hat{y}, y)\middle| \indset(\bba) \subseteq \indset(\Bba),\Bb=\bb, \Bbz_{\indset(\bba)}=\bz_{\indset(\bba)}\right]\\
&=\argmin_{\hat{y}} \expect_\distp\left[\mathcal{L}(\hat{y}, y)\middle| \Bba_{\indset(\bba)}=\BOne,\Bb=\bb, \Bbz_{\indset(\bba)}=\bz_{\indset(\bba)}\right]\\
&=\fexpect_{\distp}\left[Y|\Bb=\bb,\Bba_{\indset(\bba)}=\BOne,\Bbz_{\indset(\bba)}=\bz_{\indset(\bba)}\right].
\end{align}
In \Cref{eqn:refpluginpbar}, we use the fact that we can  express the distribution $\distpo$ for a subset of inputs with $\Bba = \bba$ as a mixture of $\distp$, averaged over all subsets of inputs $S$ with more optional features than $\bba$, weighted equally but with the optional information erased. This is a result of the data augmentation procedures that defines $\distpo$. The total weight is just a factor and does not play a role in the $\argmin$ operation. The following steps are just reformulations of the expression.
\hfill$\square$ 
\subsection{Proof: Convergence of the sample approximation for a finite feature space}\label{app:finite_sample_proof}
In this section we provide a general estimation of the error of a non-parametric regressor from a finite number of samples on a finite feature space $\spaceX$ (e.g., finite, discrete features) and a label space $\spaceY$ that can be either continuous or discrete.
Before we can prove the main result, we establish the following lemma.
\begin{lemma}
The density $\distpo$ that is obtained from $\distp$ by applying the augmentation strategy described in the paper (\OFFIDA) is related to the original density through the following relation:
\[
\forall \bx \in \spaceX: \distpo(\bx) \geq \frac{1}{2^r} \distp(\bx).
\]
In particular, this implies that the support of $\distpo$ is at least as big as the support of $\distp$.
\label{lem:relationdistpo}
\end{lemma}
\textbf{Proof.} The resampling procedure consists of two steps. First, a reweighting is done. As we state in the main text, this reweighting from $\distpo$ can be implemented through rejection sampling with samples from $\bx=(\bb, \bba, \bzstar) \sim \distp$. Samples are passed on the the next stage with a probability of $\frac{2^{\lvert\indset(\bba)\rvert}}{2^r}$. Using this scheme, we know that for a certain $\bx=(\bb, \bba, \bzstar)$, the probability of the sample to be observed after applying only the reweighting step is bounded by $\frac{2^{\lvert\indset(\bba)\rvert}}{2^r}\distp(\bx)$. To see this, we can consider the worst case, where all other samples are passed on with probability $1$ and only the considered vector $\bx$ is downweighted by a factor of $\frac{2^{\lvert\indset(\bba)\rvert}}{2^r}$. If the other samples are also downweighted, this is a strict lower bound.
In the second step, some optional features are dropped at random with a probability of $\frac{1}{2}$. We are interested in $\distpo(\bx)$, the probability of obtaining the exact original sample with all its optional features still present. The probability that all optional features remain present with the Bernoulli distribution used, is given by $\frac{1}{2^{\lvert\indset(\bba)\rvert}}$. Bringing it all together we obtain:
\begin{align}
\forall \bx \in \spaceX: \distpo(\bx) \geq \frac{2^{\lvert\indset(\bba)\rvert}}{2^r}\frac{1}{2^{\lvert\indset(\bba)\rvert}}\distp(\bx) = \frac{1}{2^r} \distp(\bx).
\end{align}

\begin{theorem}[Convergence of Finite Sample Approximation]
Suppose a finite feature space $\spaceX$ and a numerical label space $\spaceY \subseteq \mathbb{R}$. Suppose all conditional expectations $\mu_\distpo(\bx) \coloneqq\expect_\distpo\left[y\middle|\bx\right]$ and the conditional variances $\sigma^2_\distpo(\bx) \coloneqq \Var_\distpo\left[y\middle|\bx\right]$ exist (and thus are finite). We can estimate a (discrete) non-parametric regressor $\hat{\mu}^\mathcal{D}: \spaceX \mapsto \mathbb{R}$ from a finite number $N$ of independent, identically distributed observations $\mathcal{D} =\left(\bx_i, y_i\right)_{i=1\ldots N}$ from $\distpo$ which satisfies:
\begin{align}
\expect_{\bx \sim \distp, \mathcal{D}\sim \overline{\distp}}\left[\left(\hat{\mu}^{\mathcal{D}}(\bx)-\mu_{\overline{\distp}}(\bx)\right)^2\right] \leq \frac{2^r\lvert\spaceX\rvert^2(\sigma_{\max}^2+\mu_{\max}^2)}{N} +\mathcal{O}\left(\frac{1}{N^2}\right),
\end{align}
where $\sigma_{\max}^2 \coloneqq \max_{\bx \in \spaceX} \sigma^2_\distpo,(\bx)$ and $\mu_{\max}^2 \coloneqq \max_{\bx \in \spaceX} \mu^2_\distpo(\bx)$. The expected squared deviation to the optimal estimator converges at an order of $\mathcal{O}\left(\frac{1}{N}\right)$.
\end{theorem}

\textbf{Proof.} Before we proof the rate of convergence, we first define the estimator for which we establish this bound. We can draw $N$ samples $\mathcal{D} \sim \left(\bx_i, y_i\right) \sim \overline{\distp}$. Then, we split these into $\lvert \spaceX \rvert$ equal batches of size $M = \Big\lfloor \frac{N}{\lvert\spaceX\rvert}\Big\rfloor$ samples. We can thus assign each possible feature value $\bx \in \spaceX$ a batch $\text{Batch}(\bx) \subset [N]$, of samples, although the value the features $\bx_i$ for $i$ in the batch corresponding to $\bx$ are still randomly distributed according to $\overline{\distp}$. We only use $M$ different samples to estimate each conditional mean. Denoting the true conditional mean by $\mu_\bx \coloneqq \mu_\distpo(\bx)$ and its estimate by $\hat{\mu}_\bx^{\mathcal{D}}\coloneqq\hat{\mu}^{\mathcal{D}}(\bx)$ for the feature $\bx \in \spaceX$, we estimate:
\begin{align}
\hat{\mu}_\bx^{\mathcal{D}} = \frac{\sum_{(\bx_i, y_i) \in \text{Batch}(\bx)} y_i \bm{\delta}_{\bx_i=\bx} }{1 + \sum_{(\bx_i, y_i) \in \text{Batch}(\bx)} \bm{\delta}_{\bx_i=\bx}},
\label{eq:cond_mean_estimator}
\end{align}
where $\bm{\delta}_{\bx_i=\bx} = \left\{1 ~\text{if}~\bx_i=\bx, \text{else}~0\right\}$ denotes the indicator function. Depending on the number $b_{\bx}=\sum_{(\bx_i, y_i) \in \text{Batch}(\bx)} \bm{\delta}_{\bx_i=\bx}$ of samples with matching feature values that are used in the estimation of $\hat{\mu}_\bx$, the estimator is slightly biased as $\expect\left[\hat{\mu}_\bx^{\mathcal{D}}\middle|b_\bx =q\right] = \frac{q\mu_\bx}{q+1}$ but the bias will vanish as $b_\bx \rightarrow \infty$. Note that $b_\bx$ is a random variable. The variance of the estimator on iid samples is $\text{Var}_{\distpo}\left[\hat{\mu}_\bx^{\mathcal{D}}\middle|b_\bx =q\right] = \frac{q\sigma_\bx^2}{(q+1)^2}$. Without loss of generality, we will suppose $\overline{p}_\bx \coloneqq \overline{\distp}(\bx) > 0$: By Lemma~\ref{lem:relationdistpo}, we obtain $\distpo(\bx)\geq \frac{1}{2^r}\distp(\bx)$. Thus, $\distpo(\bx)=0$ implies $\distp(\bx)=0$ and the error of the estimator will not play a role in expected squared error we are interested in obtaining. By the well-known Bias-Variance decomposition, the square error of the single estimator $\mu_\bx$ for a given $b_\bx=q$ can be written as:
\begin{align}
\expect_{\mathcal{D}\sim\overline{\distp}}\left[\left(\hat{\mu}_\bx^{\mathcal{D}} - \mu_\bx\right)^2 \middle| b_\bx =q \right] &=  \left(\expect_{\overline{\distp}}\left[\hat{\mu}_\bx^{\mathcal{D}}\middle|b_\bx =q\right] - \mu_\bx\right)^2 + \text{Var}_{\overline{\distp}}\left[\hat{\mu}_\bx^{\mathcal{D}} \middle| b_\bx =q \right]\\
&=\left(\frac{q\mu_\bx}{q+1} -\mu_\bx\right)^2 + \frac{q\sigma_\bx^2}{(q+1)^2} = \left(\frac{1}{q+1}\right)^2\mu_\bx^2 + \frac{q\sigma_\bx^2}{(q+1)^2}\\
&\leq \frac{1}{q+1} \mu^2_\bx + \frac{(q+1)\sigma_\bx^2}{(q+1)^2} = \frac{1}{q+1} \left(\mu^2_\bx + \sigma_\bx^2 \right).
\end{align}
Due to the sampling procedure, the $b_\bx$ are independently binomially distributed with $b_\bx \sim \text{Bin}\left(M, \overline{p}_\bx\right)$.
Therefore, we can first aggregate the results for a single $\bx$ and then average over the entire distribution over $\spaceX$.
We obtain:
\begin{align}
    \expect_{\mathcal{D}\sim \overline{\distp}}\left[\left(\hat{\mu}_\bx^{\mathcal{D}} - \mu_\bx\right)^2 \right] &= \sum_{q=0}^M p(b_\bx=q)\expect_{\mathcal{D}\sim \distp}\left[\left(\hat{\mu}_\bx^{\mathcal{D}} - \mu_\bx\right)^2 \middle| b_\bx =q \right] \\
    &\leq  \sum_{q=0}^M \text{Bin}(q; M, \overline{p}_\bx) \frac{1}{q+1} \left(\mu^2_\bx + \sigma_\bx^2 \right) = \left(\mu^2_\bx + \sigma_\bx^2 \right)\expect_{b_\bx\sim \text{Bin}(M,\overline{p}_\bx)}\left[\frac{1}{q+1}\right] \\
    &= \left(\mu^2_\bx + \sigma_\bx^2 \right)\left(\frac{1}{\overline{p}_\bx(M+1)}\right)\left(1-(1-\overline{p}_\bx)^{M+1}\right)\label{eqn:binomialexpect}\\
    &\leq \left(\mu^2_\bx + \sigma_\bx^2 \right)\left(\frac{1}{\overline{p}_\bx(M+1)}\right) < \left(\mu^2_\bx + \sigma_\bx^2 \right)\left(\frac{1}{\overline{p}_\bx M}\right),
\end{align}
where $\text{Bin}(q; M, \overline{p}_\bx)=\binom{M}{q}(\overline{p}_\bx)^q(1-\overline{p}_\bx)^{M-q}$ is the probability given by the binomial law and the equality in \Cref{eqn:binomialexpect} is provided in \citep[p.271]{cribari2000note}. 
We aggregate this result to an expected value over samples from the original distribution $\distp$. The sample $\bx\sim\distp$ that the estimator is evaluated on and the data set $\mathcal{D}\sim\distpo$ are independent, and we can derive an expected error for the distribution $\distp$ over all features $\bx$, as:
\begin{align}
    \expect_{\bx \sim \distp,\mathcal{D} \sim \overline{\distp}}\left[\left(\hat{\mu}(\bx)-\mu(\bx)\right)^2\right] &= \sum_{\bx\in\spaceX} p_\bx  \expect_{\mathcal{D}\sim \distp}\left[\left(\hat{\mu}_\bx^{\mathcal{D}} - \mu_\bx\right)^2 \right] \\
    &< \sum_{\bx\in\spaceX} p_\bx \left(\mu^2_\bx + \sigma_\bx^2 \right)\left(\frac{1}{\overline{p}_\bx M}\right) \leq \sum_{\bx\in\spaceX} 2^r\,\overline{p}_\bx \left(\mu^2_\bx + \sigma_\bx^2 \right)\left(\frac{1}{\overline{p}_\bx M}\right)\\
    &\leq \sum_{\bx\in\spaceX} \frac{2^r(\mu_{\max}^2 + \sigma_{\max}^2)}{M}\\
    &=\lvert\spaceX\rvert \frac{2^r(\mu_{\max}^2 + \sigma_{\max}^2)}{M} =  \frac{\lvert\spaceX\rvert^2(2^r(\mu_{\max}^2 + \sigma_{\max}^2))}{M\lvert\spaceX\rvert} \leq \frac{2^r\lvert\spaceX\rvert^2(\mu_{\max}^2 + \sigma_{\max}^2)}{N-\lvert\spaceX\rvert +1}\\
    &=\frac{2^r\lvert\spaceX\rvert^2(\sigma_{\max}^2+\mu_{\max}^2)}{N} +\mathcal{O}\left(\frac{1}{N^2}\right),
\end{align}
where we use the fact that $2^r\,\overline{p}_\bx \geq p_\bx$ and the definitions of $\mu_{\max}^2, \sigma_{\max}^2$ as specified in the theorem.

\hfill$\square$

\subsection{Algorithms}\label{app:algorithms}
An example of how \puc through data augmentation can be incorporated in an SGD-type algorithm is provided in \Cref{alg:offsgd}.
\begin{algorithm}[tb]
\caption{PUC-SGD: SGD with Protected User Consent\label{alg:offsgd}}
\begin{algorithmic}
\Require Data set $\mathcal{D}$, Loss function $\mathcal{L}$, predictor $f_{\bm{\theta}}$ with parameters $\bm{\theta}$
%\Procedure{OFF-SGD}{$\mathcal{D}, \mathcal{L}, f_{\bm{\theta}}, \bm{\theta}$}\Comment{Perform fair SGD}
\State $\bw \gets \{\text{Distribution over}~\mathcal{D}~ \text{with}~\bw(\bx) \propto w(\bx)\}$
\While{$r\not=0$}
\State Sample batch $(\bx^{(1)}, y^{(1)}),\ldots, (\bx^{(k)},y^{(k)}) \sim \bw$
\For{$j = 1,\ldots,k$} \Comment{$\bx^{(j)}{=}(\bb^{(j)}, \bba^{(j)}, \bz^{*(j)})$}
\State $\bq \gets \text{Bernoulli}(0.5)$\Comment{iid. Bernoulli vector}
\State $\overline{\bba}^{(j)} = \bq \odot \bba^{(j)}$
\State $\overline{\bz}^{*(j)}_i = \left\{\bz_i^{*(j)}~\text{if}~\overline{\bba}^{(j)}_i{=}1,~\text{else}~ \texttt{N/A}\right\}, i\in[r]$
\State $\overline{\bx}^{(j)} \gets (\bb^{(j)}, \overline{\bba}^{(j)}, \overline{\bz}^{*(j)})$
\EndFor
\State $\bm{d}{\bm{\theta}} \gets \nabla_{\bm{\theta}}\left(\frac{1}{k}\sum_{j=1}^k\mathcal{L}\left(f_{\bm{\theta}}(\overline{\bx}^{(j)}), y^{(j)}\right)\right)$
\State $\bm{\theta} \gets \bm{\theta} - \gamma \bm{d}{\theta}$
\EndWhile\label{euclidendwhile}
\State \textbf{return} $\bm{\theta}$
%\Comment{The gcd is b}
%\EndProcedure
\end{algorithmic}
\end{algorithm}

\section{\puc on Simulated Distributions}
In this section we introduce two types of parametric data distributions with optional information that we use in our experiments with simulated data. They allow to independently control the complexity and to obtain as many samples as needed to study the convergence behavior. The first family is based on a Naive Bayes model (\Cref{sec:app_naivebayes}) with binary features, whereas the second one introduced in \Cref{sec:app_densityfamiliyofflr} allows for continuous features with logistic distributions.

\subsection{Na\"ive Bayes models revisited}
\label{sec:app_naivebayes}
We can also consider a Na\"{i}ve Bayes models with binary features which can possibly be unavailable as in Poole et al.~ \cite{poole2020conditioning}. Suppose that we have a Naive Bayes model with independent availability mechanisms, i.e., the availability of feature $i$ is only dependent on the label $y$ and the corresponding feature value $z_i$ and thus
$\distp(\bb, \bba, \bz, y) = \left( \prod_{i=1}^n \distp(b_i|y)\right)\left(\prod_{i=1}^r \distp(z_i| y) \distp(a_i|z_i, y)\right) \distp(y)$. A graphical representation of this model can be found in \Cref{fig:naivebayesmultipleindep}.
In this case, we can express the odds ratio as:
\begin{align}
    \text{odds}(Y=1|\bb, \bzstar_\indset, \Bba_\mathcal{I}=\BOne) = \frac{\distp(Y=1,\bb, \bz_\indset, \Bba_\mathcal{I}=\BOne)}{\distp(Y=0,\bb, \bz_\indset, \Bba_\mathcal{I}=\BOne)} =\\ 
    \left( \prod_{i=1}^n \frac{\distp(b_i|Y=1)}{\distp(b_i|Y=0)}\right)\left( \prod_{i \in \mathcal{I}}\frac{\distp(z_i|Y{=}1) \distp(A_i{=}1 |z_i, Y{=}1)}{\distp(z_i|Y{=}0) \distp(A_i{=}1 |z_i, Y{=}0)} \right) \frac{\distp(Y=1)}{\distp(Y=0)}.
\end{align}
As we furthermore suppose the features are binary, the odds are specified through the ratios $\frac{\distp(b_i|Y{=}1)}{\distp(b_i|Y{=}0 )}$ for $b_i \in \left\{0,1\right\}$ and  $\frac{\distp(z_i|Y{=}1)}{\distp(z_i|Y{=}0)}\frac{\distp(A_i{=}1 |z_i, Y{=}1)}{\distp(A_i{=}1 |z_i, Y{=}0)}$ for $z_i \in \left\{0,1\right\}$. This requires only $2r + 2n$ parameters to be specified in total.

\begin{figure}[ht]
    \centering
    \begin{tikzpicture}[circ/.style={circle,draw, inner sep=1pt,minimum
  size=5ex},tcirc/.style={circle, inner sep=1pt,minimum
  size=5ex}]
    \node (B2) at (-3,1) [tcirc] {$\cdots$};
    \node (Z2) at (3,1) [tcirc] {$\cdots$};
    \node (Y) at (0,3) [circ] {$\mathbf{Y}$};
    \node (B1) at (-5,1) [circ] {$\mathbf{B_1}$};
    \node (BN) at (-1,1) [circ] {$\mathbf{B_n}$};
    \node[dotted] (Z1) at (1,1) [circ] {$\mathbf{Z_1}$};
    \node[dotted] (ZR) at (5,1) [circ] {$\mathbf{Z_r}$};
    \node (M1) at (1,0) [circ] {$\mathbf{A_1}$};
    \node (MR) at (5,0) [circ] {$\mathbf{A_r}$};
    \node (Zstar1) at (1.9,-0.5) [circ] {$\mathbf{Z^*_1}$};
    \node (ZstarR) at (5.9,-0.5) [circ] {$\mathbf{Z^*_r}$};
   
    \draw[->] (Y) -> (Z1);
    \draw[->] (Y) -> (ZR);
    \draw[->] (Y) -> (Z2);
    \draw[->] (Y) -> (B1);
    \draw[->] (Y) -> (B2);
    \draw[->] (Y) -> (BN);
    \draw[->] (Z1) -> (M1);
    \draw[->] (ZR) -> (MR);
    \draw[->] (M1) -> (Zstar1);
    \draw[->] (MR) -> (ZstarR);
    \draw[->] (Z1) to [out=-35,in=90] (Zstar1);
    \draw[->] (ZR) to [out=-35,in=90] (ZstarR);
 \end{tikzpicture}
    \caption{The Naive Bayes model with independent availability mechanisms. We observe the Label $Y$, the base features $B_1$ to $B_n$ and the possibly unavailable features $Z^*_i= A_i\cdot Z_i$}
    \label{fig:naivebayesmultipleindep}
\end{figure}
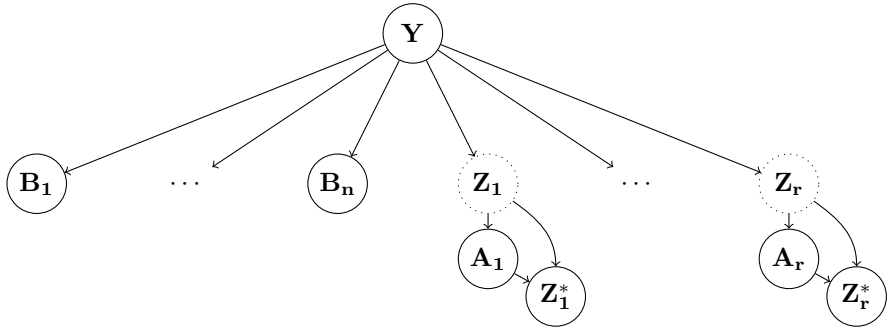

\begin{figure}[ht]     
\centering
     \scalebox{0.75}{
\begin{tikzpicture}[circ/.style={circle,draw, inner sep=1pt,minimum
  size=5ex},tcirc/.style={circle, inner sep=1pt,minimum
  size=5ex}]
    \node (Z2) at (3,0) [tcirc] {$\cdots$};
    \node (Y) at (0,2) [circ] {$\mathbf{Y}$};
    \node (B) at (3,2) [circ] {$\underset{i=1\ldots n}{\left\{\mathbf{B_i}\right\}}$};
    \node[dotted] (Z1) at (0,0) [circ] {$\mathbf{Z_1}$};
    \node[dotted] (ZR) at (5,0) [circ] {$\mathbf{Z_r}$};
    \node (M1) at (1,0) [circ] {$\mathbf{A_1}$};
    %\node (M2) at (4,0) [tcirc] {$\cdots$};
    \node (MR) at (6,0) [circ] {$\mathbf{A_r}$};
    \node (Zstar1) at (1,-1) [circ] {$\mathbf{Z^*_1}$};
    \node (ZstarR) at (6,-1) [circ] {$\mathbf{Z^*_r}$};
   
    \draw[->] (Y) -> (Z1);
    \draw[->] (Y) -> (ZR);
%    \draw[->] (Y) -> (Z2);
    \draw[->] (Y) -> (M1);
    \draw[->] (Y) -> (MR);
%    \draw[->] (Y) -> (M2);
    \draw[->] (B) -> (Z1);
    \draw[->] (B) -> (ZR);
%    \draw[->] (B) -> (Z2);
    \draw[->] (B) -> (M1);
    \draw[->] (B) -> (MR);
%    \draw[->] (B) -> (M2);
    \draw[->] (Y) -> (B);
    \draw (Z1) -- (M1);
    \draw (ZR) -- (MR);
    \draw[->] (M1) -> (Zstar1);
    \draw[->] (MR) -> (ZstarR);
    \draw[->] (Z1) to [out=-90,in=180] (Zstar1);
    \draw[->] (ZR) to [out=-90,in=180] (ZstarR);
 \end{tikzpicture}
 }
     \caption{The relaxed graphical model with independent missingness mechanisms given the the Label $Y$ and the base features $B_1$ to $B_n$. The observed, possibly missing features are $Z^*_i= M_i\cdot Z_i$.}
     \label{fig:relaxeddependence}
 \end{figure}
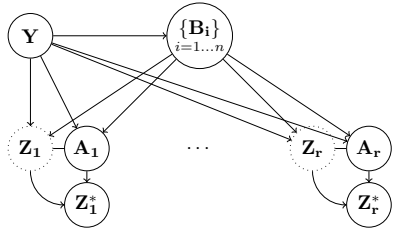

% Suppose that we have a Naive Bayes model, where the missingness is only dependent on the label
% $p(\bb, \Bbz, \Bba, y) = p(\bb|y)\left(\prod_{i=1}^r p(\Bbz_i|y) p(\Bba_i|y)\right) p(y)$. A graphical representation of this density is provided in \Cref{fig:naivebayesmultipleindep}.
% \begin{align}
%     \text{odds}(y=1|\bb, \bzstar_\indset, \Bba_\mathcal{I}=\BOne) = \frac{p(y=1,\bb, \Bbz_\indset, \Bba_\mathcal{I}=\BOne)}{p(y=0,\bb, \Bbz_\indset, \Bba_\mathcal{I}=\BOne)} =\\ 
%     \frac{p(\bb|y=1)}{p(\bb|y=0)}\left( \prod_{i \in \mathcal{I}}\frac{p(\Bbz_i|y{=}1) p(\Bba_i{=}1 |y{=}1)}{p(\Bbz_i|y{=}0) p(\Bba_i{=}1 |y{=}0)} \right)
% \end{align}
% Using common rearrangement techniques for binary variables such as
% \begin{align}
%     \frac{p(\Bbz_i|y{=}1)}{p(\Bbz_i|y{=}0)} &= \frac{p(\Bbz_i=1|y{=}1)}{p(\Bbz_i=1|y{=}0)}^{z_i}\frac{p(\Bbz_i=0|y{=}1)}{p(\Bbz_i=0|y{=}0)}^{1-z_i}\\
%     &=\left(q_i^+\right)^{z_i} \left(q_i^-\right)^{1-z_i}
% \end{align}
% and 
% \begin{align}
%     \frac{p(\Bba_i{=}1|z_i, y{=}1)}{p(\Bba_i{=}1|z_i,y{=}0)}
%     \coloneqq \left(\mu_i^+\right)^{z_i} \left(\mu_i^-\right)^{1-z_i}
% \end{align}
% we arrive at 
% \begin{align}
% &\text{odds}(y=1|\bb, \bzstar_\indset, \Bba_\mathcal{I}=\BOne)\\
% =&\exp \left[ \bw^\top\bb +t + \left( \sum_{i \in \mathcal{I}}  \Bbz_i \underset{\beta_i}{\underbrace{(\log q_i^+ - \log q_i^{-} +\log \mu_i^+ - \log \mu_i^{-})}} +  \underset{s_i}{\underbrace{\log \mu_i^- + \log q_i^-}} \right)\right]
% \end{align}
\newcommand{\pushright}[1]{\ifmeasuring@#1\else\omit\hfill$\displaystyle#1$\fi\ignorespaces}
\subsection{A parametric familiy of distributions with logistic subset models}
\label{sec:app_densityfamiliyofflr}
In this section, we describe a set of conditions that can be used to construct a family of densities that will have a logistic form when applying \PUC. Formally, this means that for each $\indset \subseteq [r]$ of optional features being present, there exists a $\bw \in \mathbb{R}^n$, $\bm{\beta} \in \mathbb{R}^{|\indset|}$, and $s \in \mathbb{R}$ that allow to represent the $\text{\normalfont odds}(Y=1|\bb, \bzstar_\indset, \Bba_\mathcal{I}=\BOne)$ in the form:
\begin{align*}
    \frac{\distp\left(Y=1\middle| \Bb=\bb, \Bbzstar_\indset=\bzstar_\indset, \Bba_\indset = \BOne \right)}{\distp\left(Y=0\middle| \Bb=\bb, \Bbzstar_\indset=\bzstar_\indset, \Bba_\indset = \BOne \right)} = \exp \left[\bw(\indset)^\top\bb + \bm{\beta}(\indset)^\top\bz^*_\indset + s(\indset) \right].
    \label{cond:logisticsubset}
\end{align*}
This allows for complex dependencies (e.g., the base feature can influence availability and value of the optional features), while also allowing to compute the ground truth \PUC model relatively easy. Formally,  we suggest the following assertions and show that they will result in logistic models for each set of features present:
\begin{align*}
&\text{C1: Base model is logistic:} & \distp(Y=1|\bb) = \sigma(\bw^\top\bb + t)&\\
& \text{C2: Availability is cond. independent} 
 &~\forall \mathcal{I} \subseteq [r]: \distp(\Bba_\mathcal{I}=\BOne |\bb,y) = \prod_{i \in \mathcal{I}}\distp(A_i = 1 |\bb,y)&\\
&\text{C3: Mut. independence when present:}&
\forall \mathcal{I} \subseteq [r]: \distp(\bz_\mathcal{I}|\bb,y, \Bba_\mathcal{I}=\BOne) = \prod_{i \in \mathcal{I}} \distp(z_i|\bb,y, A_i{=}1)&\\
&\text{C4: Availability is sigmoidal:}
&~\distp(A_i = 1 |\bb,Y=1) = N(\bb)\sigma\left(\bu_i^\top \bb + \lambda_i\right)&\\
&&~\distp(A_i = 1 |\bb,Y=0) =  N(\bb)-\distp(A_i = 1 |\bb,Y=1)&\\
&\text{C5: Base-dependent Normal distributions:}&~\distp(z_i|\bb,y, A_i=1) \sim \mathcal{N}\left(\bv_i^\top \bb + \tau_i(y), \eta^2\right)&.
\end{align*}
Intuitively, after ensuring that the base feature model has a logistic form (C1), the next two assumptions follow directly from the graphical dependency model (C2, C3), see \Cref{fig:relaxeddependence}. C4 suggests the availability should sigmoidally depend on the base features with a different offset for each class. The last condition (C5) allows the $z_i$ to depend on the base features $\bb$ with the same $\bv_i$ for both classes $y$. However, a different offset by  the coefficient $\tau_i$ can be added for each class.  The entire distribution can be specified through the parameters $\bw$, $t$, and $\bm{u}_i$, $\bm{v}_i$, $\lambda_i$, $\tau_i(0)$, $\tau_i(1)$ and $s_i$ for each missing feature in $i=1\ldots r$.

In this special case, we can show that each of the models required will have the form of logistic regression again.
We start by determining some density ratios that will arise later:
\begin{align}
    \frac{\distp(A_i = 1 |\bb,Y{=}1)}{\distp(A_i = 1 |\bb,Y{=}0)} = \frac{N(\bb)\sigma\left(\bu_i^\top \bb + \lambda_{i} \right)}{N(\bb)\left(1-\sigma\left(\bu_i^\top \bb + \lambda_i\right)\right)} = 
    \exp \left(\bu_i^\top \bb + \lambda_i \right),
\end{align}
where the identity $\frac{\sigma(x)}{1-\sigma(x)}=\exp(x)$ was used.
Furthermore,
\begin{align}
    \frac{\distp(z_i|\bb,Y=1, A_i=1)}{\distp(z_i|\bb,Y=0, A_i=1)} = \frac{\exp\left(-\frac{\left(z_i - \bv_i^\top \bb - \tau_{i1}\right)^2}{2\eta^2}\right)}{\exp\left(-\frac{\left(z_i - \bv_i^\top \bb - \tau_{i0}\right)^2}{2\eta^2}\right)} \\
    = \exp \left[ -\frac{\left(z_i - \bv_i^\top \bb - \tau_{i1}\right)^2-\left(z_i - \bv_i^\top \bb - \tau_{i0}\right)^2}{2\eta^2}\right] \\
    = \exp \left[\frac{\left(z_i - \bv_i^\top \bb \right)^2 -2\left(z_i - \bv_i^\top \bb\right)\tau_{i1} +\tau_{i1}^2-\left(z_i - \bv_i^\top \bb \right)^2 +2\left(z_i - \bv_i^\top \bb\right)\tau_{i0} -\tau_{i0}^2}{-2\eta^2}\right]\\
    = \exp \left[\frac{2\left(\tau_{i1}-\tau_{i0}\right)\left(z_i - \bv_i^\top \bb\right)+\tau_{i0}^2-\tau_{i1}^2}{2\eta^2}\right]\\
    = \exp \left[\underset{\beta_i}{\underbrace{\eta^{-2}\left(\tau_{i1}-\tau_{i0}\right)}}z_i - \underset{\bm{\gamma}^\top_i}{\underbrace{\eta^{-2}\left(\tau_{i1}-\tau_{i0}\right)\bv_i^\top}} \bb+\underset{\theta_i}{\underbrace{\frac{1}{2}\eta^{-2}\left(\tau_{i0}^2-\tau_{i1}^2\right)}}\right].
\end{align}
Let $\mathcal{I} \subseteq [r]$ be the set index set of of present features once again. We can insert the previous results and obtain:
\begin{align}
    \text{odds}(Y=1|\bb, \Bbz_{\mathcal{I}}, \Bba_{\indset}{=}\BOne) &= \frac{\distp(Y=1,\bb, \Bbz_{\mathcal{I}}, \Bba_{\indset}{=}\BOne)}{\distp(Y=0,\bb, \Bbz_{\mathcal{I}}, \Bba_{\indset}{=}\BOne)}\\
    &= \frac{\distp(Y=1|\bb)}{\distp(Y=0|\bb)}\frac{\distp(\Bba_\mathcal{I}=\BOne |\bb,Y{=}1)}{\distp(\Bba_\mathcal{I}=\BOne |\bb,Y{=}0 )}\frac{\distp(\Bbz_\mathcal{I}|\bb,Y{=}1, \Bba_\mathcal{I}{=}\BOne)}{\distp(\Bbz_\mathcal{I}|\bb,Y{=}0, \Bba_\mathcal{I}{=}\BOne)}\\
    &= \frac{\distp(Y=1|\bb)}{\distp(Y=0|\bb)}\left( \prod_{i \in \mathcal{I}}\frac{\distp(z_i|\bb,Y{=}1, A_i{=}1) \distp(A_i{=}1 |\bb,Y{=}1)}{\distp(z_i|\bb,Y{=}0, A_i{=}1) \distp(A_i{=}1 |\bb,Y{=}0)} \right)\\
    &=\exp(\bw^\top\bb + t + \sum_{i \in \mathcal{I}}\underset{\bm{\omega}_i^\top}{\underbrace{(\bu_i -\bm{\gamma}_i)^\top}} \bb + \beta_i z_i + \underset{s_i}{\underbrace{\lambda_i + \theta_i}}).
\end{align}
As this derivation shows, each subset model will again be of the logistic form.
On a sidenote, the probability of a true model with no fairness constraints can be estimated as:
\begin{align}
 \text{odds}(Y=1|\bb, \Bbz, \Bba) &= \frac{\distp(Y=1,\bb, \Bbz_{\mathcal{I}}, \Bba_{\indset}{=}\BOne), \Bba_{\overline{\indset}}{=}\mathbf{0})}{\distp(Y=0,\bb, \Bbz_{\mathcal{I}}, \Bba_{\indset}{=}\BOne, \Bba_{\overline{\indset}}{=}\mathbf{0})}\\
    &= \frac{\distp(Y=1|\bb)}{\distp(Y=0|\bb)}\frac{\distp(\Bba_\mathcal{I}=\BOne |\bb,Y{=}1)}{\distp(\Bba_\mathcal{I}=\BOne |\bb,Y{=}0 )}\frac{\distp(\Bbz_\mathcal{I}|\bb,Y{=}1, \Bba_\mathcal{I}{=}\BOne)}{\distp(\Bbz_\mathcal{I}|\bb,Y{=}0, \Bba_\mathcal{I}{=}\BOne)}\frac{\distp(\Bba_{\overline{\indset}}{=}\mathbf{0}|\bb,Y{=}1)}{\distp(\Bba_{\overline{\indset}}{=}\mathbf{0}|\bb,Y{=}0)}.
 \end{align}

\section{Additional Experimental Results and Details}
\subsection{Comparing \PUC to existing fairness notions}
\label{sec:app_empiricalfairnessnotions}
\begin{figure*}[tb]
\centering
\begin{subfigure}[h]{0.23\linewidth}
    \centering
    \includegraphics[width=\linewidth]{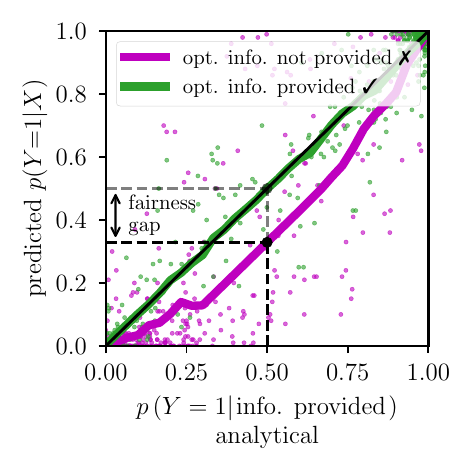}
    \caption{Fairness through unawareness}
    \label{fig:vanilla_classifier}
\end{subfigure}
\begin{subfigure}[h]{0.23\linewidth}
    \centering
    \includegraphics[width=\linewidth]{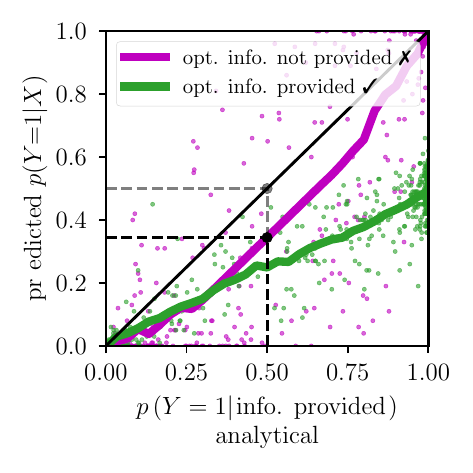}
    \caption{Statistical parity\\~}
\end{subfigure}
\begin{subfigure}[h]{0.23\linewidth}
    \centering
    \includegraphics[width=\linewidth]{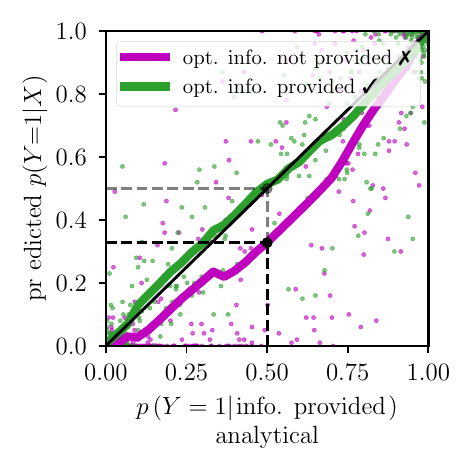}
    \caption{Equalized odds\label{fig:equalizedoss}\\~}
\end{subfigure}
\begin{subfigure}[h]{0.23\linewidth}
    \centering
    \includegraphics[width=\linewidth]{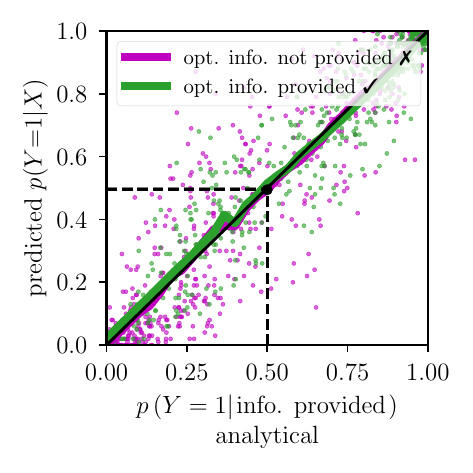}
    \caption{Protected User Consent (ours)}
\end{subfigure}
\caption{
\textbf{Standard models treat users who do not share optional information not according to the data they provided}.
In this work, users can provide information on optional features and only the provided information should be used in the decision making.
We show calibration curves for a model without fairness considerations (a) and with common fairness constraints enforcing statistical parity (b) and equalized odds (c) with respect to a model that uses only the explicity provided information (\bfm in case of no optional information, \ffm in case of optional information).
The first three models can penalize users not sharing the optional information (fairness gap in left panel), whereas a model trained with Protected User Consent through \OFFIDA (d) exhibits no systematic bias.
Models are probabilistic Random Forests trained on a synthetic data set (see \Cref{sec:app_teaseroff}). \label{fig:teaseroff}}
\end{figure*}

\label{sec:app_teaseroff}
In this section, we visually show the effect of not compensating for information contained in the decision to share data. We refer \Cref{fig:teaseroff}, were we compute probabilities for positive outcomes for a standard model ("fairness through unawareness") and other fairness-constrained models. The figure shows that all models apart from PUC, are not calibrated \emph{with respect to the data explicitly provided}. The data set used to create the Figure was sampled according to the logistic family described in \Cref{sec:app_densityfamiliyofflr}. The feature value distributions follow a logistic form. There were two base features and one optional feature. The availability and the values of this feature was dependent on the label and the value of the base features as described in the mentioned section. Specifically, the following parameters were used to instantiate the logistic family described in \Cref{sec:app_densityfamiliyofflr}:
\begin{center}
    \begin{tabular}{cc}
    \toprule
        base features & n=2, $\bb \sim \mathcal{N}(\bm{0}, 5\bm{I})$, $\bm{w}{=}(-1.5, 1.0)^\top$, $t{=}0$ \\
        opt. feature 1 & $\bm{u}_1 =(0.8, 0.4)^\top$, $\bm{v}_1=(0, 1)^\top$, $\lambda_1{=}0.7$, $\tau_1(0){=}-0.25$, $\tau_1(1){=}0.25$\\
    \bottomrule  
\end{tabular}
\end{center}

The models used in these experiments were \texttt{sklearn} RandomForests with default parameters. To incorporate the Fairness constraints of Statistical Parity and Equalized Odds, we leverage the \texttt{fairlearn}\footnote{\url{https://fairlearn.org/}} library \citep{bird2020fairlearn}, which implements the ExponentiatedGradient algorithm by \citet {agarwal2018reductions}. Although this algorithms only returns an approximate solution, we verified that the corresponding fairness gaps for Statistical Parity and Equalized Odds were substantially improved.

\label{sec:app_experimentalresults}
\subsection{Data Sets and Preprocessing}
\label{sec:app_preprocessing}
The diabetes data set\footnote{\url{https://www.kaggle.com/s/mathchi/diabetes-data-set}} was collected by the National Institute of Diabetes and Digestive and Kidney Diseases. It contains diagnostic measurements of female patients that are at least 21 years old. % and of Pima Indian Heritage. 
The target variable "Outcome" describes whether or not a person has diabetes.

The COMPAS data set\footnote{\url{https://www.kaggle.com/s/danofer/compass}} was originally collected by ProPublica and contains features describing criminal defendants in Broward County, Florida. It also contains their respective recividism score provided by the COMPAS algorithm and whether or not they reoffended within the following two years. For our analysis, we only kept features relevant for the prediction of recividism within the next two years and dropped irrelevant features such as name or date. Furthermore, we turned the categorical features race, sex and
charge degree into numerical features by encoding the categories with integers.

The UCI adult data set\footnote{\url{https://archive.ics.uci.edu/ml/datasets/Adult/}} is one of the most poplar tabular data sets and has appeared in over 300 publications \citep{ding2021retiring}. The goal is to predict whether an individuals yearly income is above 50k\$ (worth of 1994). 

The California Housing data set\footnote{\url{https://www.kaggle.com/datasets/camnugent/california-housing-prices}} contains infomration and average prices of properties in certain areas in the state of California, USA. The regression target is to predict the value of a property.
Because the income values range over several values of magnitudes, we apply $\log$ normalization to the label.

The ACSIncome data set (``income'') is derived from US census data in the work of \citet{ding2021retiring}. Code to download it is available online\footnote{\url{https://github.com/zykls/folktables/tree/main/folktables}}. As for Adult, the goal is to predict an inviduals yearly income. It features are similar from the one used in the Adult data set, however the exact incomes of each person are reported, and the data set can therefore be used in a regression setting. Because the income values range over several values of magnitudes, we apply $\log$ normalization to the labels.

%The Home Equity Line of Credit (HELOC) data set\footnote{\url{https://community.fico.com/s/explainable-machine-learning-challenge?tabset-158d9=3}} is a large collection of HELOC applications from anonymized homeowners, collected by the financial services provider FICO. The target variable RiskPerformance is "Bad" if the applicant was at least 90 past due within the two years after opening the credit account. "Good" and "Bad" were encoded as 1 and 0, respectively. We dropped the feature "ExternalRiskEstimate" from the data set since it seemed heavily correlated with all other important features (most likely it is a function of the other features) and thus minimized the effect of introducing stochastic availability.

%The Law School Admission data set\footnote{\url{https://github.com/mkusner/counterfactual-fairness}} contains information on students from law schools across the United States. Features are collected prior to their entry to law school and include race, sex, entrance exam scores (LSAT), grade-point average (GPA) and regional group. The predicted variable is the z-score of the first year average grade (ZFYA). Note that we use this data set in a binary classification manner and only predict if the z-score is above the average.

The Health Insurance (``insurance'') dataset\footnote{\url{https://api.openml.org/d/44993}} contains insurance data from individuals. It is a regression dataset, where inferences about the number of hours worked are to be made (whrswk, hours worked per week). We use the experience (years of potential work experience) as optional feature in the task.

We furthermore use two datasets with natural missing features.
The UCI horse colic dataset\footnote{\url{https://archive.ics.uci.edu/ml/datasets/Horse+Colic}} (``colic'') is a database of leasion surgeries on horses and contains a number of health attributes such as temperatures, pulse, respiratory rate and others. The target feature describes the outcome of the pathology. We use the feature abdominocentesis appearance as optional feature, which describes the appearance of fluid that is obtained from the abdominal cavity. This information is not available for each horse in the database and thus comes with natural missingness.

The water treatment dataset\footnote{\url{https://api.openml.org/d/940}} contains features describing the operational state of a water treatment plant, which is to be classified as either postive or negative. We use the feature RD-DBO-P (``oxygen demand'') as optional feature, which describes the Biological demand of oxygen in primary settler and comes with missing values.
%The Student performance data set\footnote{\url{https://archive.ics.uci.edu/ml/datasets/Student+Performance}}, referred to by the name of ``grades'', in this work can be found on the UCI Machine Learning repository.. The prediction goal is to predict the grades from certain demographic features. Many features in this data set are not highly predictive, which is why we only use the predictive features: 'school', 'sex', 'age', 'address', 'famsize', 'Medu', 'Fedu', 'traveltime', 'studytime', 'failures', 'famrel', 'freetime', 'goout', 'Dalc', 'Walc', 'health', 'absences'. Details on these features can be found with the data set's source. The label is the average of the three grades reported. We scale the value range from the orignal 0-20 to a range of $[-1,1]$.

Across all data sets, multi-value categorical features were one-hot encoded. We provide an overview of the characteristics of the different data sets in \Cref{tab:s}.
\begin{table}[ht]
\centering
\begin{tabular}{cccc}
\toprule
Data Sets & Label & Num. features & Num. samples ($N$)  \\ 
\midrule
diabetes  & Outcome           & 8             & 768    \\     
compas    & two\_year\_recid  & 9             & 7192    \\    
adult & ZFYA           & 5             & 21791         \\
california housing  & med\_house\_val & 9  & 20640 \\
income & income & 10 & 19567 \\
insurance & whrswk & 11 & 22272 \\
water & binaryClass & 36 & 527 \\
colic & pathology\_cp\_data & 26 & 368 \\
\bottomrule
\end{tabular}
\caption{Characteristics of the data sets studied in this work.\label{tab:s}}
\end{table}

\textbf{Stochastic Availability:} We make values available by the following scheme over continuous features $z_i \in \spaceXz$:
\begin{align}
    \distp(A_i = 0|z_i) = \text{sigmoid}\left(\lambda_i (z_i - \bar{z_i}) \right) = \frac{1}{1+\exp(-\lambda_i(z_i - \bar{z_i}))},
\end{align}
where we denote the empirical feature mean by $\bar{z_i}$ and $\lambda_i \in \mathbb{R}$ denotes a parameter that specifies how quickly the probability of unavailability ($A_1=0)$ increases with higher feature values (for positive values of $\lambda_i$). For negative values of $\lambda_i$, values of the feature that are lower than the mean are more likely to be unavailable. We chose $\lambda_i$ such that values which were negatively influcing the prediction were more likely to be missing. We show the probabilities curves used of the feature distribution with the corresponding values of $\lambda_i$ in \Cref{fig:datasets_missingness}.

\textbf{Adversarial Availability:} We also experiment with adversarial sharing decision as discussed in the paper. To this end, we first train a full feature model (with no missing data) and a base feature model. We then modify the dataset and drop all optional feature values where the full feature model would lead to a lower regression score or chance of the positive outcome and retrain the corresponding classifiers on this dataset. As a final check, we replace all \OFFIDA prediction that are higher than the base model's predictions by the base model's prediction to arrive at the aformentioned bonus system.

\textbf{Models.} We use standard models from the \texttt{sklearn} library \citep{scikit-learn}. Across all experiments, we used these models with the following parameters:

\begin{center}
\begin{tabular}{rc}
    \toprule
    model & parameters \\
    \midrule
    RandomForestClassifier / RandomForestRegressor & default parameters \\
    ExtraTreesClassifier / ExtraTressRegressor & min\_samples\_split=10 \\
    GradientBoostingClassifier / GradientBoostingRegressor & min\_samples\_split=10 \\
    DecisionTreeClassifier / DecisionTreeRegressor & min\_samples\_split=10 \\
    MLPClassifier / MLPRegressor & hidden\_layer\_sizes= [30,40], max\_iter=500 \\
    \bottomrule
\end{tabular}
\end{center}

% Table with lambda values has been commented out because the values are in Fig. 4
% \begin{comment}
% \begin{table}[h]
%     \centering
%     \caption{Optional features per data set with the corresponding lambda parameter for introducing stochastic availability.}
%     \begin{tabular}{lll}
%     \toprule
%               & Optional feature & $\lambda$  \\ 
%     \hline
%     HELOC     & AverageMInFile   & -0.05   \\
%     COMPAS    & priors\_count    & 1.0      \\
%     Admission & LSAT             & -0.1      \\
%     Diabetes  & Glucose          & 0.1   
%     \end{tabular}
% \end{table}
% \end{comment}

\begin{figure}[h]
    \centering
    \begin{subfigure}[b]{0.45\textwidth}
        \includegraphics[width=\textwidth]{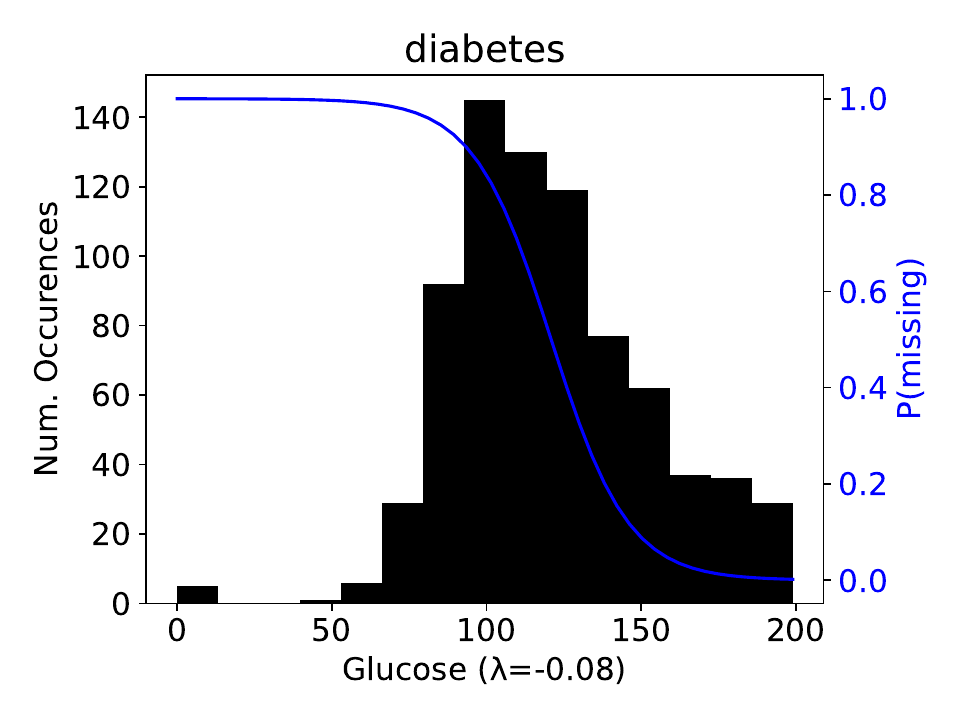}
        \label{fig:diabetes_missingness1}
    \end{subfigure}
    \hfill
    \begin{subfigure}[b]{0.45\textwidth}
        \includegraphics[width=\textwidth]{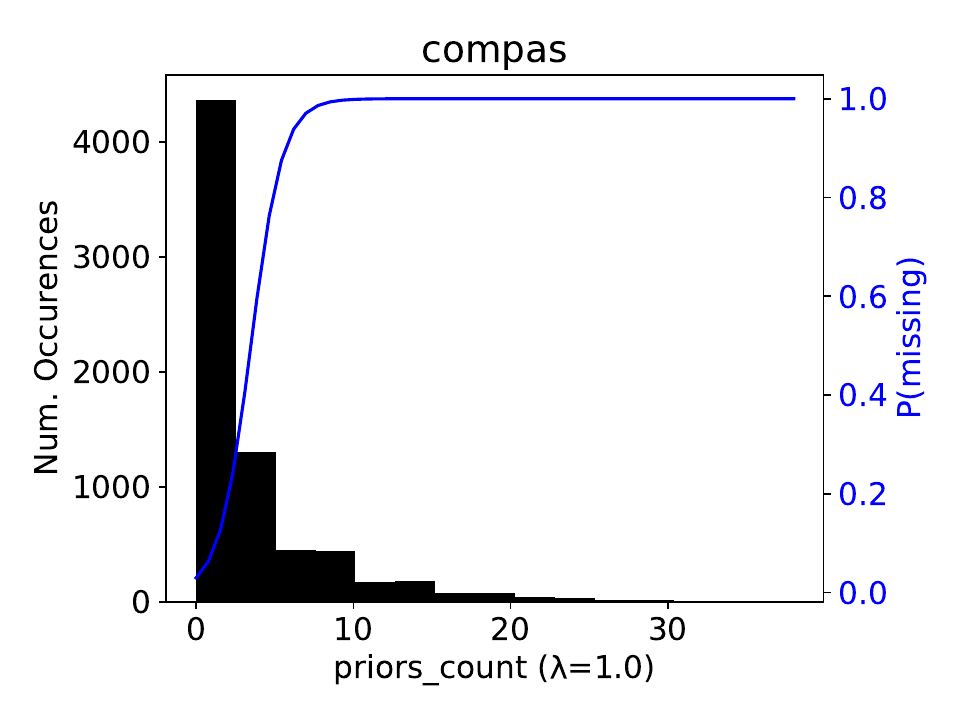}
        \label{fig:compas_missingness}
    \end{subfigure}
    \begin{subfigure}[b]{0.45\textwidth}
        \includegraphics[width=\textwidth]{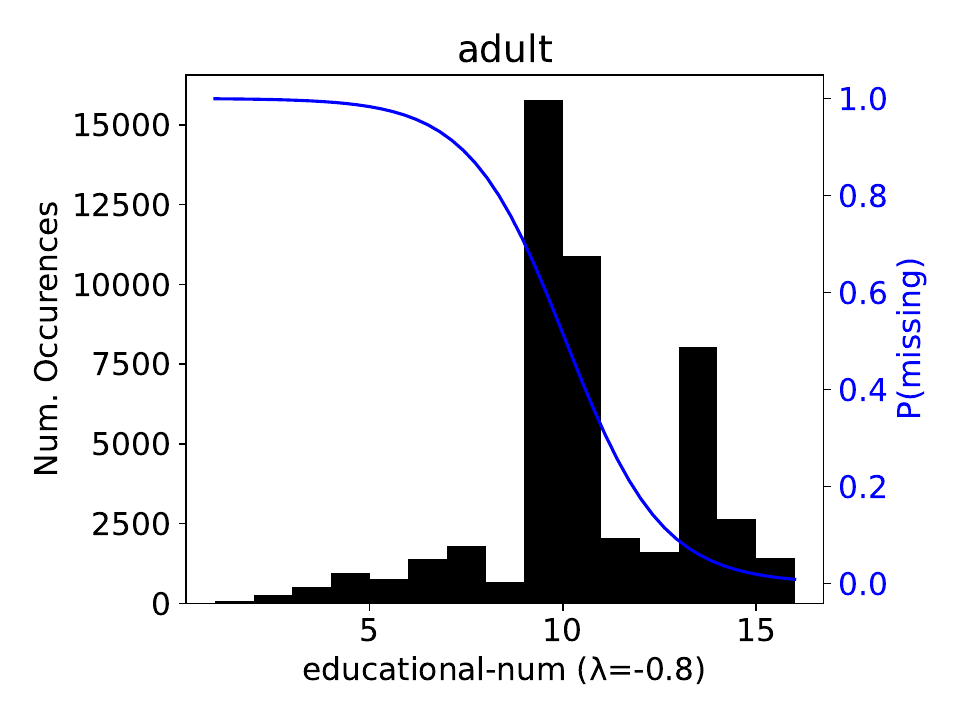}
        \label{fig:adult_missingness}
    \end{subfigure}
    \hfill
    \begin{subfigure}[b]{0.45\textwidth}
        \includegraphics[width=\textwidth]{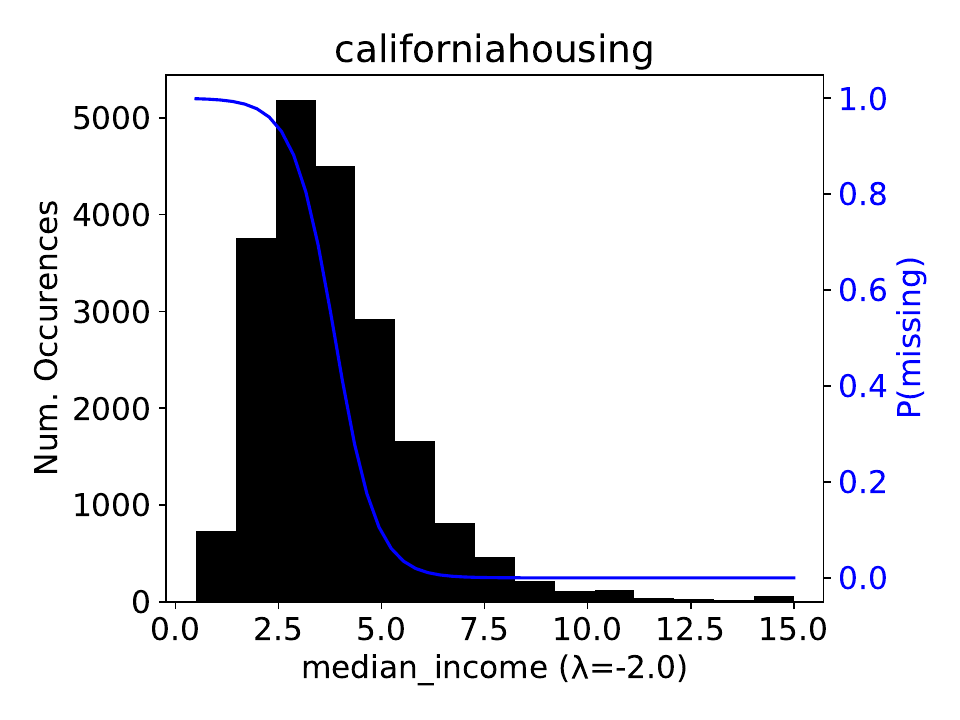}
        \label{fig:californiahousing_missingness}
    \end{subfigure}
    \begin{subfigure}[b]{0.45\textwidth}
        \includegraphics[width=\textwidth]{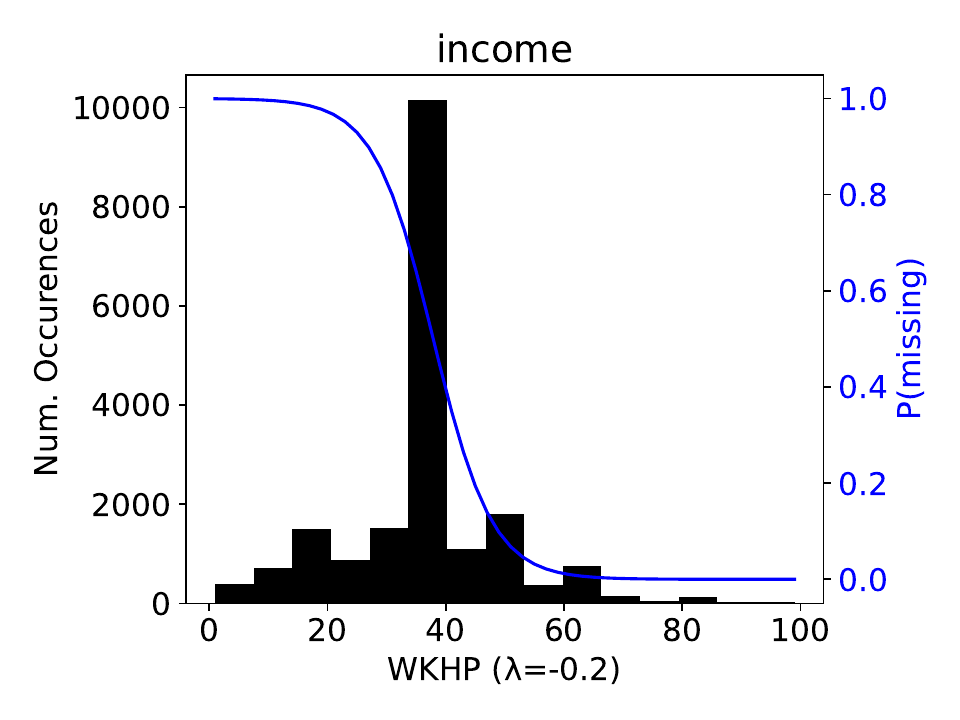}
        \label{fig:income_missingness}
    \end{subfigure}
    \hfill
    \begin{subfigure}[b]{0.45\textwidth}
        \includegraphics[width=\textwidth]{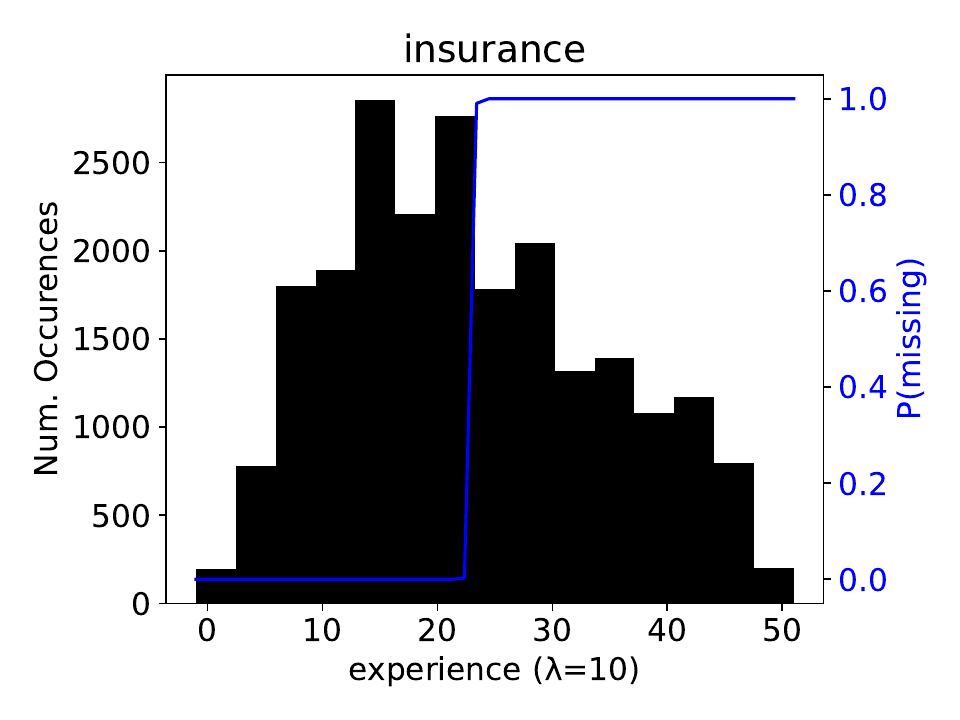}
        \label{fig:insurancemissingness}
    \end{subfigure}
    \caption{Value distribution of the respective optional features per data set and corresponding function $\distp(A_i = 0|z_i)$ with parameter $\lambda$ used to introduce stochastic availability.\label{fig:datasets_missingness}}
\end{figure}

If not stated otherwise, we report averages over 5 runs with a random 80/20 test split. Code to reproduce experiments is provided in the supplementary material and will be publicly released in case of acceptance.

\subsection{Experiment 1: Protecting User consent on real-world data sets}
\label{sec:app_exp1}
This section provides additional results for Experiment 1 (\Cref{tab:fairnessgapreal}) showing that \air is violated. %Tables \ref{tab:minsamplessplit10} -- \ref{tab:gradientboosting} show the results for the same experiment, but with different probabilistic classification models or different hyperparameter settings. 
%While the numbers in \Cref{tab:fairnessgapreal} shows that Non-Penalization is better fulfilled using \OFFIDA, we also show that Incentivization is obeyed.

%This can be seen in \Cref{tab:incentivization1}, where we show that the optional information is used for the group of consenting users. We do so by showing that the performance of the \PUC-model on this group is on par with the unconstrained model. We observe no significant deterioration on all by the ``income'' data set, where the \PUC-score is however still far away from the base model, thus confirming the successful incorporation of optional information in the model. We use 5-fold cross-validation to obtain predictions for every data point conduct five runs with different splits to estimate the averages and standard errors.

\subsubsection{Ablation studies}
\label{sec:app_ablations}
To test the robustness of the results shown in \Cref{tab:fairnessgapreal}, we performed three ablation studies. For all alternative parameters tested, the results are not qualitatively different from the original ones.

\textbf{Imputation Values.} In \Cref{tab:fairnessgapreal}, imputed data points are replaced by zeros. Alternatively, one could also use the mean or the median of the voluntary feature as imputation values, which does not lead to substantial changes as expemplarily shown for classification datasets in \Cref{tab:median_imputation}. We conclude that it is hard to stop Penalization through simple imputation. Note: Our current implementation of the data augmentation strategy is implicitly converts missing values to zero for all missing values, so the results are the same as in the main paper for \OFFIDA.

\textbf{Random forest hyperparameters.} In \Cref{tab:fairnessgapreal}, the default parameters of random forest are used (min\_samples\_split=2, n\_estimators=100, max\_depth=None). The ablation studies with different hyperparameters are shown in Tables \ref{tab:minsamplessplit10} --\ref{tab:estimators500}.

\textbf{Different models.} As an alternative to random forest, we test gradient boosting models (see \Cref{tab:gradientboosting}) and ExtRaTrees by \citet{geurts2006extremely} (see \Cref{tab:ablextratree}). While the extent of change differs to some extent, for every model and hyperparameter configuration, the full feature model uses the information in the sharing decision and the individuals that do not have feature values are rated worse. Occasionally, \PUC models can be non-significantly better than base models, but this is due to statistical errors (as indicated through the standard deviations).

\begin{table}[h]
\centering
\begin{tabular}{rrcccc}
\toprule
task & data & opt. feature  & \Bfm & \PUC & \Ffm \\
\midrule
\texttt{C} & diab. & Glucose & \res{24.75\%}{1.78}  & \textcolor{black}{\res{20.22\%}{2.01}} & \textcolor{black}{\res{19.90\%}{2.35}}\\
\texttt{C} & compas & \#priors & \res{42.02\%}{0.20}  & \textcolor{black}{\res{36.89\%}{0.42}} & \textcolor{black}{\res{36.81\%}{0.51}}\\
\texttt{C} & adult & edu-num & \res{18.75\%}{0.08}  & \textcolor{black}{\res{17.95\%}{0.07}} & \textcolor{black}{\res{17.85\%}{0.12}}\\
\midrule
\texttt{R} & income & WKHP & \res{63.56}{1.08} & \textcolor{black}{\res{54.66}{0.83}} & \textcolor{black}{\res{56.22}{1.13}}\\
\texttt{R} & calif. & m\_income & \res{12.76}{0.11} & \textcolor{black}{\res{8.51}{0.16}} & \textcolor{black}{\res{8.60}{0.17}}\\
\texttt{R} & insurance & experience & \res{245.00}{0.47} & \textcolor{black}{\res{223.53}{0.45}} & \textcolor{black}{\res{230.95}{0.34}}\\
\bottomrule
\end{tabular}
\caption{\textbf{Performance for sharers is maintained with \OFFIDA.} For the setup corresponding to \Cref{tab:fairnessgapreal}, we montior performance measures for the subgroup of sharers. We report missclassification rate (1-Acc) for classification task and MSE ($\times 100$) for regression tasks. We show that the performance in this group is close to the unconstrained model, an indication that their optional information is used.\label{tab:incentivization1}}
\end{table}

\begin{table}[h]
\centering
    \begin{tabular}{cccccc}
    \toprule
    task & data & opt. feature  & base model & imputed & \texttt{\OFFIDA}\\
    \cmidrule(lr){1-3} \cmidrule(lr){4-4} \cmidrule(lr){5-5}\cmidrule(lr){6-6} 
    \texttt{C} & diab. & Glucose & \res{29.30\%}{0.62}  & \res{25.83\%}{0.41} & \res{26.95\%}{0.82}\\
    \texttt{C} & compas & \#priors & \res{42.89\%}{0.10}  & \res{38.51\%}{0.59} & \res{39.55\%}{0.37}\\
    \texttt{C} & adult & edu-num & \res{16.05\%}{0.03}  & \res{15.38\%}{0.11} & \res{15.62\%}{0.09}\\
    \midrule
    \texttt{R} & income & WKHP & \res{85.09}{0.12} & \res{79.76}{0.47} & \res{81.52}{0.28}\\
    \texttt{R} & calif. & m\_income & \res{13.98}{0.06} & \res{11.90}{0.15} & \res{12.55}{0.05}\\
    \texttt{R} & insurance & experience & \res{262.43}{0.21} & \res{249.31}{0.27} & \res{254.84}{0.13}\\
    \bottomrule
    \end{tabular}
    \caption{Costs for \PUC with non-adversarial sharing decisions. Otherwise the setup is equivalent to Table 4a.\label{tab:nonadversarialcosts}}
\end{table}

\begin{table}[h]
\centering
    \begin{tabular}{cccccc}
    \toprule
    task & data & opt. feature  & base model & imputed & \texttt{\OFFIDA}\\\midrule
    \multicolumn{6}{c}{Non-Adversarial Sharing} \\
    \midrule
    \texttt{C} & diab. & Glucose & \res{23.96}{0.11}  & \res{20.35}{0.43} & \res{21.85}{0.69}\\
\texttt{C} & compas & \#priors & \res{40.85}{0.06}  & \res{34.88}{0.54} & \res{36.85}{0.21}\\
\texttt{C} & adult & edu-num & \res{11.90}{0.03}  & \res{11.18}{0.07} & \res{11.10}{0.05}\\
\texttt{C} & water & oxygen. dem. & \res{3.97}{0.42}  & \res{3.47}{0.25} & \res{3.19}{0.25}\\
\texttt{C} & colic & abdom. app. & \res{12.07}{0.31}  & \res{9.08}{0.30} & \res{9.13}{0.28}\\\midrule
    \multicolumn{6}{c}{Adversarial Sharing} \\\midrule
    \texttt{C} & diab. & Glucose & \res{23.96}{0.11}  & \res{18.25}{0.85} & \res{19.91}{0.46}\\
    \texttt{C} & compas & \#priors & \res{40.85}{0.06}  & \res{32.03}{0.35} & \res{34.86}{0.26}\\
    \texttt{C} & adult & edu-num & \res{11.90}{0.03}  & \res{10.26}{0.05} & \res{10.57}{0.03}\\
    \bottomrule
    \end{tabular}
    \caption{Costs for \PUC when using 100$\times$(1-ROCScores) as cost functions for the classification models instead of accuray. Setup as in Table 4a.\label{tab:costsrocsinglefeature}}
\end{table}

%%%%% IMPUTATION ABLATION
\begin{table}[h]
\begin{subtable}{.44\textwidth}
\centering
\adjustbox{max width=\columnwidth}{
\setlength{\tabcolsep}{2pt}
\begin{tabular}{cccccc}
\multicolumn{6}{r}{\texttt{C}: (1-Acc)$\times$100, \texttt{R}: MSE$\times$100}\\
\toprule
task & data & opt. feature  & base model & \Ffm & \texttt{\OFFIDA}\\
\cmidrule(lr){1-3} \cmidrule(lr){4-4} \cmidrule(lr){5-5}\cmidrule(lr){6-6} 
\texttt{C} & diab. & Glucose & \res{35.00\%}{1.57}  & \textcolor{BrickRed}{\res{32.24\%}{1.67}} & \textcolor{Green}{\res{34.67\%}{1.07}}\\
\texttt{C} & compas & \#priors & \res{44.55\%}{0.45}  & \textcolor{BrickRed}{\res{41.84\%}{0.96}} & \textcolor{Green}{\res{44.56\%}{0.51}}\\
\texttt{C} & adult & edu-num & \res{13.40\%}{0.13}  & \textcolor{BrickRed}{\res{13.02\%}{0.27}} & \textcolor{Green}{\res{13.37\%}{0.20}}\\
\midrule
\texttt{R} & income & WKHP & \res{109.09}{1.05} & \textcolor{BrickRed}{\res{107.80}{1.09}} & \textcolor{Green}{\res{110.12}{1.27}}\\
\texttt{R} & calif. & m\_income & \res{17.83}{0.25} & \textcolor{BrickRed}{\res{16.41}{0.34}} & \textcolor{Green}{\res{19.04}{0.18}}\\
\texttt{R} & insurance & experience & \res{282.99}{0.78} & \textcolor{BrickRed}{\res{278.27}{0.46}} & \textcolor{Green}{\res{284.88}{0.93}}\\
\bottomrule
\end{tabular}}
\caption{Corresponding to \Cref{tab:A}, costs, mean imputation.}
\end{subtable}
\hfill
\begin{subtable}{.54\textwidth}
\adjustbox{max width=\columnwidth}{
\setlength{\tabcolsep}{2pt}
\begin{tabular}{cccccccc}
\toprule
& & & &  \multicolumn{2}{c}{\Ffm} &  \multicolumn{2}{c}{\OFFIDA}\\
\cmidrule(lr){5-6}\cmidrule(lr){7-8} 
 task & data & optional & \Bfm & pred. & change & pred. & change\\
\cmidrule(lr){1-3} \cmidrule(lr){4-4} \cmidrule(lr){5-6}\cmidrule(lr){7-8} 
\texttt{C} & diab. & Glucose & 64.17\% & 51.17\% & \textcolor{BrickRed}{\res{-13.00\%}{3.51}} & 63.12\% & \textcolor{Green}{\res{-1.05\%}{2.92}} \\
\texttt{C} & compas & \#priors & 51.39\% & 33.77\% & \textcolor{BrickRed}{\res{-17.63\%}{0.84}} & 51.18\% & \textcolor{Green}{\res{-0.21\%}{0.14}} \\
\texttt{C} & adult & edu-num & 13.77\% & 11.35\% & \textcolor{BrickRed}{\res{-2.42\%}{0.16}} & 13.77\% & \textcolor{Green}{\res{0.01\%}{0.03}} \\
\midrule
\texttt{R} & income & WKHP & 100.0\% & 81.5\% & \textcolor{BrickRed}{\res{-18.5\%}{0.48}} & 101.4\% & \textcolor{Green}{\res{1.4\%}{0.18}}\\
\texttt{R} & insurance & experience & 100.0\% & 94.9\% & \textcolor{BrickRed}{\res{-5.1\%}{0.10}} & 100.1\% & \textcolor{Green}{\res{0.1\%}{0.05}}\\
\texttt{R} & calif. & m\_income & 100.0\% & 95.3\% & \textcolor{BrickRed}{\res{-4.7\%}{0.28}} & 104.2\% & \textcolor{Green}{\res{4.2\%}{0.42}}\\
\bottomrule
\end{tabular}}
\caption{Corresponding to \Cref{tab:B}, absolute predictions, mean imputation.}
\end{subtable}
\begin{subtable}{.44\textwidth}
\centering
\adjustbox{max width=\columnwidth}{
\setlength{\tabcolsep}{2pt}
\begin{tabular}{cccccc}
\multicolumn{6}{r}{\texttt{C}: (1-Acc)$\times$100, \texttt{R}: MSE$\times$100}\\
\toprule
task & data & opt. feature  & base model & \Ffm & \texttt{\OFFIDA}\\
\cmidrule(lr){1-3} \cmidrule(lr){4-4} \cmidrule(lr){5-5}\cmidrule(lr){6-6} 
\texttt{C} & diab. & Glucose & \res{35.00\%}{1.57}  & \textcolor{BrickRed}{\res{32.22\%}{1.18}} & \textcolor{Green}{\res{34.67\%}{1.07}}\\
\texttt{C} & compas & \#priors & \res{44.55\%}{0.45}  & \textcolor{BrickRed}{\res{41.70\%}{0.97}} & \textcolor{Green}{\res{44.56\%}{0.51}}\\
\texttt{C} & adult & edu-num & \res{13.40\%}{0.13}  & \textcolor{BrickRed}{\res{12.99\%}{0.22}} & \textcolor{Green}{\res{13.37\%}{0.20}}\\
\midrule
\texttt{R} & income & WKHP & \res{109.09}{1.05} & \textcolor{BrickRed}{\res{107.70}{1.18}} & \textcolor{Green}{\res{110.12}{1.27}}\\
\texttt{R} & calif. & m\_income & \res{17.83}{0.25} & \textcolor{BrickRed}{\res{16.28}{0.31}} & \textcolor{Green}{\res{19.04}{0.18}}\\
\texttt{R} & insurance & experience & \res{282.99}{0.78} & \textcolor{BrickRed}{\res{278.29}{0.64}} & \textcolor{Green}{\res{284.88}{0.93}}\\
\bottomrule
\end{tabular}}
\caption{Corresponding to \Cref{tab:A}, costs, median imputation.}
\end{subtable}
\hfill
\begin{subtable}{.54\textwidth}
\adjustbox{max width=\columnwidth}{
\setlength{\tabcolsep}{2pt}
\begin{tabular}{cccccccc}
\toprule
& & & &  \multicolumn{2}{c}{\Ffm} &  \multicolumn{2}{c}{\OFFIDA}\\
\cmidrule(lr){5-6}\cmidrule(lr){7-8} 
 task & data & optional & \Bfm & pred. & change & pred. & change\\
\cmidrule(lr){1-3} \cmidrule(lr){4-4} \cmidrule(lr){5-6}\cmidrule(lr){7-8} 
\texttt{C} & diab. & Glucose & 64.17\% & 50.83\% & \textcolor{BrickRed}{\res{-13.34\%}{4.10}} & 63.12\% & \textcolor{Green}{\res{-1.05\%}{2.92}} \\
\texttt{C} & compas & \#priors & 51.39\% & 33.53\% & \textcolor{BrickRed}{\res{-17.86\%}{0.92}} & 51.18\% & \textcolor{Green}{\res{-0.21\%}{0.14}} \\
\texttt{C} & adult & edu-num & 13.77\% & 11.29\% & \textcolor{BrickRed}{\res{-2.48\%}{0.17}} & 13.77\% & \textcolor{Green}{\res{0.01\%}{0.03}} \\
\midrule
\texttt{R} & income & WKHP & 100.0\% & 81.9\% & \textcolor{BrickRed}{\res{-18.1\%}{0.60}} & 101.4\% & \textcolor{Green}{\res{1.4\%}{0.18}}\\
\texttt{R} & insurance & experience & 100.0\% & 94.9\% & \textcolor{BrickRed}{\res{-5.1\%}{0.09}} & 100.1\% & \textcolor{Green}{\res{0.1\%}{0.05}}\\
\texttt{R} & calif. & m\_income & 100.0\% & 94.9\% & \textcolor{BrickRed}{\res{-5.1\%}{0.52}} & 104.2\% & \textcolor{Green}{\res{4.2\%}{0.42}}\\
\bottomrule
\end{tabular}}
\caption{Corresponding to \Cref{tab:B}, absolute predictions, median imputation.}
\end{subtable}
    \vspace{0.1cm} 
    \caption{Same setup as  \Cref{tab:fairnessgapreal}  using \emph{mean imputation} (upper line) and \emph{median imputation} (lower line). The differencees between the two imputation techniques are minimal. \label{tab:median_imputation}}
\end{table}

\begin{table}[h]
    \centering
\begin{subtable}{.44\textwidth}
\centering
\adjustbox{max width=\columnwidth}{
\setlength{\tabcolsep}{2pt}
\begin{tabular}{cccccc}
\multicolumn{6}{r}{\texttt{C}: (1-Acc)$\times$100, \texttt{R}: MSE$\times$100}\\
\toprule
task & data & opt. feature  & base model & \Ffm & \texttt{\OFFIDA}\\
\cmidrule(lr){1-3} \cmidrule(lr){4-4} \cmidrule(lr){5-5}\cmidrule(lr){6-6} 
\texttt{C} & diab. & Glucose & \res{34.38\%}{1.60}  & \textcolor{BrickRed}{\res{33.52\%}{1.30}} & \textcolor{Green}{\res{34.25\%}{1.01}}\\
\texttt{C} & compas & \#priors & \res{42.53\%}{0.40}  & \textcolor{BrickRed}{\res{39.21\%}{0.60}} & \textcolor{Green}{\res{42.92\%}{0.43}}\\
\texttt{C} & adult & edu-num & \res{12.04\%}{0.11}  & \textcolor{BrickRed}{\res{11.75\%}{0.19}} & \textcolor{Green}{\res{11.99\%}{0.19}}\\
\midrule
\texttt{R} & income & WKHP & \res{104.62}{0.56} & \textcolor{BrickRed}{\res{103.15}{0.80}} & \textcolor{Green}{\res{105.85}{0.60}}\\
\texttt{R} & calif. & m\_income & \res{17.84}{0.23} & \textcolor{BrickRed}{\res{16.15}{0.50}} & \textcolor{Green}{\res{19.11}{0.46}}\\
\texttt{R} & insurance & experience & \res{260.05}{0.29} & \textcolor{BrickRed}{\res{256.41}{0.19}} & \textcolor{Green}{\res{262.31}{0.54}}\\
\bottomrule
\end{tabular}}
\caption{Corresponding to \Cref{tab:A} (costs).}
\end{subtable}
\hfill
\begin{subtable}{.54\textwidth}
\adjustbox{max width=\columnwidth}{
\setlength{\tabcolsep}{2pt}
\begin{tabular}{cccccccc}
\toprule
& & & &  \multicolumn{2}{c}{\Ffm} &  \multicolumn{2}{c}{\OFFIDA}\\
\cmidrule(lr){5-6}\cmidrule(lr){7-8} 
 task & data & optional & \Bfm & pred. & change & pred. & change\\
\cmidrule(lr){1-3} \cmidrule(lr){4-4} \cmidrule(lr){5-6}\cmidrule(lr){7-8} 
\texttt{C} & diab. & Glucose & 64.86\% & 52.13\% & \textcolor{BrickRed}{\res{-12.73\%}{2.15}} & 66.11\% & \textcolor{Green}{\res{1.25\%}{1.89}} \\
\texttt{C} & compas & \#priors & 51.89\% & 29.92\% & \textcolor{BrickRed}{\res{-21.97\%}{0.97}} & 52.11\% & \textcolor{Green}{\res{0.22\%}{0.57}} \\
\texttt{C} & adult & edu-num & 12.08\% & 9.49\% & \textcolor{BrickRed}{\res{-2.59\%}{0.06}} & 12.18\% & \textcolor{Green}{\res{0.10\%}{0.09}} \\
\midrule
\texttt{R} & income & WKHP & 100.0\% & 81.4\% & \textcolor{BrickRed}{\res{-18.6\%}{0.36}} & 101.3\% & \textcolor{Green}{\res{1.3\%}{0.31}}\\
\texttt{R} & insurance & experience & 100.0\% & 94.8\% & \textcolor{BrickRed}{\res{-5.2\%}{0.06}} & 100.2\% & \textcolor{Green}{\res{0.2\%}{0.07}}\\
\texttt{R} & calif. & m\_income & 100.0\% & 94.6\% & \textcolor{BrickRed}{\res{-5.4\%}{1.00}} & 104.1\% & \textcolor{Green}{\res{4.1\%}{0.75}}\\
\bottomrule
\end{tabular}}
\caption{Corresponding to \Cref{tab:B} (absolute predictions).}
\end{subtable}
\vspace{0.1cm}
\caption{\textbf{\air is violated by \ffms}. Same setup as \Cref{tab:fairnessgapreal} using a \emph{Random Forest model} with $min\_samples\_split=10$. \label{tab:minsamplessplit10}}
\end{table}

\begin{table}[h]
    \centering
\begin{subtable}{.44\textwidth}
\centering
\adjustbox{max width=\columnwidth}{
\setlength{\tabcolsep}{2pt}
\begin{tabular}{cccccc}
\multicolumn{6}{r}{\texttt{C}: (1-Acc)$\times$100, \texttt{R}: MSE$\times$100}\\
\toprule
task & data & opt. feature  & base model & \Ffm & \texttt{\OFFIDA}\\
\cmidrule(lr){1-3} \cmidrule(lr){4-4} \cmidrule(lr){5-5}\cmidrule(lr){6-6} 
\texttt{C} & diab. & Glucose & \res{36.49\%}{0.45}  & \textcolor{BrickRed}{\res{34.27\%}{0.99}} & \textcolor{Green}{\res{37.87\%}{0.95}}\\
\texttt{C} & compas & \#priors & \res{43.99\%}{0.70}  & \textcolor{BrickRed}{\res{37.34\%}{0.45}} & \textcolor{Green}{\res{50.43\%}{0.96}}\\
\texttt{C} & adult & edu-num & \res{13.29\%}{0.23}  & \textcolor{BrickRed}{\res{13.96\%}{0.13}} & \textcolor{Green}{\res{13.18\%}{0.10}}\\
\midrule
\texttt{R} & income & WKHP & \res{117.24}{0.73} & \textcolor{BrickRed}{\res{115.76}{0.79}} & \textcolor{Green}{\res{123.08}{1.01}}\\
\texttt{R} & calif. & m\_income & \res{26.08}{0.10} & \textcolor{BrickRed}{\res{20.29}{0.21}} & \textcolor{Green}{\res{25.51}{0.12}}\\
\texttt{R} & insurance & experience & \res{251.99}{0.13} & \textcolor{BrickRed}{\res{251.21}{0.12}} & \textcolor{Green}{\res{258.73}{0.16}}\\
\bottomrule
\end{tabular}}
\caption{Corresponding to \Cref{tab:A} (costs).}
\end{subtable}
\hfill
\begin{subtable}{.54\textwidth}
\adjustbox{max width=\columnwidth}{
\setlength{\tabcolsep}{2pt}
\begin{tabular}{cccccccc}
\toprule
& & & &  \multicolumn{2}{c}{\Ffm} &  \multicolumn{2}{c}{\OFFIDA}\\
\cmidrule(lr){5-6}\cmidrule(lr){7-8} 
 task & data & optional & \Bfm & pred. & change & pred. & change\\
\cmidrule(lr){1-3} \cmidrule(lr){4-4} \cmidrule(lr){5-6}\cmidrule(lr){7-8} 
\texttt{C} & diab. & Glucose & 73.66\% & 58.45\% & \textcolor{BrickRed}{\res{-15.22\%}{4.23}} & 79.07\% & \textcolor{Green}{\res{5.41\%}{2.30}} \\
\texttt{C} & compas & \#priors & 65.82\% & 27.97\% & \textcolor{BrickRed}{\res{-37.85\%}{4.26}} & 79.98\% & \textcolor{Green}{\res{14.16\%}{0.75}} \\
\texttt{C} & adult & edu-num & 2.34\% & 1.54\% & \textcolor{BrickRed}{\res{-0.80\%}{0.41}} & 2.54\% & \textcolor{Green}{\res{0.20\%}{0.27}} \\
\midrule
\texttt{R} & income & WKHP & 100.0\% & 75.5\% & \textcolor{BrickRed}{\res{-24.5\%}{0.78}} & 108.6\% & \textcolor{Green}{\res{8.6\%}{0.38}}\\
\texttt{R} & insurance & experience & 100.0\% & 94.6\% & \textcolor{BrickRed}{\res{-5.4\%}{0.09}} & 103.5\% & \textcolor{Green}{\res{3.5\%}{0.04}}\\
\texttt{R} & calif. & m\_income & 100.0\% & 83.9\% & \textcolor{BrickRed}{\res{-16.1\%}{0.30}} & 99.4\% & \textcolor{Green}{\res{-0.6\%}{0.21}}\\
\bottomrule
\end{tabular}}
\caption{Corresponding to \Cref{tab:B} (absolute predictions).}
\end{subtable}
\caption{\textbf{\air is violated by \ffms}. Same setup as \Cref{tab:fairnessgapreal} using a \emph{Random Forest model} with $max\_depth=4$. \label{tab:maxdepth4}}
\end{table}

\begin{table}[h]
    \centering
\begin{subtable}{.44\textwidth}
\centering
\adjustbox{max width=\columnwidth}{
\setlength{\tabcolsep}{2pt}
\begin{tabular}{cccccc}
\multicolumn{6}{r}{\texttt{C}: (1-Acc)$\times$100, \texttt{R}: MSE$\times$100}\\
\toprule
task & data & opt. feature  & base model & \Ffm & \texttt{\OFFIDA}\\
\cmidrule(lr){1-3} \cmidrule(lr){4-4} \cmidrule(lr){5-5}\cmidrule(lr){6-6} 
\texttt{C} & diab. & Glucose & \res{34.07\%}{0.87}  & \textcolor{BrickRed}{\res{33.61\%}{0.43}} & \textcolor{Green}{\res{33.58\%}{0.92}}\\
\texttt{C} & compas & \#priors & \res{44.64\%}{0.20}  & \textcolor{BrickRed}{\res{41.40\%}{0.61}} & \textcolor{Green}{\res{44.78\%}{0.42}}\\
\texttt{C} & adult & edu-num & \res{13.31\%}{0.06}  & \textcolor{BrickRed}{\res{12.83\%}{0.14}} & \textcolor{Green}{\res{13.31\%}{0.05}}\\
\midrule
\texttt{R} & income & WKHP & \res{107.98}{0.37} & \textcolor{BrickRed}{\res{106.71}{0.32}} & \textcolor{Green}{\res{109.29}{0.54}}\\
\texttt{R} & calif. & m\_income & \res{17.70}{0.10} & \textcolor{BrickRed}{\res{16.11}{0.31}} & \textcolor{Green}{\res{18.93}{0.20}}\\
\texttt{R} & insurance & experience & \res{281.96}{0.12} & \textcolor{BrickRed}{\res{277.03}{0.51}} & \textcolor{Green}{\res{283.79}{0.19}}\\
\bottomrule
\end{tabular}}
\caption{Corresponding to \Cref{tab:A} (costs).}
\end{subtable}
\hfill
\begin{subtable}{.54\textwidth}
\adjustbox{max width=\columnwidth}{
\setlength{\tabcolsep}{2pt}
\begin{tabular}{cccccccc}
\toprule
& & & &  \multicolumn{2}{c}{\Ffm} &  \multicolumn{2}{c}{\OFFIDA}\\
\cmidrule(lr){5-6}\cmidrule(lr){7-8} 
 task & data & optional & \Bfm & pred. & change & pred. & change\\
\cmidrule(lr){1-3} \cmidrule(lr){4-4} \cmidrule(lr){5-6}\cmidrule(lr){7-8} 
\texttt{C} & diab. & Glucose & 63.30\% & 51.63\% & \textcolor{BrickRed}{\res{-11.67\%}{1.20}} & 63.21\% & \textcolor{Green}{\res{-0.09\%}{1.14}} \\
\texttt{C} & compas & \#priors & 51.31\% & 32.62\% & \textcolor{BrickRed}{\res{-18.69\%}{1.27}} & 51.36\% & \textcolor{Green}{\res{0.05\%}{0.43}} \\
\texttt{C} & adult & edu-num & 13.93\% & 11.47\% & \textcolor{BrickRed}{\res{-2.46\%}{0.24}} & 13.91\% & \textcolor{Green}{\res{-0.02\%}{0.09}} \\
\midrule
\texttt{R} & income & WKHP & 100.0\% & 81.4\% & \textcolor{BrickRed}{\res{-18.6\%}{0.46}} & 101.3\% & \textcolor{Green}{\res{1.3\%}{0.12}}\\
\texttt{R} & insurance & experience & 100.0\% & 94.8\% & \textcolor{BrickRed}{\res{-5.2\%}{0.08}} & 100.1\% & \textcolor{Green}{\res{0.1\%}{0.02}}\\
\texttt{R} & calif. & m\_income & 100.0\% & 94.1\% & \textcolor{BrickRed}{\res{-5.9\%}{0.86}} & 103.8\% & \textcolor{Green}{\res{3.8\%}{0.48}}\\
\bottomrule
\end{tabular}}
\caption{Corresponding to \Cref{tab:B} (absolute predictions).}
\end{subtable}
\caption{\textbf{\air is violated by \ffms}. Same setup as \Cref{tab:fairnessgapreal} using a \emph{Random Forest model} with $n\_estimators=500$. \label{tab:estimators500}}
\end{table}

\begin{table}[h]
    \centering
\begin{subtable}{.44\textwidth}
\centering
\adjustbox{max width=\columnwidth}{
\setlength{\tabcolsep}{2pt}
\begin{tabular}{cccccc}
\multicolumn{6}{r}{\texttt{C}: (1-Acc)$\times$100, \texttt{R}: MSE$\times$100}\\
\toprule
task & data & opt. feature  & base model & \Ffm & \texttt{\OFFIDA}\\
\cmidrule(lr){1-3} \cmidrule(lr){4-4} \cmidrule(lr){5-5}\cmidrule(lr){6-6} 
\texttt{C} & diab. & Glucose & \res{34.07\%}{0.87}  & \textcolor{BrickRed}{\res{33.61\%}{0.43}} & \textcolor{Green}{\res{33.58\%}{0.92}}\\
\texttt{C} & compas & \#priors & \res{44.64\%}{0.20}  & \textcolor{BrickRed}{\res{41.40\%}{0.61}} & \textcolor{Green}{\res{44.78\%}{0.42}}\\
\texttt{C} & adult & edu-num & \res{13.31\%}{0.06}  & \textcolor{BrickRed}{\res{12.83\%}{0.14}} & \textcolor{Green}{\res{13.31\%}{0.05}}\\
\midrule
\texttt{R} & income & WKHP & \res{107.98}{0.37} & \textcolor{BrickRed}{\res{106.71}{0.32}} & \textcolor{Green}{\res{109.29}{0.54}}\\
\texttt{R} & calif. & m\_income & \res{17.70}{0.10} & \textcolor{BrickRed}{\res{16.11}{0.31}} & \textcolor{Green}{\res{18.93}{0.20}}\\
\texttt{R} & insurance & experience & \res{281.96}{0.12} & \textcolor{BrickRed}{\res{277.03}{0.51}} & \textcolor{Green}{\res{283.79}{0.19}}\\
\bottomrule
\end{tabular}}
\caption{Corresponding to \Cref{tab:A} (costs).}
\end{subtable}
\hfill
\begin{subtable}{.54\textwidth}
\adjustbox{max width=\columnwidth}{
\setlength{\tabcolsep}{2pt}
\begin{tabular}{cccccccc}
\toprule
& & & &  \multicolumn{2}{c}{\Ffm} &  \multicolumn{2}{c}{\OFFIDA}\\
\cmidrule(lr){5-6}\cmidrule(lr){7-8} 
 task & data & optional & \Bfm & pred. & change & pred. & change\\
\cmidrule(lr){1-3} \cmidrule(lr){4-4} \cmidrule(lr){5-6}\cmidrule(lr){7-8} 
\texttt{C} & diab. & Glucose & 62.80\% & 52.96\% & \textcolor{BrickRed}{\res{-9.84\%}{2.46}} & 63.75\% & \textcolor{Green}{\res{0.95\%}{1.35}} \\
\texttt{C} & compas & \#priors & 62.82\% & 21.29\% & \textcolor{BrickRed}{\res{-41.54\%}{1.69}} & 64.14\% & \textcolor{Green}{\res{1.32\%}{0.22}} \\
\texttt{C} & adult & edu-num & 9.73\% & 5.98\% & \textcolor{BrickRed}{\res{-3.76\%}{0.10}} & 9.31\% & \textcolor{Green}{\res{-0.43\%}{0.07}} \\
\midrule
\texttt{R} & income & WKHP & 100.0\% & 79.6\% & \textcolor{BrickRed}{\res{-20.4\%}{0.13}} & 100.0\% & \textcolor{Green}{\res{0.0\%}{0.09}}\\
\texttt{R} & insurance & experience & 100.0\% & 94.4\% & \textcolor{BrickRed}{\res{-5.6\%}{0.05}} & 101.2\% & \textcolor{Green}{\res{1.2\%}{0.02}}\\
\texttt{R} & calif. & m\_income & 100.0\% & 91.1\% & \textcolor{BrickRed}{\res{-8.9\%}{0.36}} & 102.9\% & \textcolor{Green}{\res{2.9\%}{0.75}}\\
\bottomrule
\end{tabular}}
\caption{Corresponding to \Cref{tab:B} (absolute predictions).}
\end{subtable}
\caption{\textbf{\air is violated by \ffms}. Same setup as \Cref{tab:fairnessgapreal} using a \emph{Gradient Boosting model}. \label{tab:gradientboosting}}
\end{table}

\begin{table}[h]
    \centering
\begin{subtable}{.44\textwidth}
\centering
\adjustbox{max width=\columnwidth}{
\setlength{\tabcolsep}{2pt}
\begin{tabular}{cccccc}
\multicolumn{6}{r}{\texttt{C}: (1-Acc)$\times$100, \texttt{R}: MSE$\times$100}\\
\toprule
task & data & opt. feature  & base model & \Ffm & \texttt{\OFFIDA}\\
\cmidrule(lr){1-3} \cmidrule(lr){4-4} \cmidrule(lr){5-5}\cmidrule(lr){6-6} 
\texttt{C} & diab. & Glucose & \res{35.27\%}{0.89}  & \textcolor{BrickRed}{\res{33.20\%}{1.46}} & \textcolor{Green}{\res{35.68\%}{1.54}}\\
\texttt{C} & compas & \#priors & \res{42.22\%}{0.41}  & \textcolor{BrickRed}{\res{36.69\%}{0.29}} & \textcolor{Green}{\res{42.66\%}{0.56}}\\
\texttt{C} & adult & edu-num & \res{11.25\%}{0.09}  & \textcolor{BrickRed}{\res{11.37\%}{0.07}} & \textcolor{Green}{\res{11.27\%}{0.12}}\\
\midrule
\texttt{R} & income & WKHP & \res{104.79}{0.72} & \textcolor{BrickRed}{\res{100.90}{0.75}} & \textcolor{Green}{\res{106.02}{0.66}}\\
\texttt{R} & calif. & m\_income & \res{16.99}{0.10} & \textcolor{BrickRed}{\res{15.50}{0.22}} & \textcolor{Green}{\res{17.74}{0.34}}\\
\texttt{R} & insurance & experience & \res{243.33}{0.12} & \textcolor{BrickRed}{\res{242.30}{0.12}} & \textcolor{Green}{\res{246.22}{0.12}}\\
\bottomrule
\end{tabular}}
\caption{Corresponding to \Cref{tab:A} (costs).}
\end{subtable}
\hfill
\begin{subtable}{.54\textwidth}
\adjustbox{max width=\columnwidth}{
\setlength{\tabcolsep}{2pt}
\begin{tabular}{cccccccc}
\toprule
& & & &  \multicolumn{2}{c}{\Ffm} &  \multicolumn{2}{c}{\OFFIDA}\\
\cmidrule(lr){5-6}\cmidrule(lr){7-8} 
 task & data & optional & \Bfm & pred. & change & pred. & change\\
\cmidrule(lr){1-3} \cmidrule(lr){4-4} \cmidrule(lr){5-6}\cmidrule(lr){7-8} 
\texttt{C} & diab. & Glucose & 71.27\% & 48.73\% & \textcolor{BrickRed}{\res{-22.54\%}{5.84}} & 68.78\% & \textcolor{Green}{\res{-2.49\%}{1.27}} \\
\texttt{C} & compas & \#priors & 53.53\% & 29.53\% & \textcolor{BrickRed}{\res{-24.00\%}{1.16}} & 53.39\% & \textcolor{Green}{\res{-0.14\%}{0.18}} \\
\texttt{C} & adult & edu-num & 12.27\% & 9.58\% & \textcolor{BrickRed}{\res{-2.69\%}{0.27}} & 12.33\% & \textcolor{Green}{\res{0.06\%}{0.06}} \\
\midrule
\texttt{R} & income & WKHP & 100.0\% & 80.1\% & \textcolor{BrickRed}{\res{-19.9\%}{0.60}} & 101.0\% & \textcolor{Green}{\res{1.0\%}{0.08}}\\
\texttt{R} & insurance & experience & 100.0\% & 94.6\% & \textcolor{BrickRed}{\res{-5.4\%}{0.07}} & 100.1\% & \textcolor{Green}{\res{0.1\%}{0.02}}\\
\texttt{R} & calif. & m\_income & 100.0\% & 90.9\% & \textcolor{BrickRed}{\res{-9.1\%}{0.55}} & 104.0\% & \textcolor{Green}{\res{4.0\%}{0.28}}\\
\bottomrule
\end{tabular}}
\caption{Corresponding to \Cref{tab:B} (absolute predictions).}
\end{subtable}
\caption{\textbf{\air is violated by \ffms}. Same setup as \Cref{tab:fairnessgapreal} using a \emph{Extra Trees model} \label{tab:ablextratree}}
\end{table}

\subsection{Experiment 2: Validating Non-Degradation and costs of fairness with respect to optional information}
This section contains additional details on the experiments leading up to \Cref{tab:costsoffairness}.
\label{sec:app_exp3real}

\textbf{Single optional feature}. We first investigate the performance of the models in the setup corresponding to \Cref{tab:fairnessgapreal}, i.e., with only a single optional feature. In \Cref{tab:nonadversarialcosts} we show the cost setup when non-stategic sharing decisions are taking, which leads to qualitatively equivalent results as in the main paper. \Cref{tab:costsrocsinglefeature} shows the results for the classification models when using the area under the 1-ROC-curve as a cost function.

Having verified these results for a single feature, we now continue with the more challenging setup of multiple optionality.

\textbf{Introducing multiple optionality.} For the real data experiment, we apply the following preprocessing steps to induce stochastic availability:
\begin{itemize}
    \item We identify the most discriminative numerical features by dropping each feature from the data set and reporting the decline in predictive performance of a model trained without the feature with respect to a model trained on all features. We rank the features starting with the one resulting in the highest performance loss.
    \item We select the $r$ most discriminative features, such that on average, each subset of missing pattern has at least $150$ samples out of the initial data set size of $N$ to be fitted with, i.e.,
    \begin{align*}
        r = \inf \left\{r' \in \mathbb{N}: \frac{N}{2^r} > 150 \right\}.
    \end{align*}
    \item We do not consider numerical features were the relation to the label is not clear (i.e., is there a positive or negative correlation). The optional features are listed in \Cref{tab:app_offgapfeatures}.
    \item We independently induce stochastic availability into each feature using the sigmoidal strategy. We use a $\lambda_i = \pm \frac{1}{\sqrt{\Var{[f_i]}}}$, which is effectively equivalent to applying a sigmoid over normalized feature values. The signs are determined by the context such that negative indicators are more likely to be not provided and are also reported in the \Cref{tab:app_offgapfeatures}.

\end{itemize}
We show the corresponding results of \Cref{tab:multiplefeatures} using 1-ROC as cost function in \Cref{tab:app_costoffairnessmultiple_roc}. We provide ablations with the two other models in \Cref{tab:multilegbt} and \Cref{tab:multipleextra}

\begin{table}[h]
\centering
\begin{tabular}{cccccrc}
\toprule
& & \multicolumn{4}{c}{Fair models} & \Ffm \\ 
\cmidrule(lr){3-6}\cmidrule(lr){7-7}
task & data (\# opt.) & \Bfm & \OFFIDA (f) & \OFFIDA (e) &  ($\times$) & zero-imputed\\
\cmidrule(r){1-2} \cmidrule(lr){3-3}\cmidrule(lr){4-4} \cmidrule(lr){5-6}  \cmidrule(lr){7-7}

\texttt{C} &diab. (2) & \res{73.14}{2.59} & \res{77.13}{3.67} & \bres{77.22}{3.92} & 2.3 & \res{78.42}{2.61}\\
\texttt{C} &compas (5) & \res{61.32}{0.92} & \res{61.90}{0.96} & \bres{62.32}{0.86} & 7.6 & \res{62.69}{1.58}\\
\texttt{C} &adult (5) & \res{84.90}{0.46} & \res{90.39}{0.35} & \bres{89.68}{0.33} & 7.4 & \res{90.57}{0.21}\\
\bottomrule
\end{tabular}
\caption{\textbf{\PUC-compliant models improve predictive performance}. Same setup as in to \Cref{tab:multiplefeatures}, but in this case we use \emph{ROC-AUC as the performance metric}. A higher ROC-AUC is preferable.\label{tab:app_costoffairnessmultiple_roc}}
\end{table}
%Furthermore, we would like to stress that in our theory, optimal estimators of the conditional mean are assumed, so even if statistically significant, the results would not be contradictory.

\begin{table}[h]
\centering
\begin{tabular}{cccccrc}
\toprule
& & \multicolumn{4}{c}{Fair models} & \Ffm \\ 
\cmidrule(lr){3-6}\cmidrule(lr){7-7}
task & data (\# opt.) & \Bfm & \OFFIDA (f) & \OFFIDA (e) &  ($\times$) & zero-imputed\\
\cmidrule(r){1-2} \cmidrule(lr){3-3}\cmidrule(lr){4-4} \cmidrule(lr){5-6}  \cmidrule(lr){7-7}
\texttt{C} &diab. (2) & \res{29.87}{2.25} & \res{28.70}{2.37} & \bres{28.18}{2.74} & 2.3 & \res{27.14}{2.67}\\
\texttt{C} &compas (5) & \res{40.78}{0.63} & \bres{37.71}{0.63} & \res{37.79}{0.90} & 7.6 & \res{35.62}{0.60}\\
\texttt{C} &adult (5) & \res{17.84}{0.41} & \res{13.43}{0.49} & \bres{13.43}{0.48} & 7.4 & \res{13.31}{0.45}\\
\midrule
\texttt{R} & calif. (4) & \res{11.01}{1.89} & \res{9.45}{0.19} & \bres{9.32}{0.34} & 5.1 & \res{9.04}{0.17}\\
\texttt{R} & income (3) & \res{46.31}{2.02} & \res{45.00}{2.42} & \bres{44.28}{1.86} & 3.4 & \res{41.89}{1.52}\\
\texttt{R} & insurance (3) & \res{230.00}{0.72} & \res{212.30}{1.73} & \bres{211.27}{2.13} & 3.2 & \res{210.72}{1.55}\\
\bottomrule
\end{tabular}
\caption{\textbf{\PUC-compliant models improve predictive performance}. Same setup as in to \Cref{tab:multiplefeatures}, but in this case we use \emph{Gradient Boosted Decision Trees}.}
\label{tab:multilegbt}
\end{table}

\begin{table}[h]
\centering
\begin{tabular}{cccccrc}
\toprule
& & \multicolumn{4}{c}{Fair models} & \Ffm \\ 
\cmidrule(lr){3-6}\cmidrule(lr){7-7}
task & data (\# opt.) & \Bfm & \OFFIDA (f) & \OFFIDA (e) &  ($\times$) & zero-imputed\\
\cmidrule(r){1-2} \cmidrule(lr){3-3}\cmidrule(lr){4-4} \cmidrule(lr){5-6}  \cmidrule(lr){7-7}
\texttt{C} &diab. (2) & \res{27.79}{4.22} & \res{28.18}{2.48} & \bres{27.92}{2.63} & 2.3 & \res{26.88}{3.50}\\
\texttt{C} &compas (5) & \res{40.83}{0.53} & \res{39.82}{0.75} & \bres{39.33}{1.38} & 7.7 & \res{39.74}{1.39}\\
\texttt{C} &adult (5) & \res{18.00}{0.37} & \bres{15.21}{0.52} & \res{15.31}{0.51} & 7.4 & \res{15.15}{0.37}\\
\midrule
\texttt{R} & calif. (4) & \res{5.83}{0.27} & \res{9.21}{0.64} & \bres{7.86}{0.27} & 5.0 & \res{7.01}{0.23}\\
\texttt{R} & income (3) & \res{48.99}{1.60} & \res{47.32}{1.87} & \bres{46.98}{1.85} & 3.4 & \res{43.78}{1.83}\\
\texttt{R} & insurance (3) & \res{251.47}{2.57} & \bres{238.28}{1.18} & \res{249.41}{2.21} & 3.2 & \res{226.85}{1.75}\\
\bottomrule
\end{tabular}
\caption{\textbf{\PUC-compliant models improve predictive performance}. Same setup as in to \Cref{tab:multiplefeatures}, but in this case we use \emph{Extra Trees}.}
\label{tab:multipleextra}
\end{table}

\subsection{Experiment 3: Convergence to analytical \PUC}
\label{sec:app_exp3converge}
In this section, we provide additional details regarding the experiment where we study the gaps to analytical \puc on our simulated data sets.

\subsubsection{Synthetic data sets} 
We initially conduct a synthetic data experiment to verify our theory.

\textbf{First Synthetic Data Set.} For the data set used in \Cref{fig:offconvergence} and \Cref{fig:offconvergencemodels}, we create binary features according to the Naive Bayes scheme described in \Cref{sec:app_naivebayes}. The probabilities of each feature pointing to the corresponding class were drawn randomly, we made the three features with the highest discriminatory power optional. We then drew probabilities of the feature values being missing also at random. FOr this example, the missingness did not depend on the feature value, but only on the label. The resulting parameters are given in \Cref{tab:parameters_naivebayes} for the sake of completeness.

\textbf{Second Synthetic Data Set.} We create a more complicated data set with continuous features as describeb by the familiy in \Cref{sec:app_densityfamiliyofflr}. 
We create two normally distributed base features and three optional features to test interesting dependency combinations by using the parameters in \Cref{tab:parameters}. This distribution includes cases where:
\begin{itemize}
    \item the availability distribution depends on the base features ($\bm{u} \neq \bm{0}$, feature 1)
    \item the availability distribution depends on the class value ($\lambda \neq 0$, feature 1, feature 2)
    \item the feature value depends on the base features and the class value ($\bm{v} \neq \bm{0}$, $\tau(0)\neq \tau(1)$, feature 1, feature 2, feature 3)
\end{itemize}
We draw increasing numbers of samples from the known distribution and fit the corresponding estimators. The test set on which the \PUC-Gap$^2$ is estimated on 5000 independently drawn samples. For \Cref{fig:offconvergencemodels}, we use 50000 samples to train each model.

\subsubsection{Additional Results}
We also conduct the approximation experiment on the more complicated continuous data set. The results can be found in \Cref{fig:offconvergencemodels_app}. Note that on this continuous data sets the models will not perfectly converge. However, we show that the \PUC-Gap is in range of the irreducible, random estimation error, by computing the average squared estimation error without \PUC on the unfair data set and adding the ranges of this error to the plot.

\begin{figure}[ht]
\centering
\includegraphics[width=0.6\textwidth]{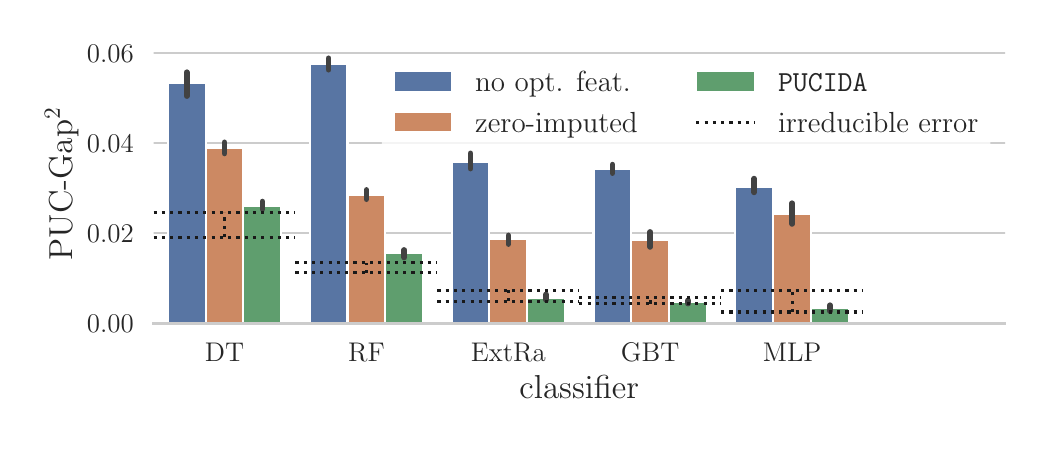}\vspace{-2mm}
\caption{\texttt{\OFFIDA} converges independently of the ML model on the second simulated data set with optional features. The fairness gaps are close to the irreducible model estimation error (due to imperfect models on this continuous data set) when applying our technique across a variety of common models on the continuous simulated data set.\label{fig:offconvergencemodels_app}}
\end{figure}

\begin{table}[h]
    \centering
    \scalebox{0.88}{
    \begin{tabular}{rl}
    \toprule
    data set & optional features \\
    \midrule
    %law &  LSAT ($-$), UGPA ($-$), sex ($+$)\\
    insurance & experience ($-$), kidslt6 ($-$), kids618 ($-$)\\
    %heloc & \begin{tabular}{c} %    NetFractionRevolvingBurden ($+$), AverageMInFile ($-$), MSinceMostRecentInqexcl7days ($-$)\\%    NumInqLast6Mexcl7days ($-$), NumBank2NatlTradesWHighUtilization ($+$)
    %\end{tabular} \\
    adult & age($-$) , educational-num ($-$), hours-per-week ($-$), capital-gain ($-$), capital-loss ($-$)\\
    compas & priors\_count ($+$), age ($-$), c\_days\_from\_compas ($-$), c\_charge\_degree ($+$), juv\_misd\_count ($+$)\\
    diabetes & Glucose ($+$), age ($+$) \\
    california housing  & housing\_median\_age ($+$), population ($-$), households ($-$), median\_income ($-$)\\
    income & AGEP ($-$), SCHL ($-$), WKHP ($-$)\\
    \bottomrule
    \end{tabular}}
    \caption{Features made optional in the experiment with multiple optional features. Direction: ($+$) means higher values more likely to be unavailable, ($-$) indicates lower values to be more likely to be unavailable. The direction was chosen such that feature values that lead to more negative outcomes tend to be undisclosed more frequently.\label{tab:app_offgapfeatures}}
\end{table}

\begin{table}[ht]
    \centering
    \begin{tabular}{rcccc}
    \toprule
        feature & $p(x_i=1|y=0)$ & $p(x_i=1|y=1)$ &$p(a_i=0|y=0)$ & $p(a_i=0|y=1)$ \\
        \midrule
        1 & 0.090 & 0.141 & -- & -- \\
        2 & 0.915 & 0.930 & -- & -- \\
        3 & 0.225 & 0.020 & -- & -- \\
        4 & 0.771 & 0.377 & -- & -- \\
        5 & 0.202 & 0.347 & -- & -- \\
        6 & 0.968 & 0.322 & 0.920 & 0.345 \\
        7 & 0.874 & 0.239 & 0.647 & 0.294 \\
        8 & 0.723 & 0.159 & 0.508 & 0.207 \\
    \bottomrule  
    \end{tabular}
    \caption{Parametric distribution parameters used in the first synthetic data set. Features are all binary. Features 1--5 are base feature which are always available. Features 6 --8 are unavailable with a certain probability given the class label.}
    \label{tab:parameters_naivebayes}
\end{table}

\begin{table}[]
    \centering
    \begin{tabular}{cc}
    \toprule
        base features & n=2, $\bb \sim \mathcal{N}(\bm{0}, 5\bm{I})$, $\bm{w}{=}(-1.5, 1.0)^\top$, $t{=}0$ \\
        opt. feature 1 & $\bm{u}_1 =(0.8, 0.4)^\top$, $\bm{v}_1=(0, 1)^\top$, $\lambda_1{=}0.7$, $\tau_1(0){=}-0.25$, $\tau_1(1){=}0.25$\\
        opt. feature 2 & $\bm{u}_2 =\bm{0}$, $\bm{v}_2=(0, -0.15)^\top$, $\lambda_2{=} 1.0$, $\tau_2(0){=}0.4$, $\tau_2(1){=}-0.4$\\
        opt. feature 3 & $\bm{u}_3 =\bm{0}$, $\bm{v}_3=(0.1, 0.2)^\top$, $\lambda_3{=} 0.0$, $\tau_3(0){=}-0.2$, $\tau_3(1){=}0.2$\\
    \bottomrule  
    \end{tabular}
    \caption{Parametric distribution parameters used in the synthetic data experiment. The used density covers all possible dependencies between availability, feature values and the base features that are allowed by the graphical model.}
    \label{tab:parameters}
\end{table}

\fi

\end{document}